\documentclass[11pt]{article}
\usepackage{float}
\usepackage{amsfonts}
\usepackage[final]{acl}

\usepackage{times}
\usepackage{latexsym}

\usepackage[T1]{fontenc}

\usepackage[utf8]{inputenc}

\usepackage{microtype}

\usepackage{inconsolata}

\usepackage{graphicx}
\usepackage{amsmath}
\usepackage{booktabs}
\usepackage{float}
\usepackage{subcaption}

\usepackage{xcolor}
\usepackage{mdframed}
\usepackage{placeins}
\usepackage{enumitem}
\usepackage{nameref}
\usepackage{algorithm}
\usepackage{algpseudocode}
\usepackage{booktabs}   
\usepackage{multirow}   
\newcommand{\smallsc}[1]{\text{\small\scshape #1}}
\usepackage{newunicodechar}
\newunicodechar{−}{-}
\usepackage{stfloats}
\usepackage{comment}
\usepackage{xcolor}
\usepackage{tikz}
\usepackage{pgfplots}
\usepackage{pgfplotstable}
\usepgfplotslibrary{groupplots}
\pgfplotsset{compat=1.18}
\pgfplotsset{
  every axis/.append style={
    tick label style={font=\scriptsize},
    label style={font=\scriptsize},
    title style={font=\scriptsize},
    legend style={font=\scriptsize},
  }
}
\newmdenv[
  backgroundcolor=gray!15, 
  linecolor=black,         
  leftmargin=0.5pt,
  rightmargin=0.5pt,
  skipabove=1em,
  skipbelow=1em,
  innerleftmargin=6pt,
  innerrightmargin=6pt,
]{smallerbox}

\newenvironment{smallermdframed}
  {\begin{smallerbox}\scriptsize} 
  {\end{smallerbox}}

\usepackage{authblk}

\newcommand{\greencheck}{\textcolor{green!60!black}{\Large$\bullet$}}
\newcommand{\redcross}{\textcolor{red!70!black}{\Large$\bullet$}}
\DeclareMathOperator*{\argmin}{arg\,min}
\newcommand{\inname}{\mathrel{\in_{\text{name}}}}
\newcommand{\notinname}{\mathrel{\notin_{\text{name}}}}

\title{ProActor: Timing-Aware Reinforcement Learning for Proactive Task Scheduling Agents
\thanks{Accepted to ACL 2026.}
}

\author[1]{Lei Ding}
\author[2]{Bin He}
\author[1]{Chenguang Wang}
\author[1]{Yang Liu}
\affil[1]{University of California, Santa Cruz}
\affil[2]{Zillow Group}

\begin{document}
\maketitle

\begin{abstract}

Proactive task-oriented agents must autonomously anticipate user needs, identify actionable opportunities, and trigger software actions at appropriate moments—fundamentally shifting from reactive systems that await explicit instructions. However, existing approaches lack generalizable end-to-end solutions for measuring and optimizing such anticipatory behaviors.

This paper introduces ProActor, a unified framework for conversational task scheduling that integrates: (1) a domain-agnostic automated annotation methodology that enables scalable proactiveness reinforcement learning (RL) by generating full opportunity time windows instead of rigid point labels, (2) systematic proactiveness metrics capturing both timing quality and reference action alignment, and (3) RL optimization using GRPO with various reward designs. Our insight is that RULER-based rewards with proactiveness rubrics are crucial for improving timing quality, and that proactiveness optimization enabled by stage-aware composite rewards is key to balancing timing quality and reference action alignment.

Timing-aware RL requires extensive exploration, demanding efficient infrastructure. We develop ART-F, an adaptive framework combining request-adaptive inference clusters with DDP-based training on single-node multi-GPU systems, enabling LoRA training of 4-bit Qwen2.5-14B-ProActor-Q4 with 4--8$\times$ speedups. Experiments on two newly auto-annotated datasets demonstrate significant improvements in proactive timing while maintaining action consistency comparable to state-of-the-art (SOTA) baselines. Ablations validate the effectiveness of distinct composite reward variations.

\end{abstract}

\begin{figure}
    \centering
    \includegraphics[width=\columnwidth]{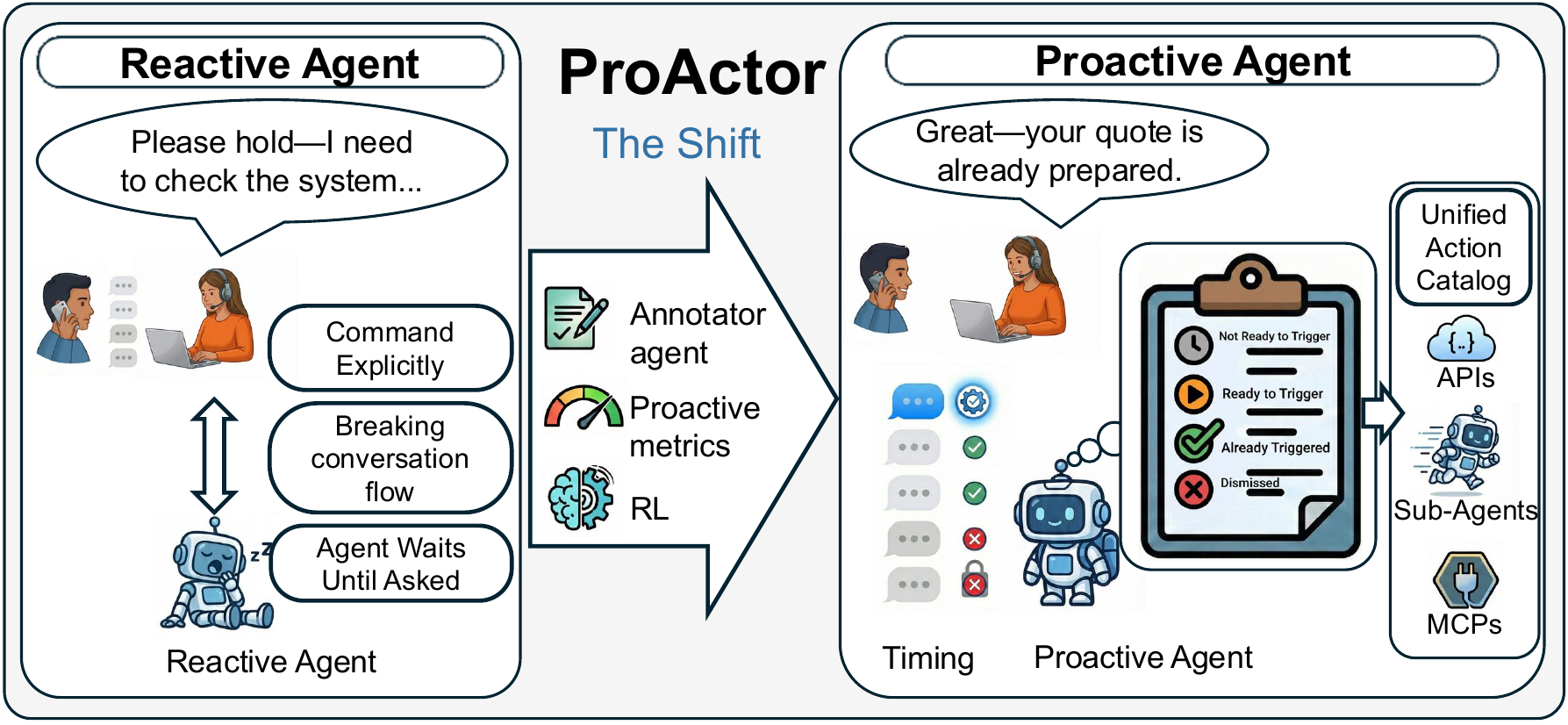}
    \caption{Our ProActor shifts conversational agents from reactive execution to proactive task scheduling. An automated annotation pipeline provides reference actions, proactiveness metrics quantify timing and prediction action alignment, and RL optimizes timing-aware policies over a unified action catalog—enabling agents to act proactively without disrupting conversation flow.}
    \label{fig:0}
\end{figure}
\section{Introduction}


As AI assistants evolve, a paradigm shift is emerging—from \textbf{reactive} systems that wait for explicit instructions to \textbf{anticipatory} agents~\citep{luproactive, wucollabllm}. Empowered by large language models (LLMs), proactive agents can anticipate user needs, surface relevant actions, and guide workflows before being asked. A case in point is an agent that continuously monitors conversations between professionals and clients, identifies emerging opportunities, and intervenes seamlessly when action is appropriate (Figure~\ref{fig:0}). Such ambient agents point toward a future in which AI agents are deeply embedded in daily interactions, supporting decision-making across customer service, enterprise automation, and real-time assistance.


\begin{figure}
    \centering
    \includegraphics[width=\columnwidth]{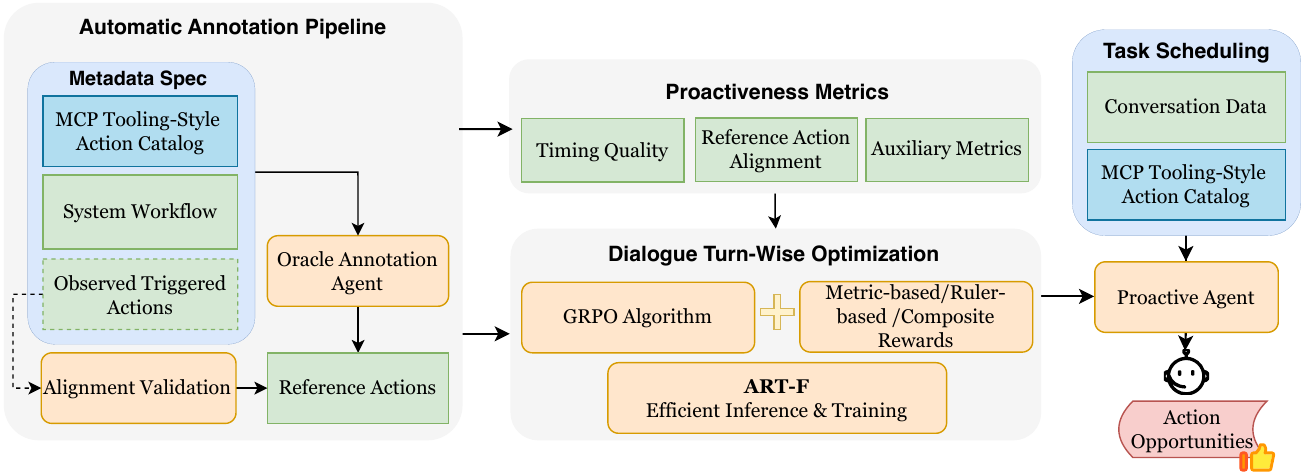}
    \caption{Overview of ProActor for end-to-end proactiveness optimization. ProActor integrates (1) an automated annotation pipeline that generates high-quality MCP-style reference actions, (2) a suite of proactiveness metrics that quantify action quality and timing, and (3) ART-F, an efficient RL framework enabling scalable training in GPU-constrained environments. Proactiveness optimization in conversational task scheduling is formalized as a dialogue turn–level RL problem.}
    \label{fig:1}
\end{figure}

Recent work has explored proactive behavior optimization via stronger prompting~\citep{deng2023prompting}, improved context engineering~\citep{yoon2025beyond}, supervised fine-tuning (SFT)~\citep{dong2025protod}, and reward-driven optimization~\citep{luproactive, wucollabllm}. However, they remain insufficient in conversational task scheduling. First, proactive behavior allows multiple valid timing choices rather than a single correct answer~\cite{sodhi2023effectiveness}, making SFT unsuitable because reproducing exact reference timings penalizes valid alternatives and obscures underlying timing principles~\citep{sodhi2023effectiveness, kim2025principles}. Moreover, the lack of automated pipelines for proactive action annotation~\cite{wu2023fine, sodhi2023effectiveness} further amplifies challenges of proactive optimization. To fill these gaps, we introduce a domain-agnostic automated annotation pipeline that constructs a unified action catalog and leverages an oracle LLM annotator to generate proactive action candidates, referred to as \textbf{reference actions}. Our pipeline also supports evaluating the quality of reference actions with the ready status by the alignment with \textbf{action observations} actually triggered, and preserving only high-quality references.

Second, existing methods~\citep{wu2023fine, deng2023prompting, zhou2024timing} struggle to quantitatively capture timing-sensitive factors—such as readiness and termination—crucial for conversational task scheduling. By contrast, we treat reference actions as proactiveness anchors and define \textbf{a suite of quantitative metrics} to evaluate timing quality and action consistency, complemented by preference-based RULER rewards~\citep{openpipe2025ruler} for ambiguity-aware proactiveness evaluation. Guided by these signals—particularly stage-aware composite rewards—we apply Group Relative Policy Optimization (GRPO)~\citep{shao2024deepseekmath} at the dialogue turn level, formalizing proactiveness optimization as a principled trade-off between better proactive timing and consistency with reference actions.

Last but not least, RL post-training for LLMs is highly resource-intensive, requiring large-scale rollout generation, interaction among policy, reference, and reward models, and tightly coordinated distributed training across GPUs~\citep{ouyang2022training, shao2024deepseekmath}. Motivated by evidence that LoRA-based RL can match full-parameter tuning~\citep{schulman2025lora}, we apply \textbf{Low-Rank Adaptation (LoRA)}~\cite{hu2021lora} to \textbf{quantized LLMs} and introduce Adaptive Resource Training Framework (\textbf{ART-F}), an efficient end-to-end RL framework that integrates a request-adaptive inference cluster with DDP-based training to maximize GPU utilization on single-node, multi-GPU systems. Using ART-F, we train Qwen2.5-14B-ProActor-Q4, a 4-bit LoRA-tuned model, achieving 4–8× training speedups on 4×H200 and 8×H100 GPUs while delivering substantial gains in proactive timing and maintaining SOTA action consistency on two newly annotated datasets, ABCD+ and Home Loan. Ablation studies further confirm that the stage-aware composite reward is critical for achieving our optimization goal. We plan to open source the ART-F and annotation pipelines in the future.


In summary, our main contributions are:
\begin{itemize}[leftmargin=*,itemsep=1pt]
    \item We introduce \textbf{an automated domain-agnostic annotation pipeline} to extract a unified action catalog and generate high-quality reference actions, delivering new datasets: ABCD+ and Home Loan.
    \item We propose \textbf{a comprehensive suite of proactiveness metrics} that quantitatively capture timing quality, alignment between predicted and reference actions, enabling systematic evaluation and optimization of task-scheduling agents.
    \item We formulate proactiveness optimization as \textbf{a trade-off between timing accuracy and reference alignment}, and introduce a dialogue turn–level RL-based proposal that leverages metric-based, RULER-based, and composite rewards to optimize proactive decision-making.
    \item We introduce \textbf{ART-F}, an efficient RL post-training framework for quantized LLMs, and demonstrate its effectiveness by training \textbf{Qwen2.5-14B-ProActor-Q4} that attains 4–8$\times$ faster training while achieving \textbf{strong gains in proactive timing} without sacrificing action consistency.
\end{itemize}

\section{Related Works}

\paragraph{Proactive Agents and Task Scheduling}

Recent advances in proactive agents~\cite{luproactive, wucollabllm, deng2024towards} fundamentally reshape intelligent assistant design~\cite{abbas2025having, liu2025proactive} and their application to task scheduling~\cite{xu2025agenttod, yoon2025beyond, yu2025passive, liu2024compeer}. To optimize proactive behaviors, researchers invested efforts in generalizing tool abstractions, standardizing action representations~\cite{rastogi2020towards, liutoolace, qin2023toolllm} and constructing data pipelines~\cite{zhang2025proactive, wei2025grounded} in specific domains. Despite progress, existing evaluations~\cite{liu2025proactiveeval, luproactive, alshikh2025towards} largely overlook time-sensitive proactive behaviors~\cite{kim2025principles}: they treat proactive timing as a single-answer problem, penalizing predictions that deviate from labels even when earlier timings represent valid proactive behavior~\citep{sodhi2023effectiveness}. This limitation motivates our RL-based approach: we generate \emph{reference action ranges} and use turn-level optimization to explore proactive timing rather than replicating exact one reference point.

\paragraph{RL Optimization \& Efficient Training}

RL has been applied to conversational task scheduling, from dialogue policy learning~\cite{peng2018deep, dora, rescue2024} to multi-step tool optimization~\cite{qin2023toolllm, workbench, progra}. Despite explicit timing models~\cite{chen2025llamapie, zhou2024timing}, time-sensitive proactivity in conversational settings remains underexplored. We address this gap by formulating proactive timing as a step-wise optimization problem with composite rewards to balance early action with correctness. Furthermore, RL training incurs substantial computational cost and infrastructure complexity~\cite{lattimore2013sample, impala2018, quaRL2019}. While distributed RL frameworks improve scalability~\cite{wang2025distflow, bartoldson2025trajectory}, LoRA-based~\cite{hu2021lora} quantized RL on single-node GPUs often suffers from rollout–training imbalance and under-utilization~\citep{dettmers2023qlora, hu2021lora}. ART-F targets this by enabling efficient, collocated RL training through a lightweight request-adaptive inference and asynchronous payload distribution in GPU-constrained environments.

\section{Methodology}

\subsection{Domain-Agnostic Automated Annotation Pipeline for Reference Actions}\label{sec:auto_data_pipeline}

\begin{figure}
    \centering
    \includegraphics[width=\columnwidth]{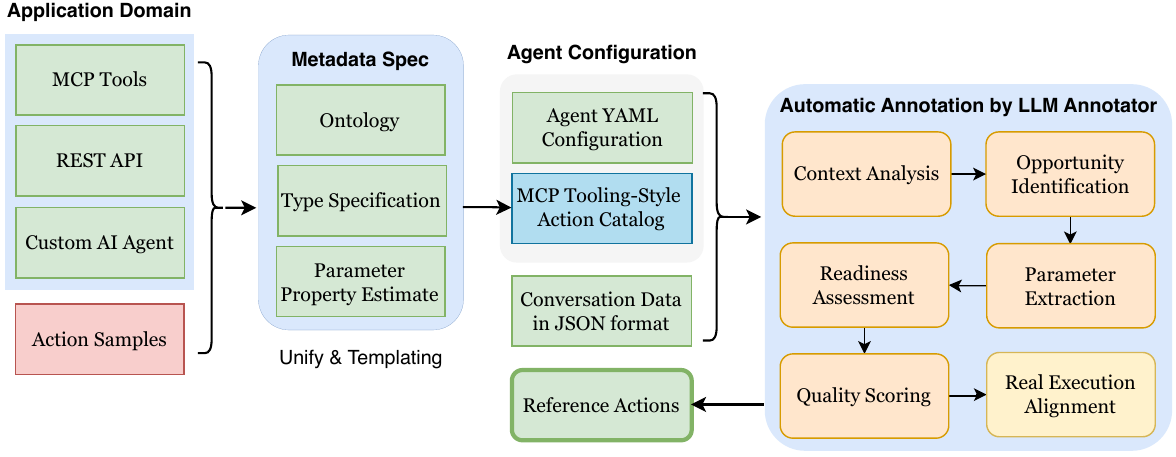}
    \caption{Automated annotation pipeline: Heterogeneous tools are unified through a domain-agnostic, configuration-driven \textit{Catalog Generator}, while an \textit{LLM annotator} with a global dialogue view produces high-quality reference actions at each dialogue turn.}
    \label{fig:2}
\end{figure}

Given RL needs large volumes of reference action ranges, which is infeasible for humans to label, it is crucial to develop an automated pipeline that can produce these annotations at scale. Following~\cite{chen2021action}, we introduce a unified metadata schema that normalizes heterogeneous action interfaces into a standardized JSON representation, capturing action ontology, type signatures, and parameter properties. The schema supports diverse automation backends and is rendered into a unified action catalog (Appendix~\ref{metadata_annotation}) using Jinja2~\cite{jinja_code_generation}.

Given a dialogue and the generated tool catalog, we initialize an oracle LLM annotator (Appendix~\ref{oracle_annotation_agent}) with access to the full conversation—including future turns—and instruct it to identify proactive opportunities at each dialogue turn by following the predefined procedure in Figure~\ref{fig:2}, analogous to hindsight experience replay~\cite{andrychowicz2017hindsight}. This deliberate design endows the annotator with \emph{action-forecasting} capability, enabling it to identify actions that \emph{would have been} beneficial at each moment rather than merely retrospectively explaining outcomes. We further compare against \textbf{a strictly causal oracle annotator} that has no access to future turns, demonstrating the necessity of incorporating hindsight into the annotation process (Appendix~\ref{comparison_to_annotation_without_future_turn}). Importantly, the entire pipeline is \textbf{fully automated and model-agnostic}: adapting to new domains or LLMs requires only configuration changes. 

\paragraph{Reference Actions: Guidance Over Ground Truth.} We term annotation outputs as \textbf{reference actions} because they represent \textit{one or more valid timing choices given each dialogue turn context}---not ground truth. The oracle annotator identifies \textit{suitable} actions, not necessarily \textit{optimal} ones; more aggressive proactive timing could also be valid. This multiplicity motivates RL over SFT: we optimize policies exploring proactive opportunities rather than replicating exact timings. For datasets with action observations, we further validate the alignment between reference actions labeled with ready status and the corresponding observed actions with triggered status(Appendix~\ref{alignment_validation_on_with_reference_dataset}).

\subsection{Proactiveness Metrics}\label{sec:proactiveness_metrics}

Proactive task scheduling requires agents to decide \textit{what}, \textit{how}, and \textit{when} to act. Accordingly, we define a set of metrics that evaluate action quality along two dimensions: \emph{action consistency} (AC, Max AC, and Difference) and \emph{proactive timing} (PT, FTR, and RAR). For convenience, predictions with status \textsc{READY\_TO\_TRIGGER} or \textsc{TRIGGERED} are treated as \emph{ready actions}. Our setting permits partial parameter specification and explicit readiness tracking, unlike standard tooling calls that assume complete parameter mappings~\cite{patil2025bfcl}. Formally, given a unified action catalog $\mathcal{A}$, we evaluate agent behavior at each dialogue turn $t$ by comparing agent prediction actions $\hat{A}_t$ with reference actions $A^{\text{ref}}_t$. 

\paragraph{Action Consistency (AC)} measures the average alignment level between $\hat{A}_t$ and $A^{\text{ref}}_t$:
\begin{equation}
\resizebox{\columnwidth}{!}{$
\text{AC}(\hat{A}_t, A^{\text{ref}}_t) =
\frac{1}{|\hat{A}_t|} 
\sum_{\hat{a}_i \in \hat{A}_t}
\max\limits_{\substack{
a_j \in A^{\text{ref}}_t \\
a_j.\text{name}=\hat{a}_i.\text{name}
}}
\frac{C(R(\hat{a}_i),R(a_j)) + C(O(\hat{a}_i),O(a_j))}{|R(a_j)| + |O(a_j)|}
$}
\end{equation}
where $R(\cdot)$, $O(\cdot)$ denote required and optional parameters, and $C(\cdot,\cdot)$ is the parameter match score, i.e., the fraction of parameters that match between prediction and reference.

\paragraph{Maximum AC (Max AC)} captures the best alignment level between $\hat{A}_t$ and $A^{\text{ref}}_t$:
\begin{equation}
\text{Max AC}(\hat{A}_t, A^{\text{ref}}_t) = \max_{\hat{a} \in \hat{A}_t} \text{AC}(\{\hat{a}\}, A^{\text{ref}}_t)
\end{equation}

\paragraph{Consistency Difference (Difference)} measures prediction reliability as the relative gap between AC and Max AC. Given AC statistics $(A, \delta_A)$, Max AC statistics $(M, \delta_M)$ over multiple runs, we define
\begin{equation}
\mu = \frac{M - A}{A},
\delta \approx \sqrt{\left(\frac{\delta_M}{A}\right)^2 + \left(\frac{M \cdot \delta_A}{A^2}\right)^2}
\end{equation}
A high value indicates inconsistent predictions, while a low value reflects more stable behaviors.

\paragraph{Reference Range}
\footnotemark\footnotetext{
For $A_t \in \{\hat{A}_t$, $A^{\text{ref}}_t\}$, we denote $a \inname A_t$, if there exists $a' \in A_t$ such that $a'.\text{name} = a.\text{name}$; correspondingly, $a \notinname A_t$ denotes that no such $a'$ exists.
} 
characterizes how ready actions are distributed across dialogue turns. For action $a \in \mathcal{A}$, its reference range is defined as
\begin{equation}
\mathcal{R}_a = \{\, t \mid a \inname A^{\text{ref}}_t  \land a \text{ is ready at turn }  t \,\},
\end{equation}
and the total reference range is $\mathcal{R} = \bigcup_{a \in \mathcal{A}} \mathcal{R}_a$.



\paragraph{Proactive Timing (PT)}\label{proactive_timing_measure} rewards prediction actions that occur \textit{no later than} the reference-ready window:
\begin{equation}
\text{PT}(t, \hat{A_t}, \mathcal{R}) = \frac{\sum_{\hat{a} \in \hat{A_t}} \mathbb{I}[\exists \tau \in \mathcal{R}_{\hat{a}} : \tau \geq t]}{|\hat{A_t}|}
\end{equation}

\paragraph{Fault Trigger Rate (FTR)} penalizes predicted ready actions that fall outside the reference coverage. Denoting the set of predicted ready actions at dialogue turn $t$ as
\[
\hat{A_t}^{\text{ready}} = \{\, \hat{a} \in \hat{A_t} \land \hat{a} \text{ is a ready action} \,\},
\]
FTR is defined as
\begin{equation}
\text{FTR}(t, \hat{A}_t^{\text{ready}}, \mathcal{R}) = \frac{|\{\hat{a} \in \hat{A}_t^{\text{ready}} \land \hat{a} \notin \mathcal{R}\}|}{|\hat{A}_t^{\text{ready}}|}
\end{equation}

\paragraph{Ready Action Rate (RAR)}\label{ready_action_rate_measure} measures the proportion of predictions marked ready at turn $t$:
\begin{equation}
\text{RAR}(t, \hat{A_t}) = \frac{|\{\hat{a} \in \hat{A_t} \land \hat{a}\text{ is a ready action}\}|}{|\hat{A_t}|}
\end{equation}
Low RAR indicates conservatism; high RAR with high FTR suggests over-eager triggering.

More metrics refer to Appendix~\ref{more_evaluation_metrics}.

\subsection{Reinforcement Learning Optimization}\label{sec:rl_method}

\paragraph{Task Definition}
Given a reference-ready dataset and a unified action catalog, our goal is to optimize a policy model $\pi$ that, at dialogue turn $t$, produces a candidate action set $\hat{A_t}$. The policy is trained to maximize accumulated rewards by aligning $\hat{A_t}$ with the corresponding reference actions $A^{\text{ref}}_t$, measured both action consistency and proactive timing. This formulation raises questions: how can reward design and reward granularity effectively translate proactiveness metrics and evaluation rubrics into learning signals?

\begin{figure}[t!]
    \centering
    \includegraphics[width=\columnwidth]{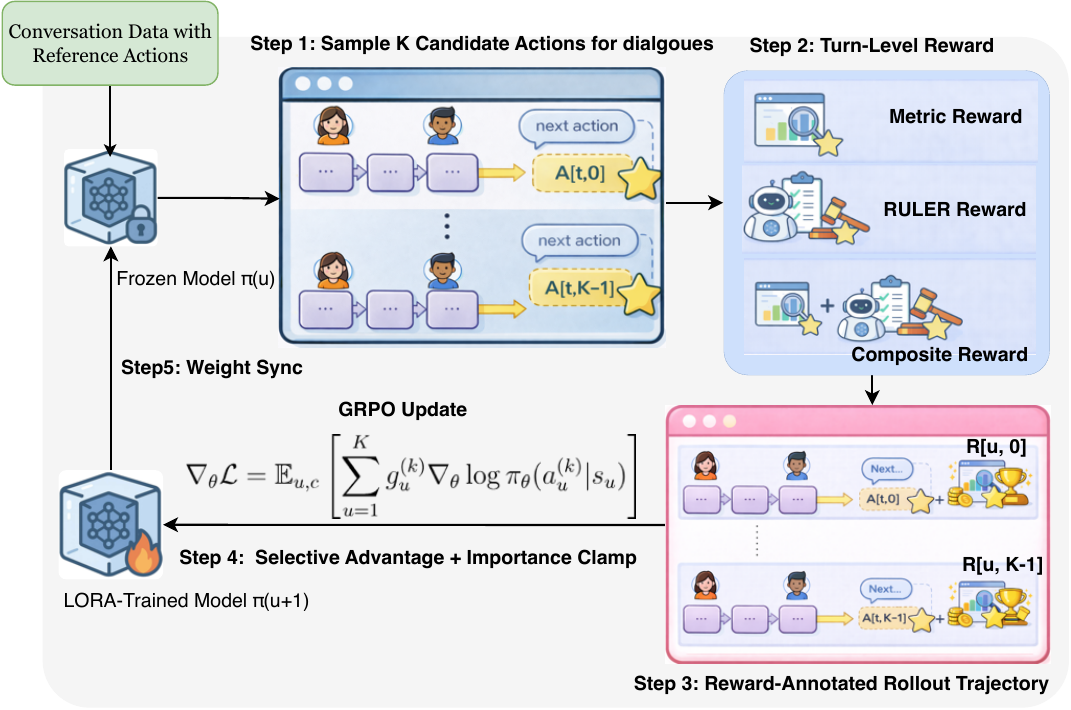}
    \caption{Turn-level GRPO optimization: we sample $K$ action candidates from the policy model $\pi_{\theta}(u)$ for each dialogue turn, score action trajectories with the chosen reward, update $\pi_{\theta}(u)$ to $\pi_{\theta}(u+1)$ using GRPO with reward-weighted gradients.}
    \label{fig:3}
\end{figure}

\paragraph{Reward Granularity Decision} In general, \textbf{trajectory-level rewards}, which average action feedback over an entire dialogue, are ill-suited for task scheduling: long dialogues ($\geq$ 23 turns) worsen credit assignment, early actions create cascading effects that are hard to disentangle, and sparse end-of-trajectory signals provide limited learning guidance. These observations are consistent with our empirical findings (Section~\ref{reward_granularity}), motivating the adoption of \textbf{turn-level reward modeling} that evaluates action quality per dialogue turn.

\paragraph{Turn-Level Optimization}
At optimization step\footnote{In this paper, $t$ denotes dialogue turns; $u$ indexes RL steps} $u$, we rollout $K$ times for each dialogue turn and update the policy model $\pi_\theta$ using turn-level GRPO, as illustrated in Figure~\ref{fig:3}:

\begin{equation}
\resizebox{\columnwidth}{!}{$
\nabla_\theta \mathcal{L} = \mathbb{E}_{u,c}\left[\sum_{k=1}^{K} g_u^{(k)} \nabla_\theta \log \pi_\theta(a_u^{(k)}|s_u)\right]
$}
\end{equation}
where $c$ is the episode context, $s_u, a_u^{(k)}$ are the state/action. $g_u^{(k)}$ is defined as the turn-level advantage $A_u^{(k)}$ scaled by an importance ratio $\bar r_u^{(k)}$ (set 10 based on Appendix~\ref{importance_sampling_ratio_cap} and~\ref{turn_level_grpo_equation}), with PPO-style clipping applied to limit gradient magnitude.

We further design \textbf{several turn-level rewards}, beginning with \emph{single-objective rewards} that just optimize action consistency or timing.

\paragraph{Metric-based rewards} use AC and Max AC to compute the alignment value between $\hat{A_t}$ and $A^{\text{ref}}_t$ as reward signals, referred to as \textit{RAC} and \textit{Max RAC}, respectively, which are intentionally designed to improve action consistency.

\paragraph{RULER-based rewards} ART’s default RULER~\citep{openpipe2025ruler, hilton2025art}, referred to as \textbf{General RULER}\label{general_ruler}, applies a generic evaluation rubric to assess how well predicted actions follow prompt instructions. We also introduce \textbf{Custom RULER}\label{custom_ruler} to \textbf{explicitly evaluate proactive behaviors} and expect it to benefit the optimization of proactive timing behaviors.

Next, we propose \emph{composite rewards} that combine action consistency and timing with fixed coefficients:

\paragraph{Weighted Metric Reward} Since PT and FTR are relatively sparse timing signals and AC/Max AC yields dense action consistency signals, we define a simple  combination reward:
\begin{equation}
\resizebox{\columnwidth}{!}{$
R_{\mathrm{wM}}^{\mathcal{F}}(u) = \mathcal{F}(u) + w_1 \cdot \mathrm{PTR}(u) - w_2 \cdot \mathrm{FTR}(u)
$}
\end{equation}
where $\mathcal{F} \in \{\mathrm{AC}, \mathrm{MaxAC}\}, w_1 = 0.05, w_2 = 0.01$.

Furthermore, we introduce \emph{stage-aware rewards} to adjust factor weights as the training progresses.

\paragraph{Adaptive Metric Reward} Given a total of $U$ training steps, this design tries to encourage broad exploration early (Max RAC + a small bonus for PT) when $u < \frac{U}{3}$ as \textit{the exploration phase}, then transitions to balanced optimization (RAC with the increased bonus for PT and penalty for FTR) when $\frac{U}{3} \leq u < \frac{2U}{3}$ as \textit{the balanced phase}, and finally becomes conservative (RAC with the final penalty for FTR) late in conversations to avoid noise when $u \geq \frac{2U}{3}$ as \textit{the conservative phase}. 

\begin{equation}
\resizebox{\columnwidth}{!}{$
R_{\text{adM}}(u) = \begin{cases}
w_1 \cdot \text{MaxRAC}(u) + (1 - w_1) \cdot \text{PTR}(u) &u < \frac{U}{3} \\
w_2 \cdot \text{RAC}(u) + w_3 \cdot \text{PTR}(u) - (1 - w_2 - w_3) \cdot \text{FTR}(u) &\frac{U}{3} \leq u < \frac{2U}{3}\\
w_4 \cdot \text{RAC}(u) - (1 - w_4) \cdot \text{FTR}(u) & \text{otherwise} \label{eq:adaptive_metrics}
\end{cases}
$}
\end{equation}
where $w_1 = 0.8$, $w_2 = 0.6$, $w_3 = 0.3$, and $w_4 = 0.6$ are fixed coefficients, selected empirically.

\paragraph{Adaptive RULER Reward} As RULER-based rewards tend to gradually overemphasize timing and Metric-based rewards promote action consistency but weaken timing, we design adaptive formulations balancing these two aspects:
\begin{equation}
R_{\text{adR}}(u) = (1-\lambda_u) R_{\text{metric}}(u) + \lambda_u R_{\text{RULER}}(u) \label{eq:adaptive_ruler}
\end{equation}
where $\lambda_u$ increases from 0 to $\lambda_{\max}=0.3$, progressively emphasizing timing quality until the upper limit. More auxiliary metrics are in Appendix~\ref{sec:rl_details}.

\subsection{Efficient Training Framework: ART-F}\label{sec:art_f_runtime_framework}

\begin{figure}
    \centering
    \includegraphics[width=\columnwidth]{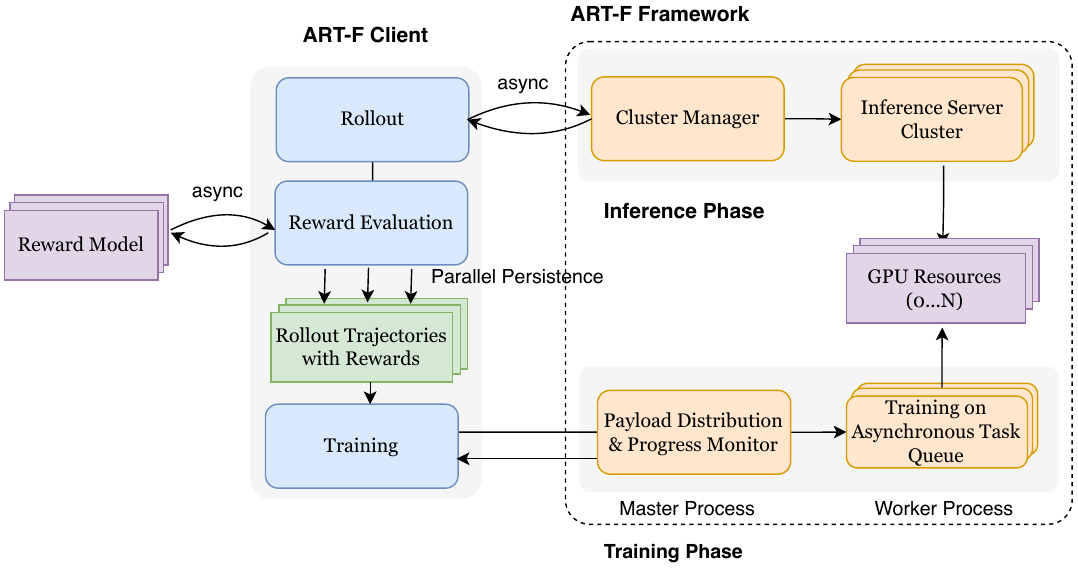}
    \caption{ART-F framework: An efficient collocated RL training system that combines an adaptive inference cluster with asynchronous, distributed training to maximize GPU utilization.}
    \label{fig:4}
\end{figure}

To address ART~\citep{hilton2025art}'s inefficiency under LoRA-based RL training, we introduce ART-F (Figure~\ref{fig:4}) that extends ART to enable an efficient collocated RL training~\citep{wu2025llamarL, wang2025colocating} through improved resource orchestration (Appendix~\ref{sec:art_f_framework} and~\ref{sec:art_f_client_implementation}).

\paragraph{Adaptive Inference Cluster}
ART-F dynamically instantiates multiple vLLM~\cite{kwon2023efficient} servers per GPU, each reserving a fixed memory fraction, thereby increasing parallel request capacity without exceeding memory limits. A centralized load-balancing cluster manager continuously monitors server health, throughput, and latency, exposing real-time routing information to rollout clients, enabling informed scheduling decisions—such as selecting available servers or early terminating rollouts during saturation—
substantially improving inference throughput and GPU utilization.

\paragraph{Asynchronous Payload-Distribution Training}
ART-F implements a custom asynchronous payload-distribution training framework with a master–worker architecture. The master process coordinates training by partitioning rollout trajectory groups, distributing them to worker processes, and tracking execution progress via asynchronous queues. ART-F supports \textit{symmetric(replicated) and asymmetric (partitioned) DDP~\cite{pytorchDDP, unsloth_ddp} modes} (Appendix~\ref{master_process_payload_distribution}), enabling trade-offs between stability and data efficiency, while step-level synchronization ensures consistent gradient updates. This design mitigates rollout–training imbalance, supports dynamic batch sizing, and enables stable, high-throughput RL optimization on a single multi-GPU node.

\section{Experiments}
\subsection{Datasets}

\paragraph{With action observations: ABCD+} extends the ABCD dataset~\citep{chen2021action} with proactive annotations in customer-support scenarios, representing settings \emph{with} historical action execution logs. It contains 7,042 dialogues (5,647/703/692 train/dev/test), 114,978 annotated actions, and an average dialogue length of $21.2 \pm 7.2$ turns.  
\paragraph{Without action observations: Home Loan} is constructed from mortgage consultation transcripts between loan officers and clients (proprietary dataset), representing domains \emph{without} observed software action triggers. It includes 968 dialogues (774/97/97 train/dev/test), 30,610 annotated actions, conversations averaging $47.4 \pm 1.1$ turns\footnote{Dataset details are in Appendix~\ref{sec:dataset_details}}.

\subsection{Setup}

\paragraph{Baselines} We design
\textbf{(1) Non-Reasoning}, where the model directly identifies action opportunities, status, and timing;
\textbf{(2) Reasoning} requires the model to explicitly reason (via \texttt{<think>}) before predicting action opportunities, status, and timing;
\textbf{(3) Reasoning + ASG} maintains an action-state graph (ASG) to track predicted actions across turns and conditions the reasoning on this evolving action state; 
\textbf{(4) SFT} fine-tune the model on the annotated dataset, validating the necessity of RL. We test baselines on GPT-5.1, Gemini-2.5-flash, Claude-sonnet-4\footnote{Abbreviated to Claude-4 in the remaining sections}, Qwen2.5-14B-Instruct\footnote{Baseline details are provided in Appendix~\ref{sec:baseline_details}. We evaluate SFT variants on the ABCD+ dataset, which offers a larger number of samples than Home LOAN; additional details are given in Appendix~\ref{sft_baseline_setup} and~\ref{sft_baseline_comparison}}.

\paragraph{RL Training}
We train Qwen2.5-14B-ProActor-Q4, a 4-bit quantized Qwen2.5-14B-Instruct model optimized via LoRA\footnote{rank 8, $\alpha=16$, applied to attention and MLP projections.}, using our ART-F client (Appendix~\ref{cluster_configuration}). Training is performed on 4$\times$H200 (ABCD+) and 8$\times$H100 (Home Loan) with a maximum 9,216-token context. We enable symmetric DDP training (Appendix~\ref{symmetric_vs_non_symmetric_epoch}) for 2 epochs, equivalent to 8 (ABCD+) and 16 (Home Loan) epochs under non-symmetric partitioning. End-to-end training completes in 3.5–5.7 days on ABCD+ and 1.5–2.15 days on Home Loan \footnote{Full training settings are in Appendix~\ref{sec:training_setup}}.

\paragraph{Proactiveness Ranking Index (PRI)} To measure whether models strike a better proactiveness balance, we introduce PRI (Appendix~\ref{proactiveness_ranking_index}), a composite ranking score that aggregates Consistency Index (CI) and Timing Index (TI) using harmonic mean: $\text{PRI} = \frac{2 \times \text{CI} \times \text{TI}}{\text{CI} + \text{TI}}$ where CI combines action consistency metrics (AC, Max AC, Difference) and TI combines timing metrics (PT, FTR, RAR), all min-max normalized within comparison groups. Rankings are reported separately for the main results (Table~\ref{table:main_results}) and ablations (Table~\ref{table:ablation_reward_merged}).

\subsection{Main Results}

\begin{table*}[t!]
    \centering
    \setlength\tabcolsep{3pt}
    \resizebox{\textwidth}{!}{
    \begin{tabular}{l|c|ccc|ccc}
    \hline\toprule
     & \textbf{PRI}{\scriptsize~$\uparrow$}
     & \multicolumn{3}{c|}{\textbf{Consistency}}
     & \multicolumn{3}{c}{\textbf{Timing}} \\
    \cline{3-8}
     &
     & \textbf{AC}{\scriptsize~$\uparrow$}
     & \textbf{Max AC}{\scriptsize~$\uparrow$}
     & \textbf{Difference}{\scriptsize~$\downarrow$}
     & \textbf{Proactive Timing}{\scriptsize~$\uparrow$}
     & \textbf{Fault Trigger Rate}{\scriptsize~$\downarrow$}
     & \textbf{Ready Action Rate}{\scriptsize~$\uparrow$}  \\
    \hline\toprule

    \multicolumn{6}{l}{\textbf{ABCD+}} \\
    \hline
    GPT-5.1 Non-Reasoning
    & 0.5104
    & 0.318$\pm$0.005 & 0.706$\pm$0.012 & 0.717$\pm$0.013
    & 0.2023$\pm$0.0019 & 0.0460$\pm$0.0009 & 0.419$\pm$0.010 \\
    GPT-5.1 Reasoning
    & 0.6003
    & 0.429$\pm$0.006 & 0.789$\pm$0.002 & 0.839$\pm$0.026
    & 0.1643$\pm$0.0018 & \textbf{0.0165$\pm$0.0002} & 0.214$\pm$0.003 \\
    GPT-5.1 Reasoning + ASG
    & 0.5547
    & 0.420$\pm$0.004 & 0.733$\pm$0.007 & 0.712$\pm$0.063
    & 0.1874$\pm$0.0022 & 0.0647$\pm$0.0019 & 0.281$\pm$0.003 \\
    Gemini-2.5-flash Non-Reasoning $^{(4)}$
    & \textbf{0.6251}
    & 0.417$\pm$0.004 & \textbf{0.834$\pm$0.001} & 1.000$\pm$0.019
    & 0.2133$\pm$0.0030 & 0.0399$\pm$0.0006 & 0.288$\pm$0.003	 \\
    Gemini-2.5-flash Reasoning
    & 0.6216
    & \textbf{0.430$\pm$0.001} & 0.813$\pm$0.002 & 0.891$\pm$0.006
    & 0.1782$\pm$0.0011 & \textbf{0.0245$\pm$0.0006} & 0.252$\pm$0.001	 \\
    Gemini-2.5-flash Reasoning + ASG
    & 0.5257
    & 0.384$\pm$0.001 & 0.737$\pm$0.001 & 0.919$\pm$0.006
    & 0.1715$\pm$0.0015 & 0.0286$\pm$0.0005 & 0.241$\pm$0.004	 \\
    Claude-4 Non-Reasoning
    & 0.6216
    & 0.427$\pm$0.001 & 0.831$\pm$0.001 & 0.946$\pm$0.005
    & 0.2119$\pm$0.0013 & 0.0526$\pm$0.0012 & 0.297$\pm$0.001 \\
    Claude-4 Reasoning $^{(3)}$
    & \textbf{0.6318}
    & 0.421$\pm$0.001 & 0.749$\pm$0.004 & 0.779$\pm$0.010
    & 0.2136$\pm$0.0023 & 0.0482$\pm$0.0023 & 0.314$\pm$0.003 \\
    Claude-4 Reasoning + ASG
    & 0.6031
    & 0.403$\pm$0.005 & \textbf{0.852$\pm$0.004} & 1.114$\pm$0.028
    & 0.2269$\pm$0.0022 & 0.0634$\pm$0.0032 & 0.360$\pm$0.002 \\
    Qwen2.5-14B-Instruct Non-Reasoning
    & 0.2996
    & 0.295$\pm$0.004 & 0.564$\pm$0.002 & 0.914$\pm$0.004
    & 0.2237$\pm$0.0012	 & 0.0773$\pm$0.0010 & 0.515$\pm$0.002 \\
    Qwen2.5-14B-Instruct Reasoning
    & 0.6246
    & 0.423$\pm$0.005 & 0.449$\pm$0.006 & 0.062$\pm$0.019
    & 0.1842$\pm$0.0012 & 0.0307$\pm$0.0008 & 0.279$\pm$0.003 \\
    Qwen2.5-14B-Instruct Reasoning + ASG
    & 0.4331
    & 0.316$\pm$0.004 & 0.385$\pm$0.004 & 0.218$\pm$0.020
    & 0.1918$\pm$0.0043 & 0.0432$\pm$0.0011 & 0.359$\pm$0.004 \\
    Qwen2.5-14B-Instruct + SFT
    & 0.1700
    & 0.272$\pm$0.001 & 0.533$\pm$0.0002 & 0.960$\pm$0.020
    & 0.2097$\pm$0.0007 & 0.0912$\pm$0.0003 & 0.531$\pm$0.003 \\
    Qwen2.5-14B-ProActor-Q4 + Custom RULER $^{(1)}$
    & \textbf{0.7293}
    & 0.426$\pm$0.015 & 0.484$\pm$0.019 & 0.136$\pm$0.048
    & \textbf{0.2347$\pm$0.0201} & 0.0708$\pm$0.0078 & \textbf{0.546$\pm$0.036} \\
    Qwen2.5-14B-ProActor-Q4 + Adaptive RULER $^{(2)}$
    & \textbf{0.6842}
    & \textbf{0.431$\pm$0.022} & 0.586$\pm$0.044 & 0.320$\pm$0.123
    & \textbf{0.2515$\pm$0.0272} & 0.1089$\pm$0.0083 & \textbf{0.521$\pm$0.052} \\
    \hline\toprule

    \multicolumn{6}{l}{\textbf{Home Loan}} \\
    \hline
    GPT-5.1 Non-Reasoning
    & 0.5047
    & 0.272$\pm$0.001 & 0.467$\pm$0.003 & 0.717$\pm$0.013
    & 0.0462$\pm$0.0013 & 0.0049$\pm$0.0002 & 0.116$\pm$0.003	 \\
    GPT-5.1 Reasoning
    & 0.5047
    & 0.363$\pm$0.007 & \textbf{0.579$\pm$0.004} & 0.595$\pm$0.033
    & 0.0186$\pm$0.0026 & \textbf{0.0003$\pm$0.0003} & 0.031$\pm$0.002\\
    GPT-5.1 Reasoning + ASG
    & 0.5010
    & 0.281$\pm$0.007 & 0.481$\pm$0.013 & 0.712$\pm$0.063
    & 0.0374$\pm$0.0061 & 0.0034$\pm$0.0004	 & 0.105$\pm$0.010  \\
    Gemini-2.5-flash Non-Reasoning
    & 0.6165
    & 0.349$\pm$0.004 & \textbf{0.688$\pm$0.004} & 0.972$\pm$0.024
    & 0.0632$\pm$0.0011 & 0.0072$\pm$0.0005 & 0.173$\pm$0.005 \\
    Gemini-2.5-flash Reasoning $^{(1)}$
    & \textbf{0.7303}
    & 0.345$\pm$0.004 & 0.527$\pm$0.006	& 0.528$\pm$0.024
    & 0.0757$\pm$0.0011 & \textbf{0.0001$\pm$0.0001} & 0.241$\pm$0.002 \\
    Gemini-2.5-flash Reasoning + ASG
    & 0.6067
    & 0.283$\pm$0.007 & 0.479$\pm$0.002 & 0.693$\pm$0.042
    & 0.0764$\pm$0.0039 & 0.0063$\pm$0.0018 & 0.251$\pm$0.012 \\
    Claude-4 Non-Reasoning
    & 0.5416
    & 0.224$\pm$0.003 & 0.347$\pm$0.004 & 0.549$\pm$0.024
    & \textbf{0.0811$\pm$0.0029} & 0.0118$\pm$0.0008 & \textbf{0.332$\pm$0.002}\\
    Claude-4 Reasoning $^{(3)}$
    & \textbf{0.7039}
    & 0.332$\pm$0.006 & 0.472$\pm$0.003 & 0.422$\pm$0.028
    & 0.0742$\pm$0.0027 & 0.0063$\pm$0.0016 & 0.261$\pm$0.003\\
    Claude-4 Reasoning + ASG $^{(2)}$
    & \textbf{0.7262}
    & \textbf{0.375$\pm$0.002} & 0.607$\pm$0.006 & 0.619$\pm$0.022
    & 0.0760$\pm$0.0031 & 0.0127$\pm$0.0013 & 0.307$\pm$0.005\\
    Qwen2.5-14B-Instruct Non-Reasoning
    & 0.4288
    & 0.253$\pm$0.002 & 0.383$\pm$0.003 & 0.514$\pm$0.017
    & 0.0120$\pm$0.0006 & 0.0003$\pm$0.0004 & 0.054$\pm$0.005 \\
    Qwen2.5-14B-Instruct Reasoning
    & 0.4012
    & 0.217$\pm$0.005 & 0.239$\pm$0.009 & 0.101$\pm$0.049
    & 0.0037$\pm$0.0010 & 0.0006$\pm$0.0001 & 0.032$\pm$0.003 \\
    Qwen2.5-14B-Instruct Reasoning + ASG
    & 0.3223
    & 0.113$\pm$0.011 & 0.144$\pm$0.006 & 0.274$\pm$0.133
    & 0.0184$\pm$0.0083 & 0.0033$\pm$0.0022 & 0.090$\pm$0.014 \\
    Qwen2.5-14B-ProActor-Q4 + Custom RULER
    & 0.5603
    & 0.206$\pm$0.024 & 0.234$\pm$0.021  & 0.137$\pm$0.161
    & \textbf{0.0846$\pm$0.0095} & 0.0355$\pm$0.0079 & \textbf{0.465$\pm$0.074}  \\
    Qwen2.5-14B-ProActor-Q4 + Adaptive RULER $^{(4)}$
    & \textbf{0.6232}
    & \textbf{0.395$\pm$0.029} & 0.466$\pm$0.038 & 0.180$\pm$0.129
    & 0.0501$\pm$0.0055 & 0.0131$\pm$0.0013 & 0.156$\pm$0.009 \\
    \bottomrule
    \end{tabular}}
    \caption{Model performance comparison on the test set: Baselines are averaged over 3 runs; ProActor-Q4 results are aggregated over the last $N{=}4$ checkpoints. Adaptive RULER uses Max RAC with $\lambda_{\max}=0.3$. (1)-(4) indicate the top-1 to top-4 methods ranked by PRI per group.}
    \label{table:main_results}
\end{table*}

\paragraph{Baseline Analysis}

\textit{(1) Reasoning improves consistency but harms timing.} Reasoning reduces Consistency Difference (–19.9/-28.6\%\footnote{$t_1/t_2$ indicates the dataset-specific metric value, with $t_1$ for ABCD+ and $t_2$ for Home Loan } on average for non-Qwen baselines, and over –80\% for Qwen), while making AC/Max AC vary across models. However, reasoning degrades timing by $\approx$10\%, with only isolated gains for Claude-4 on ABCD+ and Gemini-2.5-flash on Home Loan (Table~\ref{tab:reasoning_change_ratio}). 

\textit{(2) ASG introduces varied timing gains at a substantial consistency cost.} ASG adds varied PT gains (-3.8\% to 14.1\% on ABCD+ and 0.9\% to 101.1\% on Home Loan) but degrades AC/Max AC (2\% to 22.6\%, except Claude-4 on Home Loan) and increases inconsistency (10\% to 40\%) for most models, showing that structure alone can't hit a better consistency and timing balance (Table~\ref{tab:asg_change_ratio}). 

\textit{(3) No baseline resolves the consistency-timing tradeoff.} Across both datasets, baselines achieving high Max AC ($\geq$0.8) incur large Consistency Difference (0.75 to 1.11) and exhibit conservative behavior with low PT ($\leq$0.23) and RAR ($\leq$0.36); while some baselines approach ProActor’s PT, none maintain Difference $< 0.2$, confirming our claim (Appendix~\ref{no_baseline_strick_balance}).

\paragraph{ProActor-Q4} 
\textit{RL enables 4-bit ProActor-Q4 models to achieve a better trade-off between action consistency and proactive timing}, as shown in Table~\ref{table:main_results}. We attribute it to reward designs that optimize action consistency via RAC/Max RAC, while leveraging explicit rubrics for proactiveness evaluation in RULER-based rewards. Two ProActor-Q4 variants exhibit complementary strengths:

\textit{Custom RULER} achieves the best timing overall, with PT = 0.2347/0.0846 and RAR = 0.546/0.465, higher than the strongest baselines(typically PT$\leq$0.2023/0.0811, RAR$\leq$0.288/0.251), while keeping Consistency Difference at 0.136/0.137, comparable to or lower than most SOTA baselines.

\textit{Adaptive RULER} delivers the most balanced improvement, reaching the highest AC (0.431/0.395) among all methods and strong PT(0.2515/0.0501), while maintaining a lower Difference than high-Max AC baselines(e.g., GPT-5.1, Gemini, Claude often exceed 0.7 to 1.1 Difference).

Explicit reasoning induces hesitation, while ProActor breaks this paralysis by using RL to internalize timing intuition. ASG's accumulation of past opportunities creates a reasoning burden that distracts static policies. RL offers a feasible path to effectively harness the rich ASG memory rather than becoming overwhelmed by it.


\subsection{Ablation Study}

\begin{table*}[t!]
\centering
\setlength\tabcolsep{3pt}
\resizebox{\textwidth}{!}{
\begin{tabular}{l|c|c|cc|cc|cc|cc|cc}
\hline\toprule
\textbf{\#Dialogs} 
& \textbf{Reward Type}
& \textbf{\textbf{PRI}{\scriptsize~$\uparrow$}}
& \multicolumn{2}{c|}{\textbf{AC}{\scriptsize~$\uparrow$}}
& \multicolumn{2}{c|}{\textbf{Max AC}{\scriptsize~$\uparrow$}}
& \multicolumn{2}{c|}{\textbf{Proactive Timing (PT)}{\scriptsize~$\uparrow$}}
& \multicolumn{2}{c|}{\textbf{Fault Trigger Rate (FTR)}{\scriptsize~$\downarrow$}} 
& \multicolumn{2}{c}{\textbf{Ready Action Rate (RAR)}{\scriptsize~$\uparrow$}}  \\
\cline{4-13}
\textbf{Train/Test} 
& 
&
& Statistics & $\Delta$
& Statistics & $\Delta$
& Statistics & $\Delta$
& Statistics & $\Delta$ 
& Statistics & $\Delta$\\
\hline

\multicolumn{10}{l}{\rule{0pt}{2.5ex}\textbf{ABCD+}} \\
\hline

100/50 
& RAC   
& 0.2596   
& 0.3239$\pm$0.01606 & -0.1528
& 0.3881$\pm$0.02590 & -0.1348 
& 0.1223$\pm$0.02364 & -0.1030 
& \textbf{0.0640$\pm$0.01960} & 0.9636  
& 0.3002$\pm$0.08088 &  0.2770  \\ 

100/50 
& Max RAC      
& 0.2264    
& 0.3263$\pm$0.02468 & -0.4740	
& \textbf{0.4384$\pm$0.02264} & -0.1587
& 0.1643$\pm$0.00207 & -0.0616	 
& 0.0886$\pm$0.01212 & 2.65
& 0.3862$\pm$0.01111 & 0.2262 \\

100/50 
& General Ruler $\dagger$ 
& \textbf{0.4918}    
& 0.3494$\pm$0.02714  & -0.2750
& 0.3978$\pm$0.02984 & -0.2436
& \textbf{0.2324$\pm$0.01984}  & 0.9056
& 0.1019$\pm$0.01910 & 3.7188 
& \textbf{0.5945$\pm$0.05405} & 1.4867 \\

100/50 
& Custom Ruler  $\star$
& \textbf{0.6140}    
& \textbf{0.3850$\pm$0.02544} & -0.1995
& 0.4203$\pm$0.03200  & -0.2476
& 0.2315$\pm$0.02393 & 0.6755
& 0.1068$\pm$0.02327  & 4.8358 
& 0.5707$\pm$0.05717  & 1.2846 \\

\hline

5647/692
& RAC 
& 0.4352      
& 0.3368$\pm$0.02401 & -0.1478
& 0.3857$\pm$0.01949 & -0.1327
& 0.1799$\pm$0.00452 & 0.0128
& 0.0621$\pm$0.00765 & 0.1760 
& 0.4148$\pm$0.02354 & 0.1818  \\

5647/692
& Max RAC  
& 0.3737      
& 0.3745$\pm$0.03439 & -0.0926
& 0.5014$\pm$0.06809 & 0.0492
& 0.1172$\pm$0.03605 & -0.0351
& \textbf{0.0365$\pm$0.01013} & 0.0919
& 0.2426$\pm$0.06160 & -0.0008  \\

5647/692
& Custom Ruler $\star$  
& \textbf{0.7217}    
& 0.4257$\pm$0.01469 & -0.4791
& 0.4837$\pm$0.01848 & -0.4731
& 0.2347$\pm$0.02014 & 0.56561
& 0.0708$\pm$0.00784 & 2.4654 
& 0.5456$\pm$0.04169 & 1.8482 \\

5647/692
& Weighted Metric (Max RAC) 
& 0.2169    
& 0.3068$\pm$0.02606 & -0.2975
& 0.4399$\pm$0.05253 & 0.0011
& 0.1465$\pm$0.03091 & 0.0338
& 0.0700$\pm$0.01253 & 1.5204
& 0.3769$\pm$0.05893 & 0.3607\\

5647/692
& Adaptive Metric 
& 0.5159   
& 0.3537$\pm$0.02264 & -0.1950
& 0.3922$\pm$0.03489 & -0.2524
& 0.2268$\pm$0.04591 & 0.2966
& 0.1137$\pm$0.01925 & 2.5661
& \textbf{0.6332$\pm$0.11113} & 1.0600 \\

5647/692
& Adaptive RULER ($\lambda_{\max}=0.3$, Max RAC) $\dagger$
& \textbf{0.6026}   
& \textbf{0.4314$\pm$0.02240} & -0.05045
& \textbf{0.5861$\pm$0.04335} & 0.1150  
& \textbf{0.2515$\pm$0.02722} & 0.4410 
& 0.1089$\pm$0.00832 & 1.9003
& 0.5212$\pm$0.05153 & 0.5818 \\

\hline\toprule


\multicolumn{10}{l}{\textbf{Home Loan}} \\
\hline

774/97
& RAC   
& 0.5668   
& \textbf{0.4701$\pm$0.03285} & 0.8483 
& \textbf{0.5195$\pm$0.03016} & 0.8462
& 0.0264$\pm$0.00590 & -0.2502 
& 0.0041$\pm$0.00245 & -0.7494
& 0.0612$\pm$0.01856 & -0.5365 \\

774/97
& Max RAC  
& 0.4311
& 0.3894$\pm$0.01327 & 0.4891
& 0.4780$\pm$0.02065 & 0.6927
& 0.0093$\pm$0.00378 & -0.8214 
& \textbf{0.0015$\pm$0.00072} & -0.9409
& 0.0177$\pm$0.00787 & -0.9178\\

774/97
& Custom Ruler 
& 0.4220   
& 0.2057$\pm$0.02409 & -0.3760
& 0.2338$\pm$0.02114 & -0.3329
& \textbf{0.0846$\pm$0.00951} & 2.5245
& 0.0355$\pm$0.00785 & 5.2040
& \textbf{0.4653$\pm$0.07412} & 4.8508\\

774/97
& Weighted Metric (RAC) 
& 0.5376   
& 0.4359$\pm$0.01376 & 0.6695
& 0.4805$\pm$0.02399 & 0.7160 
& 0.0205$\pm$0.00349 & -0.2859
& 0.0020$\pm$0.00066 & -0.8762
& 0.0481$\pm$0.00722 & -0.6039  \\

774/97
& Weighted Metric (Max RAC) 
& 0.4809  
& 0.4133$\pm$0.02947 & 0.6461
& 0.5346$\pm$0.03620 & 0.9601
& 0.0270$\pm$0.00730 & -0.4212 
& 0.0049$\pm$0.00090 & -0.4268 
& 0.0506$\pm$0.02000 & -0.7042\\

774/97
& Adaptive Metric 
& 0.5742  
& 0.4677$\pm$0.01760 & 0.8090
& 0.5184$\pm$0.02184 & 0.8390 
& 0.0282$\pm$0.00552 & -0.3170
& 0.0042$\pm$0.00121 & -0.7683
& 0.0653$\pm$0.01628 & -0.5919 \\

774/97
& Adaptive RULER ($\lambda_{\max}=0.3$, RAC) $\star$ 
& \textbf{0.6154}   
& 0.4173$\pm$0.00768 & 0.5111
& 0.4403$\pm$0.01019 & 0.4510
& 0.0397$\pm$0.00528 & 0.4382
& 0.0071$\pm$0.00070 & -0.1682  
& 0.1231$\pm$0.01985 & 0.4329     \\

774/97
& Adaptive RULER ($\lambda_{\max}=0.3$, Max RAC) $\dagger$ 
& \textbf{0.5734} 
& 0.3950$\pm$0.02936 & 0.5366
& 0.4658$\pm$0.03821 & 0.6764
& 0.0501$\pm$0.00547 & 1.0451
& 0.0131$\pm$0.00131 & 1.5299
& 0.1561$\pm$0.00930 & 0.5190  \\

\bottomrule
\end{tabular}}
\caption{Performance under different rewards on the test set. $\Delta$ denotes the relative change of each metric after RL training w.r.t.\ its pre-training value. $^{\star}$ and $^{\dagger}$ indicate top-1 and top-2 methods by PRI per group.}
\label{table:ablation_reward_merged}
\end{table*}

\paragraph{Reward Variations} \textbf{RULER-based} rewards favor smaller-scale settings, where data patterns are simple and limited. In contrast, \textbf{Adaptive RULER} better captures the proactiveness trade-off at larger scales, though some reward variants outperform on individual metrics. In addition, Adaptive RULER combined with RAC typically achieves a higher PRI than its Max RAC counterpart, because AC aims at optimizing average rather than the best-case action alignment.

\textit{Single-objective Rewards at ABCD+ 100/50} RULER-based rewards dominate PRI (0.49 to 0.61) over RAC and Max RAC (0.23 to 0.26), driven by substantially higher PT ($\approx$0.23 vs.\ 0.12 to 0.16) and RAR ($\approx$0.57 to 0.59 vs.\ 0.30 to 0.39). While Max RAC rewards maintain a strong Max AC(0.4384), they exhibit conservative behavior with smaller PT($\leq$0.165) and RAR($\leq$0.4). As General RULER underperforms Custom RULER in PRI and other metrics are close, we omit it in later experiments.

\textit{Reward Variations at ABCD+ 5647/692 \& Home Loan 774/97} Adaptive RULER attains top-2 PRI(0.60–0.62) on both dataset, reflecting a balanced trade-off. In contrast, some single-objective rewards (e.g. RAC on Home Loan) achieve strong AC/Max AC (up to 0.52) but exhibit low PT and RAR (PT < 0.03), showing a greater imbalance.

\paragraph{RULER robustness}
To validate the robustness of RULER-based rewards, we examine three dimensions: judge model, evaluation rubric, and task domain.
\textit{(i) Cross-judge:} Training Adaptive RULER with three judge models- GPT-4.1-mini, Gemini-2.5-flash, and Claude-Opus-4.6 on 100/50 dialogues yields consistent timing (PT: 0.169-0.190, RAR: 0.365-0.413) and action quality (AC: 0.400-0.429), with ${\leq}$7\% inter-judge variation (Appendix~\ref{sec:judger_model}).
\textit{(ii) Cross-rubric:} General and Custom RULER both improve timing over baselines; Custom RULER further raises PRI (0.614 vs.\ 0.492, Table~\ref{table:main_results}).
\textit{(iii) Cross-domain:} Gains persist on both ABCD+ and Home Loan despite different action distributions (Tables~\ref{table:main_results},~\ref{table:ablation_reward_merged}).

\begin{figure}[t!]
    \centering
    \includegraphics[width=\columnwidth]{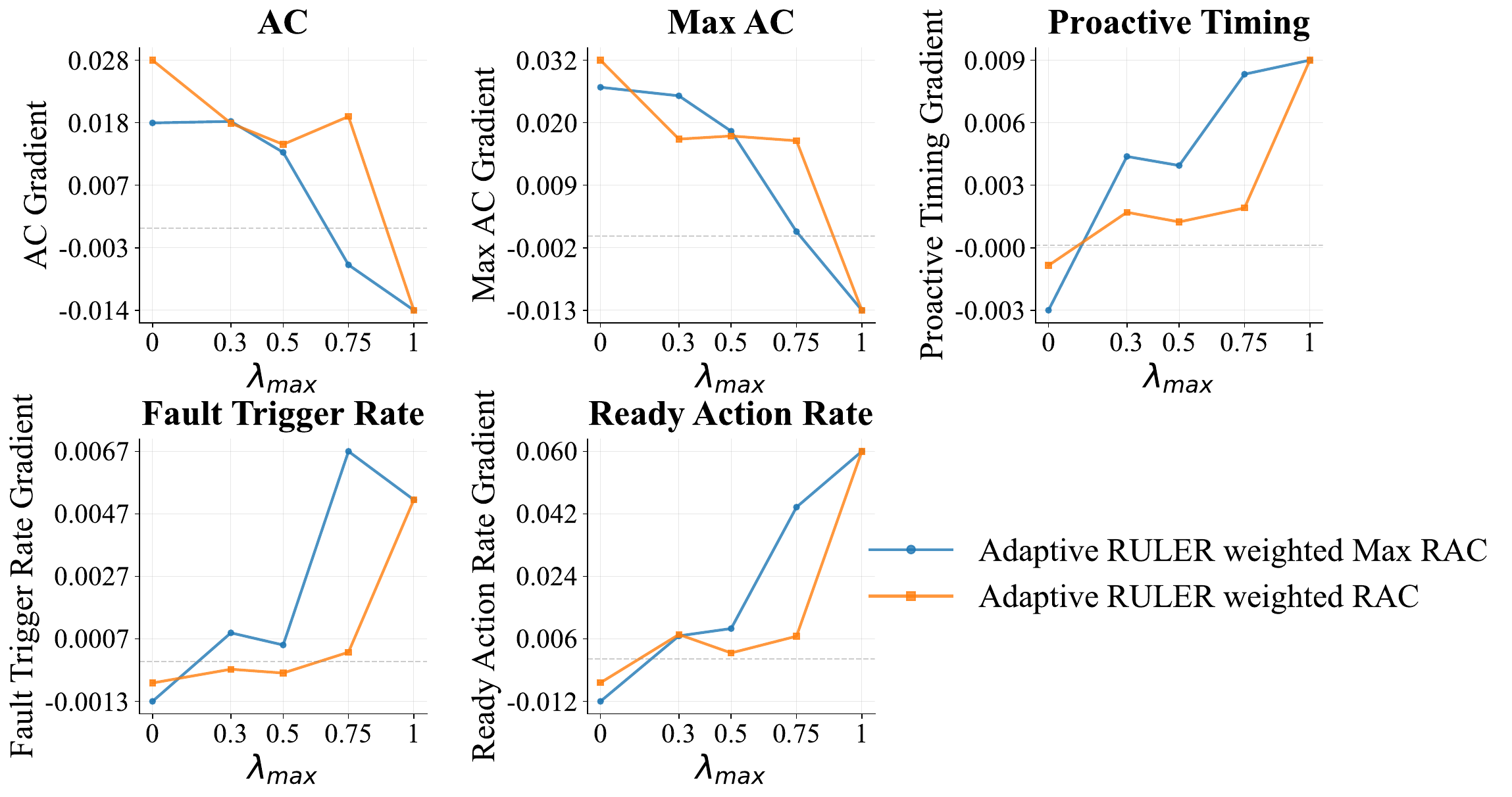}
    \caption{Effect of $\lambda_{max}$ on Adaptive RULER weighted by Max RAC (blue) and RAC (orange), showing how \textbf{Average Gradient} of six main metrics change with $\lambda_{max} \in \{0, 0.3, 0.5, 0.75, 1.0\}$.}
    \label{fig:4_ablation}
\end{figure}

\paragraph{$\lambda_{\max}$ Effect.} For Adaptive RULER rewards weighted by RAC and Max RAC, $\lambda_{\max}$ functions as a speed-control knob for proactive timing: those with higher $\lambda_{\max}$ tend to optimize timing metrics more aggressively. We enumerate $\lambda_{\max} \in \{0, 0.3, 0.5, 0.75, 1.0\}$ on the Home Loan 774/97 setting, with 5 runs per configuration, and compute \textbf{the average metric gradients on the testing set for each neighborhood training step} in each run. Illustrated by Figure~\ref{fig:4_ablation}, the average gradients of PT, FTR, and RAR scale \textbf{approximately proportionally with} $\lambda_{\max}$, whereas those of AC and Max AC exhibit \textbf{an inverse trend} (Appendix~\ref{sec:ablation_details}).

\section{Conclusion}

We present ProActor, a unified framework for training timing-aware proactive task-scheduling agents. Our approach integrates: (1) a domain-agnostic annotation pipeline generating \textit{reference actions} with multiple valid timing choices, (2) systematic proactiveness metrics with rubric evaluations, and (3) turn-level GRPO with several reward variations, especially composite rewards to controllably balance action consistency and timing. The key insight is that turn-level rewards with an explicit evaluation rubric for proactivity are essential for learning timing-aware policies.

Experiments show significant improvements in proactive timing while maintaining action consistency. Meanwhile, ART-F enables this at scale, with a 4-8$\times$ speedup. Domain adaptation with only configuration changes validates ProActor's generalizability for real-world task automation.

\section{Limitations}

\paragraph{Action Observation Scope.}\label{sec:limitations_observation_scope}
A fundamental limitation concerns the nature of observed triggers in our evaluation. Current datasets only capture actions that \textit{actually occurred} in the system, but suitable proactive actions form a broader range---not limited to what happened in the observed trajectory. This is precisely why we term our annotations \textit{reference actions} rather than ground truth, which represent one valid timing choice among potentially many. Our quality filtering aligns against these actual triggers, but this represents a subset of valid opportunities. To address this, we describe a human annotation protocol (Appendix~\ref{sec:human_annotation_protocol}) for capturing the earliest valid triggering time---and optionally the full valid timing window---for each action, enabling more comprehensive evaluation of proactive behaviors beyond observed trajectories.

\paragraph{Evaluation Scope.}

Our experiments focus on two English-language datasets: ABCD+ (7,042 dialogues) and Home Loan (968 dialogues). Multilingual evaluation is an important open direction: different languages exhibit distinct conversational norms---such as turn-taking patterns, levels of directness, and formality conventions---that may affect when proactive actions are contextually appropriate. Low-resource languages introduce additional data sparsity challenges for both annotation quality and RL training stability. Our annotation pipeline's configuration-driven, language-independent design (Section~\ref{sec:auto_data_pipeline}) requires only configuration changes to adapt to new languages, lowering the barrier to multilingual extension. We plan to release the annotation tools and processed ABCD+ annotations to support community expansion to additional languages and domains.

\paragraph{Model and Training Constraints.}

Our RL experiments focus on 4-bit quantized Qwen2.5-14B-Instruct with LoRA adapters. While the RL formulation (turn-level GRPO), reward computation (metric-based and RULER), and ART-F training infrastructure are model-agnostic by design, we have not yet validated generalization across model families (e.g., Llama, Mistral), parameter scales, or quantization schemes. Extending to other architectures requires no changes to the framework---only model-specific LoRA and quantization configurations---making cross-family evaluation a natural next step. Similarly, applying our pipeline to multilingual LLMs or language-adapted variants (e.g., multilingual Gemma, Aya) would enable proactive timing optimization across diverse linguistic contexts. We limit training to dialogues with $\leq$50 turns; performance on longer conversations remains to be evaluated.

\paragraph{ART-F Extension.}

ART-F currently supports training on a single node with multiple GPUs. For our ProActor-Q4 model (14B parameters, 4-bit quantization, 9,216-token context), each vLLM inference server requires approximately 45~GB of GPU memory (Appendix~\ref{cluster_configuration}). The minimum viable configuration is 4 GPUs with $\geq$80~GB VRAM each (e.g., 4$\times$A100-80GB or 4$\times$H100), supporting one inference server per GPU during the rollout phase and DDP training across all GPUs. Our experiments used 4$\times$H200 (141~GB each) and 8$\times$H100 configurations, achieving 4--8$\times$ training speedups (Appendix~\ref{tensor_parallelization_and_speed_up}). Extending ART-F to multi-node distributed training via ML platforms (Databricks, Azure, Google Cloud) or community serverless GPU resources~\citep{awesome-serverless-gpu} is a natural next step toward further scalability.

\section{Ethical Considerations}
This work utilizes two datasets: a previously published public corpus of customer support transcripts ABCD+ and a proprietary dataset of loan officer-borrower interactions. Due to the sensitivity of the financial-related data, neither the proprietary dataset nor the model fine-tuned by the proprietary data will be publicly released. To ensure data privacy and security, the model training and experimental analysis were conducted within a secure compute environment with strict access controls, consistent with our organization's compliance standards. 

We emphasize that this work is strictly experimental. Any real-world deployment of the framework would necessitate rigorous fairness testing to mitigate potential allocative harms. LLM and AI agents that misjudge situations may trigger unwanted actions, and miscalibration remains possible in out-of-distribution scenarios. We recommend thorough human-in-the-loop assessment before any production use to ensure both robustness and operational safety.

\paragraph{Consent and Transparency.} In deployed proactive systems, users should be informed that an AI assistant may initiate actions on their behalf. For irreversible actions (e.g., submitting applications, initiating transactions), explicit user consent should be obtained before execution. Our framework's action status taxonomy---distinguishing \textsc{PENDING}, \textsc{READY\_TO\_TRIGGER}, and \textsc{TRIGGERED} states---naturally supports a propose-and-confirm workflow where the agent identifies opportunities but defers execution to human authorization.

\paragraph{Human-Override Safeguards.} A human professional should retain the ability to override, delay, or cancel any action proposed by the proactive agent. Production deployments should implement confirmation gates for high-stakes actions and provide clear audit trails of all agent-proposed and agent-executed actions. Our experimental evaluation measures action \textit{proposals} rather than autonomous execution, which aligns with a human-in-the-loop deployment model where the agent augments---rather than replaces---professional decision-making.

\paragraph{Bias Risks in Sensitive Domains.} In domains such as financial services, proactive timing decisions could inadvertently correlate with protected attributes if training data reflects historical service disparities. For instance, if certain customer demographics historically received less proactive service, an RL agent trained on such patterns may learn to replicate these inequities. Before deployment in sensitive domains, we recommend auditing proactive timing distributions across demographic groups and incorporating fairness-aware constraints into the reward design where disparities are detected.

\section*{Acknowledgments}
We thank Andy Martin for valuable feedback and guidance throughout this work, and Jyoti Prakash Maheswari, Taleb Zeghmi, Supriya Anand, and Amir Reza Rahmani for helpful technical discussions. This work was conducted during an internship at Zillow Group.

\bibliography{custom}

\appendix

\section{More Evaluation Metrics}\label{more_evaluation_metrics}
Besides metrics presented in Section~\ref{sec:proactiveness_metrics}, we include several auxiliary evaluation metrics that capture alternative aspects. Because of limited experimental bandwidth, these metrics are reported only for baseline evaluations and are omitted for RL-trained models.

\paragraph{Information Consistency (IC)} In task-oriented conversations, questions are typically used to fill information gaps required to advance the task. To assess whether an agent can proactively pose appropriate questions that naturally align with human behavior, we evaluate its ability of raising appropriate questions that gather missing information in a manner consistent with the original conversations. Specifically, an appropriate question should satisfy two criteria: (1) it exhibits high semantic similarity to questions posed by human professionals, and (2) it corresponds to missing parameters in the action state graph. Formally, we define Information Consistency (IC) based on the average semantic similarity between predicted questions and the corresponding reference questions in the original dialogue turn, as follows: 

\begin{equation}
\text{IC}(Q^{\text{ref}}_t, \hat{Q}_t) = \frac{1}{|Q^{\text{ref}}_t||\hat{Q}_t|} \sum_{\substack{q^{\text{ref}} \in Q^{\text{ref}}_t \\\hat{q} \in \hat{Q}_t}} \text{sim}(q^{\text{ref}}, \hat{q})
\end{equation}

where $\hat{Q}_t, Q^{\text{ref}}_t$ are the predicted question set and the reference question set at the same dialogue turn $t$, $\text{sim}(\cdot, \cdot)$ is the semantic similarity function between two questions.

\paragraph{Action Dependency Alignment (AD)} Since predicted actions are scheduled sequentially, valid action sequences should respect the system’s default workflows and dependency constraints. We therefore define Action Dependency Alignment (AD) to measure the extent to which consecutive predicted actions $\hat{a_{t-1}}, \hat{a_t}$ conform to these dependencies:

{\small
\begin{equation}
\text{AD}(\hat{A}) = \frac{1}{|\hat{A}_t|} \sum_{\hat{a}_t \in \hat{A}_t} \left( \sum_{\hat{a}_{t-1} \in \hat{A}_{t-1}} P(\hat{a}_t \mid \hat{a}_{t-1}) \right)
\end{equation}
}

where $P(\hat{a}_t \mid \hat{a}_{t-1})$ represents the conditional probability of action $\hat{a}_t$ following $\hat{a}_{t-1}$ based on the learned action dependency graph $G$ from the system's default workflows stored in the knowledge base $\mathcal{A}$. Specifically, $P(\hat{a}_t \mid \hat{a}_{t-1})$ is computed by normalizing 1-gram frequency counts from $G$ \textbf{with Laplace smoothing for unseen transitions}:

\begin{equation}
P(\hat{a}_t \mid \hat{a}_{t-1}) = \frac{C(\hat{a}_{t-1}, \hat{a}_t) + \epsilon}{\sum_{\hat{a}' \in \mathcal{A}} [C(\hat{a}_{t-1}, \hat{a}') + \epsilon]}
\end{equation}
where $\epsilon = 1e^{-7}$ as the smoothing parameter for unseen action pairs, $C(\hat{a}, \hat{a}_t)$ is the count of transitions from action $\hat{a}_{t-1}$ to action $\hat{a}_t$ in the knowledge base $\mathcal{A}$, $\sum_{\hat{a}'} C(\hat{a}_{t-1}, \hat{a}')$ is the total count of all transitions starting from action $\hat{a}_{t-1}$.

Without losing generality, the rate-based metrics introduced so far treat datasets with and without observed triggers equally. However, \textbf{in settings without observed triggers at the dialogue turn $t$}, the alignment between predicted action opportunities $\hat{a}_t \in \hat{A}_t$ and the truly-triggered actions in the system $A_t$ requires more special considerations. To this end, we introduce the following three metrics.

\paragraph{Actual Trigger Rate (ATR)} \label{actual_trigger_rate} To measure how many triggered actions actually appear in the predicted action opportunities, we use the percentage of triggered actions out of all identified opportunities:

\begin{equation}
\resizebox{\columnwidth}{!}{$
\text{ATR}(t, \hat{A}_t)
=
\frac{
\left|\left\{
\hat{a} \in \hat{A}_t \land \hat{a} \inname A^{\text{ref}}_t \land
\hat{a}\text{ is a TRIGGERED action}\}
\right\}\right|
}{
|\hat{A}_t|
}
$}
\end{equation}

\paragraph{False Triggered Action Rate (FTR)} \label{false_triggered_action_rate} As a counterpart to ATR, we measure the proportion of falsely triggered actions—i.e., actions predicted as triggered do not appear in the observed triggered action set. FTR quantifies the extent of over-triggering by computing the fraction of predicted actions that are not actually triggered:

\begin{equation}
\resizebox{\columnwidth}{!}{$
\text{FTR}(t, \hat{A}_t)
=
\frac{
\left|\left\{
\hat{a} \in \hat{A}_t \land \hat{a} \notinname A^{\text{ref}}_t \land
\hat{a}\text{ is a TRIGGERED action}\}
\right\}\right|
}{
|\hat{A}_t|
}
$}
\end{equation}

\paragraph{False Ready Action Rate (FRR)} \label{false_ready_action_rate_measure} As the counterpart to RAR~\ref{ready_action_rate_measure},  we measure the proportion of incorrectly marked-as-ready actions that do not appear in the reference action set $A_t$:

{\small
\begin{equation}
\resizebox{\columnwidth}{!}{$
\text{FRR}(t, \hat{A}_t)
=
\frac{
\left|\left\{
\hat{a} \in \hat{A}_t \land
\hat{a} \notinname A^{\text{ref}}_t \land
\hat{a} \text{ is a ready action}\}
\right\}\right|
}{
|\hat{A}_t|
}
$}
\end{equation}
}

Together, ATR, FTR, and FRR serve as complementary metrics for evaluating whether actions are triggered appropriately, ensuring correct activation while minimizing unnecessary side effects.

\paragraph{Proactiveness Rating (PR)} \label{subjective_proactive_timing_measure} To quantify the degree of proactiveness exhibited by predicted action opportunities, we adopt an LLM-based subjective evaluation that assigns a discrete proactiveness score to each action on a scale from $-1$ to $5$:
\begin{itemize}
\item 5: Highly proactive assistance without any explicit request
\item 4: Proactive with minimal prompting
\item 3: Moderately proactive
\item 2: Slightly proactive
\item 1: Minimally proactive
\item -1: Cannot be evaluated
\end{itemize}

In contrast to \textbf{Proactive Timing (PT)} in Section~\ref{proactive_timing_measure}, which provides an objective measure of alignment with reference actions, \textbf{PR} serves as a simplified "subjective" assessment of proactive behavior, evaluated by a profile-based LLM-based judger configured (Appendix~\ref{fig:baseline1_config1_non_cot}, \ref{fig:baseline1_config1_cot} and \ref{fig:baseline1_config1_asg}).

\section{Annotation Data Pipeline} \label{annotation_data_pipeline}\label{sec:annotation_pipeline}

\begin{figure}
    \centering
    \includegraphics[width=\columnwidth]{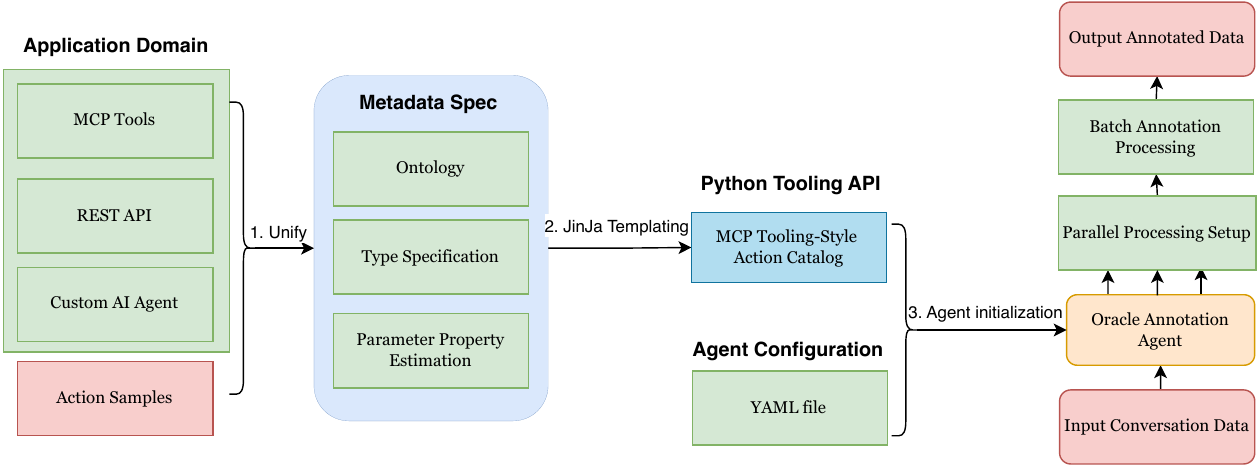}
    \caption{The annotation data pipeline unifies domain metadata and dialogue inputs into standardized tool catalogs, initializes a configurable AI agent, and performs scalable batch annotation to produce structured annotated outputs.}
    \label{fig:7}
\end{figure}

The workflow of the annotation data pipeline is illustrated in Figure~\ref{fig:7}. Section~\ref{metadata_annotation} briefly describes how metadata specifications are generated from automation domain APIs and automation samples. Section~\ref{oracle_annotation_agent} then details how the oracle agent is initialized using a configuration-driven design, and how conversation data are partitioned and annotated in parallel to produce the final annotated outputs. Finally, we describe the alignment validation procedure for datasets with action observations, such as ABCD+.

\subsection{Metadata Specification \& Tooling Catalog Model}\label{metadata_annotation}

Metadata specification generation aims to define and reconsolidate task automation operations in a unified representation, enabling the subsequent task scheduling process to be \textbf{reformulated as action opportunity discovery in a manner analogous to MCP-style tooling calls}. A key distinction from standard MCP tooling, however, is that MCP tool calls typically assume complete parameter mapping, restricting agents to cases where all required parameters have been fully resolved. In contrast, our framework \textbf{explicitly supports partial parameter mapping and readiness analysis}, which is essential for the annotation process as well as for baseline and RL-trained models. 

As the entry point of the entire pipeline, metadata specification generation interfaces with target systems through supported APIs—including MCP tools, REST APIs, or custom AI agents—to extract the following components:

\begin{enumerate}
\item \textbf{Ontology}: a JSON file that defines the hierarchical structure of actions, the workflows composed of these actions, and their inter-action relationships.
\item \textbf{Type specification}: a JSON file that describes each supported action, including its semantic description and detailed parameter types.
\item \textbf{Parameter property estimation}: a JSON file that determines whether each parameter is required or optional. To derive this, we employ a heuristic approach that generates automation samples from system interactions and evaluates parameter presence across all cases. Parameters observed in every instance are marked as \emph{required}, while others are marked as \emph{optional}.
\end{enumerate}

These three components jointly serve as inputs to a Jinja-based\cite{jinja_code_generation} text templating process, which compiles them into a self-described Python dictionary containing all available function call definitions. This resulting artifact constitutes what we refer to as the \emph{MCP tooling-style action catalog}.

\subsubsection{Unified Tool Catalog Architecture}

Our tool catalog supports three distinct types of automation capabilities:

\begin{enumerate}
\item \textbf{MCP (Model Context Protocol) Tools}: Standard MCP-compliant tools with JSON schema validation, providing native compatibility for modern LLM tool use.

\item \textbf{API Definitions}: REST API endpoints with configurable headers, methods, and parameters, enabling integration with existing services.

\item \textbf{Custom AI Agents}: Complex multi-step workflows and specialized automation routines that extend beyond simple API calls.
\end{enumerate}

Essentially, this unified workflow converts domain-specific metadata into standardized and consistent tool catalogs that integrate seamlessly with the annotation pipeline, independent of the underlying business domain.

\subsubsection{Configuration-Driven Design}

The annotation system adopts a configuration-driven design in which a unified tool catalog is specified via JSON configuration files. This abstraction enables rapid adaptation to new domains without requiring code-level modifications. Figure~\ref{fig:tool_catalog_json} illustrates a representative tool catalog schema.

This explicit separation between tool definitions and annotation logic allows the same annotation pipeline to generalize across domains—ranging from customer service scenarios in the ABCD dataset to financial workflows such as loan applications—without architectural changes. Moreover, the configuration-based design facilitates the evolution of available tools by supporting straightforward addition, deprecation, or modification of catalog entries.

\begin{figure}[htbp!]
\begin{smallermdframed}
\text\{\par
\text\ "catalog\_metadata": \{\par
\text\ \ \ "name": "domain\_catalog",\par
\text\ \ \ "version": "1.0.0",\par
\text\ \ \ "domain": "target\_domain",\par
\text\ \}\,\par
\text\ "mcp\_tools": [...],\par
\text\ "api\_definitions": [...],\par
\text\ "custom\_agents": [...]\par
\text\}\par
\end{smallermdframed}
\caption{A JSON snippet of Tool Catalog}
\label{fig:tool_catalog_json}
\end{figure}

\subsubsection{Oracle Annotation Process}

Given a dialogue and tool catalog, an LLM-powered oracle agent identifies proactive opportunities by:

\begin{enumerate}
\item \textbf{Context Analysis}: Processing the full dialogue to understand user intent and information state
\item \textbf{Opportunity Identification}: Matching dialogue context against available tools to identify actionable moments
\item \textbf{Parameter Extraction}: Determining which tool parameters can be filled from the conversation
\item \textbf{Readiness Assessment}: Evaluating whether sufficient information exists to trigger each tool
\item \textbf{Quality Scoring}: Assigning confidence scores to filter high-quality annotations
\end{enumerate}

The oracle sees the complete conversation (past and future), providing reference annotations for what proactive actions would have been beneficial at each turn. This hindsight-enabled annotation creates realistic training targets that online agents can learn to approximate.

\subsubsection{Validation with Action Observations}

For systems/datasets with execution logs (we formally name them after action observations), e.g., ABCD, we align output prediction annotations with action observations (in ABCD, they are triggered actions at the ASSISTANT turns) to validate annotation quality. This provides an additional quality control and enables analysis of the gap between ideal proactive behaviors and current reactive patterns.

\subsection{Oracle Annotation Agent}\label{oracle_annotation_agent}

To facilitate faster annotation, the oracle annotation agent is initialized using a YAML configuration with flexible, easily modifiable parameters, and the annotation workflow can be parallelized according to available system resources.

\paragraph{Agent Initialization}\label{oracle_annotation_agent_yaml_config} Figure~\ref{fig:agent_config_yaml} shows a typical YAML configuration for the oracle annotation agent, enabling configurable LLM backends, tool catalog integration, and annotation customization. The agent is instantiated using the specified backend API (e.g., OpenAI or Novita).

\begin{figure}[htbp!]
\begin{smallermdframed}

\text{\# LLM Settings}\par
\text{llm:}\par
\text{\ model: "gpt-4.1-mini"}\par
\text{\ temperature: 0.1}\par
\text{\ max\_tokens: 4000}\par
\par

\text{\# Tool Catalog Settings}\par
\text{tool\_catalog:}\par
\text{\ use\_common\_tools: true}\par
\text{\ custom\_tools: \{\}}\par
\par

\text{\# Annotation Settings}\par
\text{annotation:}\par
\text{\ batch\_size: 5}\par
\text{\ validate\_output: true}\par
\text{\ retry\_on\_failure: true}\par
\text{\ max\_retries: 3}\par
\par

\text{\# Output Settings}\par
\text{output:}\par
\text{\ output\_suffix: "\_annotated"}\par
\text{\ pretty\_print: true}\par
\text{\ save\_individual\_turns: false}\par
\par

\text{\# Prompt Settings}\par
\text{prompts:}\par
\text{\ system\_prompt: |}\par
\text{\ \ You are an expert AI assistant specialized in analyzing dialogues to}\par
\text{\ \ identify proactive automation opportunities.}\par
\text{\ \ }\par
\text{\ \ Your task is to annotate dialogue turns to identify when proactive actions}\par
\text{\ \ could be beneficial in multi-party conversations.}\par
\text{\ \ For each dialogue turn, you will identify:}\par
\text{\ \ 1. Action opportunities that could be triggered}\par
\text{\ \ 2. Required and optional inputs for each action}\par
\text{\ \ 3. Readiness maturity of participants}\par
\text{\ \ 4. Confidence in triggering the action}\par
\text{\ \ 5. Current status of the action trigger}\par\par
\text{\ \ Always provide output in valid JSON format following the}\par
\text{\ \ specified schema exactly.}\par
\text{\ task\_prompt: |}\par
\text{\ \ \#\# TASK}\par
\text{\ \ Analyze the following dialogue and annotate turn \{turn\_number\}}\par
\par
\text{\ \ \#\# FULL DIALOGUE CONTEXT}\par
\text{\ \ \{dialogue\_context\}}\par
\par
\text{\ \ \#\# CURRENT TURN TO ANNOTATE}\par
\text{\ \ Turn \{turn\_number\}: \{current\_speaker\} says: "\{current\_text\}"}\par
\par
\text{\ \ \#\# AVAILABLE ACTION OPPORTUNITIES}\par
\text{\ \ \{tool\_catalog\}}\par
\text{\ \ \#\# ANNOTATION SCHEMA}\par
\text{\ \ \{\{}\par
\text{\ \ \ \ "dialogue\_turn": \{turn\_number\},}\par
\text{\ \ \ \ "speaker": "\{current\_speaker\}",}\par
\text{\ \ \ \ "text": "\{current\_text\}",}\par
\text{\ \ \ \ "proactive\_annotations": [\{\{}\par
\text{\ \ \ \ \ "action\_opportunity": \{\{}\par
\text{\ \ \ \ \ "name": "ActionName",}\par
\text{\ \ \ \ \ "description": "What this action does",}\par
\text{\ \ \ \ \ "inputs": \{\{}\par
\text{\ \ \ \ \ \ \ "required": [ \{\{ }\par
\text{\ \ \ \ \ \ \ \ \ "input\_name": "InputName",}\par
\text{\ \ \ \ \ \ \ \ \ "provided": true/false,}\par
\text{\ \ \ \ \ \ \ \ \ "value": "The value of the input if provided",}\par
\text{\ \ \ \ \ \ \ \}\}], }\par
\text{\ \ \ \ \ \ \ "optional": [ \{\{ }\par
\text{\ \ \ \ \ \ \ \ \ "input\_name": "OptionalInput",}\par
\text{\ \ \ \ \ \ \ \ \ "provided": true/false,}\par
\text{\ \ \ \ \ \ \ \ \ "value": "The value of the input if provided",}\par
\text{\ \ \ \ \ \ \ \}\}], }\par
\text{\ \ \ \ \ \ \ "readiness\_maturity": "low/medium/high",}\par
\text{\ \ \ \ \ \ \ "trigger\_confidence": "low/medium/high",}\par
\text{\ \ \ \ \ \ \ "action\_trigger\_status": "pending/ready\_to\_trigger/triggered/}\par
\text{\ \ \ \ \ \ \ \ \ \ repeatable/dismissed"}\par
\text{\ \ \ \ \ \ \}\}]}\par
\text{\ \ \}\} }\par
\text{\ \#\# GUIDELINES}\par
\text{\ - Only include action opportunities that are relevant to the current turn}\par
\text{\ - Consider the full dialogue context when assessing input availability}\par
\text{\ - Be conservative with "ready\_to\_trigger" status - only use when}\par
\text{\ conditions are clearly met}\par
\text{\ - If no proactive opportunities exist for this turn, return empty}\par
\text{\ "proactive\_annotations" array}\par
\text{\ - Ensure all JSON is valid and follows the schema exactly}\par
\par
\text{\ \#\# RESPONSE}\par
\text{\ Provide only the JSON response, no additional text.}\par
\par

\text{\# Logging Settings}\par
\text{logging:}\par
\text{\ \ level: "INFO"}\par
\text{\ \ log\_file: "annotation.log"}\par
\text{\ \ log\_to\_console: true}\par

\end{smallermdframed}
\caption{Oracle Agent Configuration}
\label{fig:agent_config_yaml}
\end{figure}

\paragraph{Parallel Processing} The agent supports parallel execution via the \texttt{max\_workers} parameter, which specifies the number of concurrent annotation workers (default: 4). For robust recovery and controlled execution, we provide \texttt{start\_index} to resume processing from a specified dialogue index, and \texttt{annotated\_max\_dialogues\_num} to cap the number of dialogues processed in a single run. The system also tracks progress by reporting completed dialogues with timing statistics and stores outputs using gzip compression to efficiently handle large-scale datasets.

\paragraph{Reference Ready Action Range} \label{reference_ready_action_range} In addition to annotated action opportunities, we record the \emph{reference ready action range} for each dialogue in the annotation data. For action $a \in \mathcal{A}$, where $\mathcal{A}$ is the unified action catalog, dialogue $D[i]$ from dialogue dataset $D, i \in [1, |D|]$ and $T$ is the maximum dialogue turn number in $D[i]$, we define:
\begin{itemize}
\item $\mathcal{R}_{a,D[i]}^{s_t} = \{(t, s_t) : t \in [1, T], s_t \in \text{Status Set}\}$ - all occurrences with status $s_t$ for dialogue $D[i]$
\item $\mathcal{R}_{a,D[i]}^{\text{ready}} = \{t : (t, s_t) \in \mathcal{R}_{a,D[i]}^{\text{all}}, s_t \text{ is ready status}\}$ - ready windows for dialogue $D[i]$
\end{itemize}

The resulting reference-ready action range $\mathcal{R}_{a,D[i]}^{\text{ready}}$ is recorded per dialogue and used to assess annotation quality and evaluate proactive timing for both baseline and RL-trained models.

\subsection{Validation of Ready Action Alignment on Datasets with Action Observations} \label{alignment_validation_on_with_reference_dataset}

To ensure action consistency between the output reference action dataset and \textbf{observed triggered actions}, we validate the alignment between the reference ready action ranges and the observed triggered actions. Dialogues that exhibit misalignment are filtered out, resulting in a more consistent annotation dataset.

\subsubsection{Alignment Validation Formalization}\label{alignment_validation}

Given \textbf{the action observations set} $G$ with observed triggers and \textbf{the reference ready action range} $\mathcal{R}_{a,D}^{\text{ready}}$ in corresponding oracle annotation dataset $D$, we try to measure the consistency level between observed triggered actions in $G[i,t]$ and ready action ranges $\mathcal{R}_{a,D[i]}^{\text{ready}}$, $i \in [1, |D|]$ at the dialogue turn $t, t \in [1, |D[i]|]$, then aggregate at dialogue level. For each observed triggered action $g \in G[i,t]$, we calculate its \textbf{Early ready criteria score (EC)} by checking whether a corresponding reference ready action of the same type becomes ready sufficiently early---i.e., no later than the early turn threshold parameter $\sigma$ before the triggering turn $t$:

\begin{equation}
\text{EC}(g)=
\begin{cases}
1.0, & \text{if } a^{\star}.\text{start} + \sigma \le t,\\
0.0, & \text{otherwise},
\end{cases}
\end{equation}
\begin{equation}
a^{\star}
=
\argmin_{\substack{
a \in \mathcal{R}^{\text{ready}}_{a,D[i]} \\
a.\text{name} = g.\text{name}
}}
a.\text{start},
\end{equation}
\noindent where $a.start$ is the first turn at which $a$ becomes ready.

Aggregating over the entire observed action set $G$, we define \textbf{the dataset-level Early-Ready Criterion score} \text{EC}(G) as:

\begin{equation}
\text{EC}(G) = \frac{1}{|G|} \sum_{g \in G} \text{EC}(g)
\end{equation}

\subsubsection{Alignment Validation on ABCD+}\label{alignment_validation_on_abcd}

To examine consistency changes in ABCD-derived annotations, we vary $\sigma \in \{0, 1, 2, 3, 4\} $ and report the distribution of actions with EC=1.0 and dialogue consistency scores above the threshold value (empirically, set by 0.8), as shown in Table~\ref{tab:early_turn_threshold_analysis}.

\begin{table}[t!]
\centering
\setlength{\tabcolsep}{6pt}
\resizebox{\columnwidth}{!}{
\begin{tabular}{c|c|c|c|c}
\hline\hline
\textbf{Threshold} & \textbf{Number of Actions} & \textbf{Percentage of Actions} & \textbf{Dataset Score} & \textbf{Dataset Score} \\
\textbf{$\sigma$}   & \textbf{with EC(g) = 1.0}     & \textbf{with EC(g) = 1.0}         &  \text{EC}(G)                       & \textbf{\text{EC}(G) $>$ 0.8}        \\ \hline
4                   & 3,477                 & 11.78\%                        & $0.1217 \pm 0.2443$     & 4.07\%                  \\
3                   & 4,982                 & 16.88\%                        & $0.1837 \pm 0.2883$     & 6.62\%                  \\
2                   & 7,424                 & 25.16\%                        & $0.2850 \pm 0.3326$     & 11.42\%                 \\
1                   & 12,602                & 42.71\%                        & $0.4682 \pm 0.3499$     & 21.10\%                 \\
0                   & 25,715                & 87.14\%                        & $0.8877 \pm 0.1926$     & 70.27\%                 \\ \hline
\end{tabular}
}
\caption{Impact of early turn threshold on action consistency (EC), and dialogue-level performance metrics.}
\label{tab:early_turn_threshold_analysis}
\end{table}

This observation aligns with our intuition that, as the early turn threshold $\sigma$ increases, the number of valid dialogues decreases gradually. When setting early\_turn\_threshold = 0—meaning that the reference ready action range covers just the observed triggered action at its exact turn—only 70.27\% of the annotated dialogues remain valid. We consider these dialogues as consistent and retain them for baseline validation and training purposes, resulting in an ABCD+ split of 5,647 training dialogues, 703 validation dialogues, and 692 test dialogues from the original set of 10,042 annotated dialogues.

\subsubsection{The necessity of annotation agent with future understanding}\label{comparison_to_annotation_without_future_turn}

To answer the question "how well the future-viewing annotation agent performs compared to those without knowing the future turns", we refine our annotation pipeline and include the -–without-future option to make \textbf{a strictly causal oracle annotator} that generates 10,042 annotated dialogues \textbf{without knowing the future turns on the ABCD dataset}.

We analyze \textbf{the following performance aspects} to measure the annotation performance for these two settings:

\begin{enumerate}
    \item \textbf{Overall coverage}: Percentage of triggered actions whose oracle annotation satisfies both early-ready and transition criteria at the most lenient threshold (threshold=0).
    \item \textbf{Oracle annotation coverage}: Percentage of triggered actions that have any oracle annotation at all (i.e., the oracle recognized the action exists).
    \item \textbf{Score consistency}: Standard deviation of per-dialogue quality scores — lower means more uniform annotation quality across dialogues.
    \item \textbf{Turn gap precision}: Average number of turns between the oracle's earliest ready prediction and the actual trigger — smaller means more precise timing.
    \item \textbf{Phantom noise rate}: Percentage of oracle-annotated action ranges that refer to actions the assistant never actually triggered — pure noise in training data.
    \item \textbf{Critical action miss rate}: Miss rate for pull-up-account, the most frequent foundational action — a proxy for how well the oracle captures essential workflow steps.
\end{enumerate}

As reflected by Table~\ref{tab:future_turn_effect}, the annotation agent with future-viewing outperforms the setting skipping the future turn information.

\begin{table}[t!]
\centering
\setlength{\tabcolsep}{6pt}
\resizebox{\columnwidth}{!}{
\begin{tabular}{lcc}
\toprule
\textbf{Metric} & \textbf{With Future} & \textbf{Without Future} \\
\midrule
Overall coverage & 87.14\% & 78.61\% \\
Oracle annotation coverage & 80.89\% & 68.76\% \\
Score consistency & $\pm 0.19$ & $\pm 0.26$ \\
Turn gap precision & mean 2.41 & mean 3.75 \\
Phantom noise rate & 47.18\% & 58.10\% \\
Critical action miss rate & 1.3\% & 45.5\% \\
\bottomrule
\end{tabular}
}
\caption{Effect of future-turn information on evaluation quality.}
\label{tab:future_turn_effect}
\end{table}

We further \textbf{quantify the impact on training data quality} in Table~\ref{tab:future_turn_impact}. Removing future-turn information reduces the number of high-quality dialogues by 18.6\%, introduces 7,958 additional noisy labels, and eliminates 4,324 correctly matched annotations. In other words, excluding future context imposes a triple penalty: fewer usable training samples, more annotation noise, and fewer correct labels, without any compensating improvement in data quality. These results suggest that future-turn information is critical for reliable annotation alignment. They also highlight the importance of incorporating targeted human oversight to further improve the annotation pipeline and mitigate residual labeling errors.

\begin{table}[t]
\centering
\setlength{\tabcolsep}{4pt}
\small
\resizebox{\columnwidth}{!}{
\begin{tabular}{lccc}
\toprule
\textbf{Metric} & \textbf{Future} & \textbf{No Future} & \textbf{Change} \\
\midrule
High-quality dialogues ($>$0.8) & 7,058 (70.3\%) & 5,746 (58.1\%) & $-18.6\%$ \\
Missing annotations & 6,973 (19.1\%) & 11,397 (31.2\%) & +63.4\% \\
Phantom ranges & 25,274 & 33,232 & +31.5\% \\
Correctly-matched ranges & 28,292 & 23,968 & $-15.3\%$ \\
\bottomrule
\end{tabular}
}
\caption{Impact of removing future-turn information on annotation quality.}
\label{tab:future_turn_impact}
\end{table}

\section{Baseline}\label{sec:baseline_details}

\begin{figure}[t!]
    \centering
    \includegraphics[width=\columnwidth]{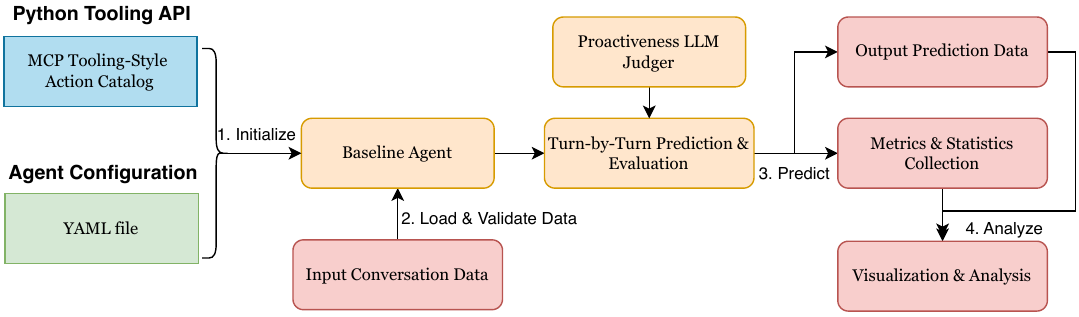}
    \caption{Training-free baseline Workflow: We initialize each training-free baseline agent using the unified action catalog and its configuration. The agent then operates in a turn-by-turn manner, generating predictions while collecting performance statistics. Beyond predefined metrics, an LLM-based judger evaluates proactive behavior, and visualization tools support further analysis.}
    \label{fig:8}
\end{figure}

This section presents the design of each training-free baseline, the YAML configurations provided, the workflow for collecting runtime metrics, and the visualization tools developed to diagnose potential issues in predicted action opportunities. The overall training-free workflow is illustrated in Figure~\ref{fig:8}. Meanwhile, we also introduce two SFT variants, LoRA-based and full parameter-tuning, using the ArcticTraining framework~\cite{bekman2025arctic} to validate the necessity of RL adaptation.

\subsection{Design Rationale}

\paragraph{Prompting Strategy Design.}
Our three prompting strategies progressively externalize reasoning and memory. Non-Reasoning relies solely on internal model knowledge, Reasoning makes deliberation explicit prior to action prediction, and ASG further maintains temporal state across turns. Each strategy isolates a specific capability—contextual understanding, deliberative reasoning, and temporal state awareness—allowing us to analyze which components of proactive behavior benefit from explicit prompting versus learning-based optimization. These observations motivate our RL approach, which learns to balance such trade-offs dynamically.

\paragraph{SFT Comparison.}
We also include supervised fine-tuning (SFT) as a baseline for Qwen models. However, from the design perspective, our application domain requires agents to learn flexible patterns from conversations rather than replicate exact behaviors from examples. The space of possible conversation patterns and action timings is too difficult to be fully covered by supervised examples. To validate this assumption about RL advantages over SFT, we also set up the SFT baseline on the ABCD+ training set through two different settings: LoRA tuning and full parameter tuning on the corresponding RL-trained model.

We evaluate these baselines using GPT-4.1-mini, GPT-5.1, Gemini-2.5-flash, Claude-sonnet-4-20250514, and Qwen2.5-14B-Instruct. We also adopt Qwen2.5-14B-Instruct as our baseline choices for SFT and training-free settings, as its 4-bit quantized variant is later fine-tuned with LoRA to obtain ProActor-Q4, enabling a direct and controlled comparison of performance gains.

\begin{figure}[t!]
    \centering
    \includegraphics[width=\columnwidth]{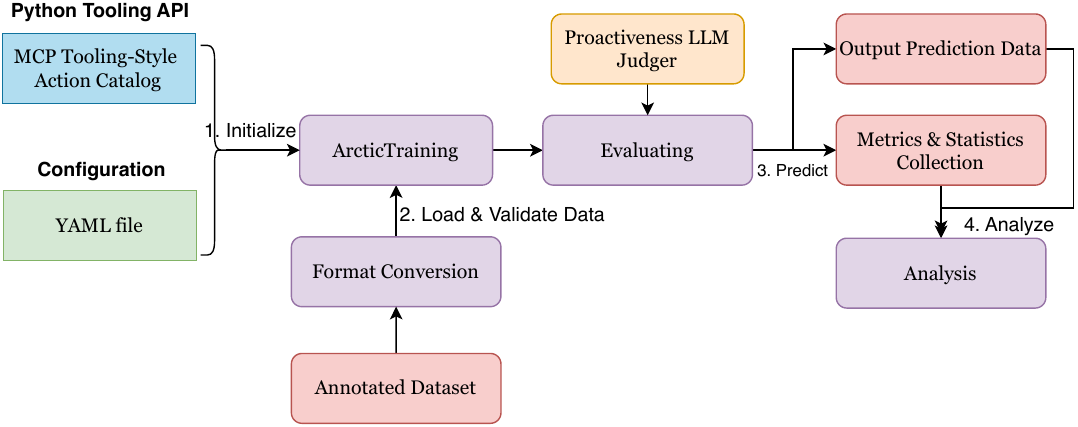}
    \caption{SFT baseline Workflow: We integrate the ArcticTraining Framework into the baseline workflow by customizing data format, plugging the evaluation at the fixed steps, and simplifying the analysis step, which is highlighted in purple color in the diagram.}
    \label{fig:9}
\end{figure}

\subsection{Non-Reasoning \& Reasoning Baseline}\label{baseline_concept}

\paragraph{Direct Prompting: Non-Reasoning Agent}
\label{direct_prompting_agent}
Our first baseline tests whether LLMs can naturally identify action opportunities, capture action dependencies, and judge triggering timing without explicit reasoning. The agent predicts actions and their parameters based solely on the current conversation, and we also prompt it to raise clarifying questions when needed. This baseline establishes the lower bound of what can be achieved through contextual understanding.

\paragraph{Reasoning Agent}
\label{reasoning_agent}
On top of direct prompting, this baseline introduces explicit proactive reasoning to improve decision-making. We prompt the model to first output a \texttt{<think>} section where it analyzes the current conversation context, identifies potential action opportunities, and reasons about appropriate triggering timing. Only after this reasoning step is completed, it outputs an \texttt{<action>} section with specific actions and parameters, and triggering readiness. This tests whether forcing models to ``think about specific aspects'' before acting improves their proactive behavior.

\subsection{ASG+Reasoning Baseline Detail} \label{asg_reasoning_details}

\begin{figure*}[t!]
  \centering
  \begin{minipage}[t]{0.95\columnwidth}
    \centering
    \includegraphics[width=\linewidth]{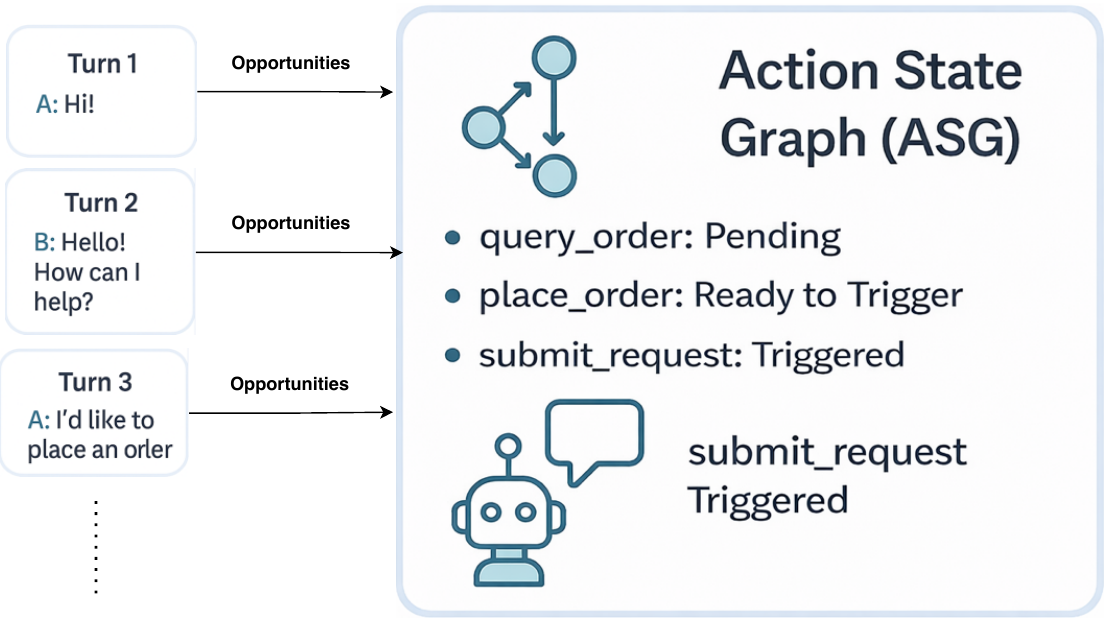}
    \caption*{\small (a) ASG overview: The Action State Graph (ASG) maintains predicted action opportunities and tracks their status transitions in application memory. The updated state is exposed to the baseline model by injecting a structured JSON representation into the prompt.}
  \end{minipage}\hfill
  \begin{minipage}[t]{\columnwidth}
    \centering
    \includegraphics[width=\linewidth]{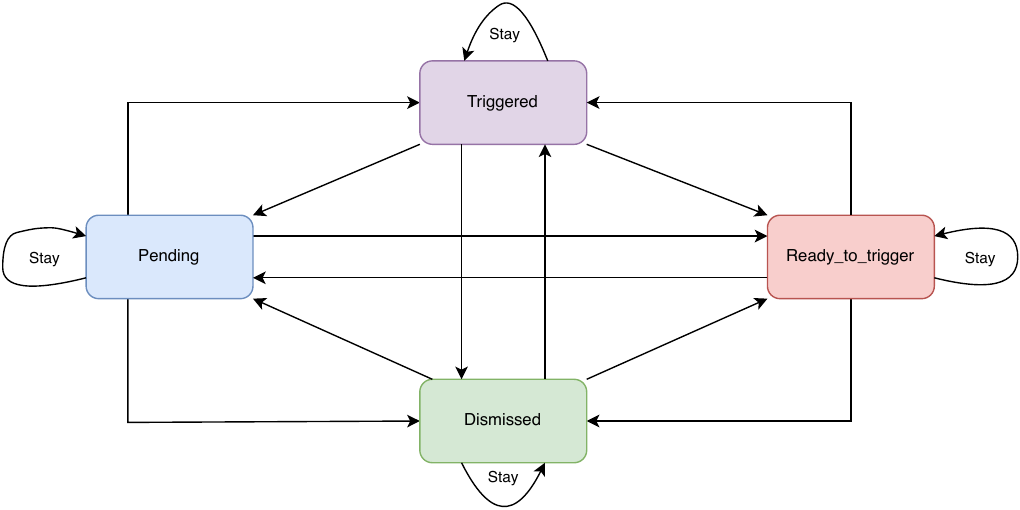}
    \caption*{\small (b) Action State Update Heuristics: For each action opportunity, we design state-transition heuristics that specify the set of permissible target statuses given the current status. In ABCD+ and Home Loan, the resulting transition graph is fully connected, allowing any action status to transition to any other status.}
  \end{minipage}
  \caption{Action State Graph (ASG). The left panel illustrates the ASG overview, while the right panel presents the heuristic update rules governing action state evolution.}
  \label{fig:asg_combined}
\end{figure*}

In this section, we focus on illustrating the details of Action State Graph (ASG) and Action State Update Heuristics shown in Figure~\ref{fig:asg_combined}.

In the ASG+Reasoning agent, the action trigger status follows a fully connected graph where any valid state can transition to any other valid state, representing a complete directed graph with self-loops allowed: 

\begin{equation}
\begin{aligned}
\mathcal{S}
&= \{\smallsc{PENDING}, \smallsc{READY\_TO\_TRIGGER}, \\
&\quad \smallsc{TRIGGERED}, \smallsc{REPEATABLE}, \smallsc{DISMISSED}\}, \\
&\forall\, s_i, s_j \in \mathcal{S}, s_i \rightarrow s_j .
\end{aligned}
\end{equation}

\paragraph{In-Memory Action State} We use the \textsc{UpdateActionStates} procedure to update the action state in the agent's runtime memory within the context of a single dialogue. For Non-Triggered action opportunities,  their state transitions must follow the defined rules. In comparison, triggered actions directly overwrite any existing state without validation. More details, please refer to Algorithm~\ref{alg:update_action_states}.

\begin{algorithm}[t]
\caption{Update Action States}
\label{alg:update_action_states}
\small
\begin{algorithmic}[1]
\Procedure{UpdateActionStates}{$new\_actions$, $is\_triggered\_action$}
    \State $actions \leftarrow$ current proactive action opportunities list

    \ForAll{$new\_action \in new\_actions$}
        \State $existing\_action \leftarrow \textsc{None}$
        \State $existing\_index \leftarrow -1$

        \Comment{Find if action already exists by name}
        \For{$i = 0$ \textbf{to} $|actions| - 1$}
            \If{$actions[i].name = new\_action.name$}
                \State $existing\_action \leftarrow actions[i]$
                \State $existing\_index \leftarrow i$
                \State \textbf{break}
            \EndIf
        \EndFor

        \If{$is\_triggered\_action = \textsc{True}$}
            \Comment{Case 1: Triggered actions - overwrite or append}
            \If{$existing\_action \neq \textsc{None}$}
                \State $actions[existing\_index] \leftarrow new\_action$
            \Else
                \State $actions.\textsc{append}(new\_action)$
            \EndIf
        \Else
            \Comment{Case 2 \& 3: Predicted actions}
            \If{$existing\_action \neq \textsc{None}$}
                \State $current\_status \leftarrow existing\_action.action\_trigger\_status$
                \State $new\_status \leftarrow new\_action.action\_trigger\_status$

                \If{\Call{IsValidTransition}{$current\_status$, $new\_status$}}
                    \State $actions[existing\_index] \leftarrow new\_action$
                \EndIf
                \Comment{If invalid transition, reject update (do nothing)}
            \Else
                \Comment{Case 3: New predicted action - append it}
                \State $actions.\textsc{append}(new\_action)$
            \EndIf
        \EndIf
    \EndFor
\EndProcedure
\end{algorithmic}
\end{algorithm}

\begin{algorithm}[t!]
\caption{Validate State Transition}
\label{alg:is_valid_transition}
\small
\begin{algorithmic}[1]
\Function{IsValidTransition}{$current\_status$, $new\_status$}
    \State $valid\_states \gets$
    \Statex \hspace{\algorithmicindent}
    $\{\smallsc{PENDING}, \smallsc{READY\_TO\_TRIGGER},$
    \Statex \hspace{\algorithmicindent}
    $\smallsc{TRIGGERED}, \smallsc{REPEATABLE}, \textsc{DISMISSED}\}$

    \If{$current\_status \notin valid\_states$ \textbf{or} $new\_status \notin valid\_states$}
        \State \Return \textsc{False}
    \EndIf
    \State \Return \textsc{True}
\EndFunction
\end{algorithmic}
\end{algorithm}

\paragraph{Action State Representation In Prompt} When the ASG+Reasoning agent predicts action opportunities or raises questions, the in-memory action state \texttt{actions} (Algorithm~\ref{alg:update_action_states}) is converted into a structured JSON representation (Figure~\ref{fig:json_representation_of_action_states}) and injected into the action- and question-related prompts by replacing the \texttt{action\_states} variable.

\begin{figure}[t!]
\begin{smallermdframed}

\text{[}\par
\text{\ \{}\par
\text{\ \ \ "action\_opportunity": \{}\par
\text{\ \ \ \ \ "name": "ActionName",}\par
\text{\ \ \ \ \ "description": "What this action does",}\par
\text{\ \ \ \ \ "inputs": \{}\par
\text{\ \ \ \ \ \ \ "required": [}\par
\text{\ \ \ \ \ \ \ \ \ \{}\par
\text{\ \ \ \ \ \ \ \ \ \ \ "input\_name": "InputName",}\par
\text{\ \ \ \ \ \ \ \ \ \ \ "provided": true,}\par
\text{\ \ \ \ \ \ \ \ \ \ \ "value": "The value of the input if provided"}\par
\text{\ \ \ \ \ \ \ \ \ \},}\par
\text{\ \ \ \ \ \ \ ],}\par
\text{\ \ \ \ \ \ \ "optional": [}\par
\text{\ \ \ \ \ \ \ \ \ \{}\par
\text{\ \ \ \ \ \ \ \ \ \ \ "input\_name": "InputName",}\par
\text{\ \ \ \ \ \ \ \ \ \ \ "provided": false,}\par
\text{\ \ \ \ \ \ \ \ \ \ \ "value": "The value of the input if provided"}\par
\text{\ \ \ \ \ \ \ \ \ \}}\par
\text{\ \ \ \ \ \ \ ]}\par
\text{\ \ \ \ \ \},} \par
\text{\ \ \ "parameters\_readiness\_maturity": "high",}\par
\text{\ \ \ "trigger\_confidence": "high",}\par
\text{\ \ \ "action\_trigger\_status": "ready\_to\_trigger"}\par
\text{\ \},}\par
\text{\ \ \ldots\ \textit{More Actions}\ \ldots}\par
\text{]}\par

\end{smallermdframed}
\caption{JSON representation of action states used in the Reasoning+ASG prompt to replace the \texttt{action\_states} variable.}
\label{fig:json_representation_of_action_states}
\end{figure}

\subsection{Agent Configuration}

Similar to the oracle annotation agent configuration (Figure~\ref{oracle_annotation_agent_yaml_config}), we provide a YAML configuration file for baseline initialization. This file specifies the LLM backend, tool catalog settings, and the prompts used for action opportunity prediction and question generation. Non-reasoning~\ref{direct_prompting_agent}, reasoning~\ref{reasoning_agent}, and ASG+reasoning~\ref{asg_reasoning_details} baselines all use the shared \texttt{dialogue\_context} and \texttt{tool\_catalog} variables to dynamically inject the current dialogue context and the unified tool catalog definition in Python. The only difference is that the ASG+reasoning agent needs to use \texttt{action\_states} to maintain the latest action states.

\paragraph{Baseline Configurations} Baseline YAML files support model switching via \texttt{llm.model} and allow configuration of non-OpenAI APIs through \texttt{llm.api\_key\_name} and \texttt{llm.base\_url}. Sample configurations are given below:
\begin{enumerate}
    \item YAML configuration of Direct Prompting agent baseline, refer to Figure~\ref{fig:baseline1_config1_non_cot}
    \item YAML configuration for Reasoning agent baseline, refer to Figure~\ref{fig:baseline1_config1_cot}
    \item YAML configuration for ASG+reasoning agent baseline, refer to Figure~\ref{fig:baseline1_config1_asg}
\end{enumerate}

\begin{figure}[t!]
\begin{smallermdframed}

\text{\# Baseline Configuration File}\par
\text{\# Configuration for LLM-based actor model}\par
\par

\text{\# LLM Settings}\par
\text{llm:}\par
\text{\ \ model: "openai/gpt-4.1-mini"}\par
\text{\ \ temperature: 0.1}\par
\text{\ \ max\_tokens: 4096}\par
\text{\ \ using\_short\_model\_name: true}\par
\par

\text{\# Tool Catalog Settings}\par
\text{tool\_catalog:}\par
\text{\ \ use\_common\_tools: true}\par
\text{\ \ custom\_tools: \{\}}\par
\par

\text{\# Prompt Settings: Action Prompts and Question Prompts}\par
\text{\# 1. Action Prompts}\par
\text{action\_prompts:}\par
\text{\ \ system\_prompt: |}\par
\text{\ \ \ \ You are an expert AI assistant for professional, specializing in}\par
\text{\ \ \ \ analyzing dialogues to identify automation opportunities and their}\par
\text{\ \ \ \ parameters based on the client (e.g., customer) and professional}\par
\text{\ \ \ \ (e.g., sales agent) conversations.}\par
\text{\ \ \ \ }\par
\text{\ \ \ \ Given the conversation history and available system action}\par
\text{\ \ \ \ opportunities, your task is to analyze the latest turn and determine}\par
\text{\ \ \ \ the following for all relevant actions:}\par
\text{\ \ \ \ 1. Action opportunities that could be triggered once ready}\par
\text{\ \ \ \ 2. Required and optional inputs for each action}\par
\text{\ \ \ \ 3. Parameter Readiness (readiness\_maturity)}\par
\text{\ \ \ \ 4. Confidence in triggering the action (trigger\_confidence)}\par
\text{\ \ \ \ 5. Current status of the action trigger (action\_trigger\_status)}\par
\text{\ \ \ \ }\par
\text{\ \ \ \ \#\# AVAILABLE ACTION OPPORTUNITIES}\par
\text{\ \ \ \ <tools>}\par
\text{\ \ \ \ \{tool\_catalog\}}\par
\text{\ \ \ \ </tools>}\par
\text{\ \ \ \ }\par
\text{\ \ \ \ \#\# RESPONSE SCHEMA}\par
\text{\ \ \ \ \{\{}\par
\text{\ \ \ \ \ \ "proactive\_action\_opportunities": [...]}\par
\text{\ \ \ \ \}\}}\par
\par

\text{\ \ task\_prompt: |}\par
\text{\ \ \ \ \#\# TASK}\par
\text{\ \ \ \ Analyze the following conversation for proactive automation}\par
\text{\ \ \ \ opportunities. Provide only the JSON response, no additional text.}\par
\text{\ \ \ \ }\par
\text{\ \ \ \ \#\# DIALOGUE CONTEXT}\par
\text{\ \ \ \ \{dialogue\_context\}}\par
\par

\text{\# 2. Question Prompts}\par
\text{question\_prompts:}\par
\text{\ \ system\_prompt: |}\par
\text{\ \ \ \ You are an expert AI assistant specialized in analyzing the}\par
\text{\ \ \ \ conversation between customer and sales agent to predict what}\par
\text{\ \ \ \ question the sales agent is asking.}\par
\text{\ \ \ \ }\par
\text{\ \ \ \ \#\# RESPONSE SCHEMA}\par
\text{\ \ \ \ \{\{}\par
\text{\ \ \ \ \ \ "raised\_questions": [...]}\par
\text{\ \ \ \ \}\}}\par
\par

\text{\ \ task\_prompt: |}\par
\text{\ \ \ \ \#\# TASK}\par
\text{\ \ \ \ Analyze the following dialogue and predict the question that the}\par
\text{\ \ \ \ sales agent is asking to gather the missing information for the}\par
\text{\ \ \ \ non-ready action opportunities.}\par
\text{\ \ \ \ }\par
\text{\ \ \ \ \#\# DIALOGUE CONTEXT}\par
\text{\ \ \ \ \{dialogue\_context\}}\par

\end{smallermdframed}
\caption{YAML Configuration of Direct Prompting Baseline Agent}
\label{fig:baseline1_config1_non_cot}
\end{figure}

\begin{figure}[t!]
\begin{smallermdframed}

\text{\# Baseline Configuration File}\par
\text{\# Configuration for LLM-based actor model}\par
\par

\text{\# LLM Settings}\par
\text{llm:}\par
\text{\ \ model: "openai/gpt-4.1-mini"}\par
\text{\ \ temperature: 0.1}\par
\text{\ \ max\_tokens: 4096}\par
\text{\ \ using\_short\_model\_name: true}\par
\par

\text{\# Tool Catalog Settings}\par
\text{tool\_catalog:}\par
\text{\ \ use\_common\_tools: true}\par
\text{\ \ custom\_tools: \{\}}\par
\par

\text{\# Prompt Settings: Action Prompts and Question Prompts}\par
\text{\# 1. Action Prompts}\par
\text{action\_prompts:}\par
\text{\ \ system\_prompt: |}\par
\text{\ \ \ \ You are an expert AI assistant for professional, specializing in}\par
\text{\ \ \ \ analyzing dialogues to identify automation opportunities...}\par
\text{\ \ \ \ }\par
\text{\ \ \ \ \#\# RESPONSE SCHEMA}\par
\text{\ \ \ \ <think>}\par
\text{\ \ \ \ Describe your thinking about how to decide the action}\par
\text{\ \ \ \ opportunities to be triggered in the last turn.}\par
\text{\ \ \ \ </think>}\par
\text{\ \ \ \ <action>}\par
\text{\ \ \ \ \{\{}\par
\text{\ \ \ \ \ \ "proactive\_action\_opportunities": [...]}\par
\text{\ \ \ \ \}\}}\par
\text{\ \ \ \ </action>}\par
\par

\text{\ \ task\_prompt: |}\par
\text{\ \ \ \ \#\# TASK}\par
\text{\ \ \ \ Analyze the following conversation for proactive automation}\par
\text{\ \ \ \ opportunities and output automation opportunities triggered by}\par
\text{\ \ \ \ the latest turn. Output only the <think> and <action> sections,}\par
\text{\ \ \ \ with no additional text.}\par
\text{\ \ \ \ }\par
\text{\ \ \ \ \#\# DIALOGUE CONTEXT}\par
\text{\ \ \ \ \{dialogue\_context\}}\par
\par

\text{\# 2. Question Prompts}\par
\text{question\_prompts:}\par
\text{\ \ system\_prompt: |}\par
\text{\ \ \ \ You are an expert AI assistant specialized in analyzing the}\par
\text{\ \ \ \ conversation between customer and sales agent...}\par
\text{\ \ \ \ }\par
\text{\ \ \ \ \#\# RESPONSE SCHEMA}\par
\text{\ \ \ \ <think>}\par
\text{\ \ \ \ Describe your thinking about how to decide the question to be}\par
\text{\ \ \ \ raised in the last turn...}\par
\text{\ \ \ \ </think>}\par
\text{\ \ \ \ <question>}\par
\text{\ \ \ \ \{\{}\par
\text{\ \ \ \ \ \ "raised\_questions": [...]}\par
\text{\ \ \ \ \}\}}\par
\text{\ \ \ \ </question>}\par
\par

\text{\ \ task\_prompt: |}\par
\text{\ \ \ \ \#\# TASK}\par
\text{\ \ \ \ Analyze the following dialogue and predict the question that the}\par
\text{\ \ \ \ sales agent is asking... Output only the <think> and <question>}\par
\text{\ \ \ \ sections, with no additional text.}\par
\text{\ \ \ \ }\par
\text{\ \ \ \ \#\# DIALOGUE CONTEXT}\par
\text{\ \ \ \ \{dialogue\_context\}}\par

\end{smallermdframed}
\caption{YAML Configuration of Reasoning Baseline Agent: reasoning with <think> and <action>/<question> sections}
\label{fig:baseline1_config1_cot}
\end{figure}

\begin{figure}[t!]
\begin{smallermdframed}

\text{\# Baseline Configuration File}\par
\text{\# Configuration for LLM-based actor model}\par
\par

\text{\# LLM Settings}\par
\text{llm:}\par
\text{\ \ model: "openai/gpt-4.1-mini"}\par
\text{\ \ temperature: 0.1}\par
\text{\ \ max\_tokens: 4096}\par
\text{\ \ using\_short\_model\_name: true}\par
\par

\text{\# Tool Catalog Settings}\par
\text{tool\_catalog:}\par
\text{\ \ use\_common\_tools: true}\par
\text{\ \ custom\_tools: \{\}}\par
\par

\text{\# Prompt Settings: Action Prompts and Question Prompts}\par
\text{\# 1. Action Prompts}\par
\text{action\_prompts:}\par
\text{\ \ system\_prompt: |}\par
\text{\ \ \ \ You are an expert AI assistant for professional...}\par
\text{\ \ \ \ }\par
\text{\ \ \ \ \#\# TRACKING ACTION STATES}\par
\text{\ \ \ \ You are also given the ACTION STATES in the input data:}\par
\text{\ \ \ \ the action opportunities predicted or triggered in the}\par
\text{\ \ \ \ previous turns. Please use it to track and decide the}\par
\text{\ \ \ \ action opportunity candidates in the latest turn.}\par
\text{\ \ \ \ }\par
\text{\ \ \ \ \#\# RESPONSE SCHEMA}\par
\text{\ \ \ \ <think>}\par
\text{\ \ \ \ Describe your thinking about how to decide the action}\par
\text{\ \ \ \ opportunities to be triggered in the last turn.}\par
\text{\ \ \ \ </think>}\par
\text{\ \ \ \ <action>}\par
\text{\ \ \ \ \{\{}\par
\text{\ \ \ \ \ \ "proactive\_action\_opportunities": [...]}\par
\text{\ \ \ \ \}\}}\par
\text{\ \ \ \ </action>}\par
\par

\text{\ \ task\_prompt: |}\par
\text{\ \ \ \ \#\# TASK}\par
\text{\ \ \ \ Analyze the following conversation for proactive automation}\par
\text{\ \ \ \ opportunities and output automation opportunities triggered}\par
\text{\ \ \ \ by the latest turn. Output only the <think> and <action>}\par
\text{\ \ \ \ sections, with no additional text.}\par
\text{\ \ \ \ }\par
\text{\ \ \ \ \#\# ACTION STATES}\par
\text{\ \ \ \ \{action\_states\}}\par
\text{\ \ \ \ }\par
\text{\ \ \ \ \#\# DIALOGUE CONTEXT}\par
\text{\ \ \ \ \{dialogue\_context\}}\par
\par

\text{\# 2. Question Prompts}\par
\text{question\_prompts:}\par
\text{\ \ system\_prompt: |}\par
\text{\ \ \ \ You are an expert AI assistant specialized in analyzing}\par
\text{\ \ \ \ the conversation between customer and sales agent...}\par
\text{\ \ \ \ }\par
\text{\ \ \ \ \#\# TRACKING ACTION STATES}\par
\text{\ \ \ \ You are also given the ACTION STATES in the input data:}\par
\text{\ \ \ \ the action opportunities predicted or triggered in the}\par
\text{\ \ \ \ previous turns. Please use it to track and decide the}\par
\text{\ \ \ \ action opportunity candidates in the latest turn.}\par
\par

\text{\ \ task\_prompt: |}\par
\text{\ \ \ \ \#\# TASK}\par
\text{\ \ \ \ Analyze the following dialogue and predict the question...}\par
\text{\ \ \ \ Output only the <think> and <question> sections,}\par
\text{\ \ \ \ with no additional text.}\par
\text{\ \ \ \ }\par
\text{\ \ \ \ \#\# ACTION STATES}\par
\text{\ \ \ \ \{action\_states\}}\par
\text{\ \ \ \ }\par
\text{\ \ \ \ \#\# DIALOGUE CONTEXT}\par
\text{\ \ \ \ \{dialogue\_context\}}\par

\end{smallermdframed}
\caption{YAML Configuration of Reasoning Baseline Agent: reasoning with <think> with Action State Graph support}
\label{fig:baseline1_config1_asg}
\end{figure}

Due to page length, we omit the schema structures of action opportunities and raised questions, which are similar to action opportunities specification in YAML file~\ref{fig:agent_config_yaml}.

Besides, a proactiveness LLM judger can be configured via the \texttt{llm\_judger\_config\_file} command-line argument, which specifies the corresponding YAML file. An example is shown in Figure~\ref{fig:llm_judger_config_zillow}.

\begin{figure}[t!]
\begin{smallermdframed}

\text{\# LLM Judger Configuration File for ASG Support}\par
\text{judgers:}\par
\text{\ \ - name: "proactiveness\_evaluator"}\par
\text{\ \ \ \ model: "openai/gpt-4.1-mini"}\par
\text{\ \ \ \ temperature: 0.1}\par
\text{\ \ \ \ max\_tokens: 4096}\par
\text{\ \ \ \ using\_short\_model\_name: true}\par
\text{\ \ \ \ api\_key\_name: <INTERNAL\_API\_KEY>}\par
\text{\ \ \ \ base\_url: <INTERNAL\_SERVER\_BASE\_URL>}\par
\text{\ \ \ \ settings:}\par
\text{\ \ \ \ \ \ max\_retries: 3}\par
\text{\ \ \ \ \ \ retry\_on\_failure: true}\par
\par

\text{\ \ \ \ system\_prompt: |}\par
\text{\ \ \ \ \ \ You are a helpful assistant tasked with judging whether a}\par
\text{\ \ \ \ \ \ proactive action prediction is truly proactive — meaning it was}\par
\text{\ \ \ \ \ \ not explicitly requested by the customer but still aligns with}\par
\text{\ \ \ \ \ \ their intent.}\par
\text{\ \ \ \ \ \ }\par
\text{\ \ \ \ \ \ You are given:}\par
\text{\ \ \ \ \ \ - PROACTIVE ACTION PREDICTIONS AT CURRENT TURN}\par
\text{\ \ \ \ \ \ - DIALOGUE UNTIL CURRENT TURN}\par
\text{\ \ \ \ \ \ }\par
\text{\ \ \ \ \ \ Your task is to:}\par
\text{\ \ \ \ \ \ 1. Determine if the customer explicitly requested the predicted action.}\par
\text{\ \ \ \ \ \ 2. Assess the proactiveness level — how well the prediction matches}\par
\text{\ \ \ \ \ \ \ \ \ the customer's intent without a direct request.}\par
\par

\text{\ \ \ \ \ \ \# RATING CRITERIA}\par
\text{\ \ \ \ \ \ The proactiveness level is calculated based on:}\par
\text{\ \ \ \ \ \ -1 - Conversation is insufficient to judge.}\par
\text{\ \ \ \ \ \ \ 1 - Explicit or almost explicit request, target action missing.}\par
\text{\ \ \ \ \ \ \ 2 - Explicit request, target action present, inaccurate parameters.}\par
\text{\ \ \ \ \ \ \ 3 - Explicit request, target action present, accurate parameters.}\par
\text{\ \ \ \ \ \ \ 4 - Implicit request, target action present, inaccurate parameters.}\par
\text{\ \ \ \ \ \ \ 5 - Implicit request, target action present, accurate parameters.}\par
\par

\text{\ \ \ \ \ \ \# RESPONSE SCHEMA}\par
\text{\ \ \ \ \ \ Return only valid JSON without any additional text:}\par
\text{\ \ \ \ \ \ \{}\par
\text{\ \ \ \ \ \ "reasoning": "Brief explanation (<=50 words)",}\par
\text{\ \ \ \ \ \ "rating": -1/1/2/3/4/5}\par
\text{\ \ \ \ \ \ \}}\par
\par

\text{\ \ \ \ task\_prompt: |}\par
\text{\ \ \ \ \ \ \#\# TASK}\par
\text{\ \ \ \ \ \ You are given the conversation history so far and the proactive}\par
\text{\ \ \ \ \ \ action predictions from the proactive agent for the latest turn.}\par
\text{\ \ \ \ \ \ }\par
\text{\ \ \ \ \ \ \# DIALOGUE UNTIL CURRENT TURN}\par
\text{\ \ \ \ \ \ \{dialogue\_context\}}\par
\text{\ \ \ \ \ \ }\par
\text{\ \ \ \ \ \ \# PROACTIVE ACTION PREDICTIONS AT CURRENT TURN}\par
\text{\ \ \ \ \ \ \{proactive\_action\_predictions\}}\par
\text{\ \ \ \ \ \ }\par
\text{\ \ \ \ \ \ Evaluate the proactiveness of these predictions and return your}\par
\text{\ \ \ \ \ \ assessment in the specified JSON format.}\par

\end{smallermdframed}
\caption{LLM Judger Configuration: Proactiveness evaluation with 5-point rating scale}
\label{fig:llm_judger_config_zillow}
\end{figure}

\subsection{Result Analysis \& Visualization}

\begin{figure}[t!]
    \centering
    \includegraphics[width=\columnwidth]{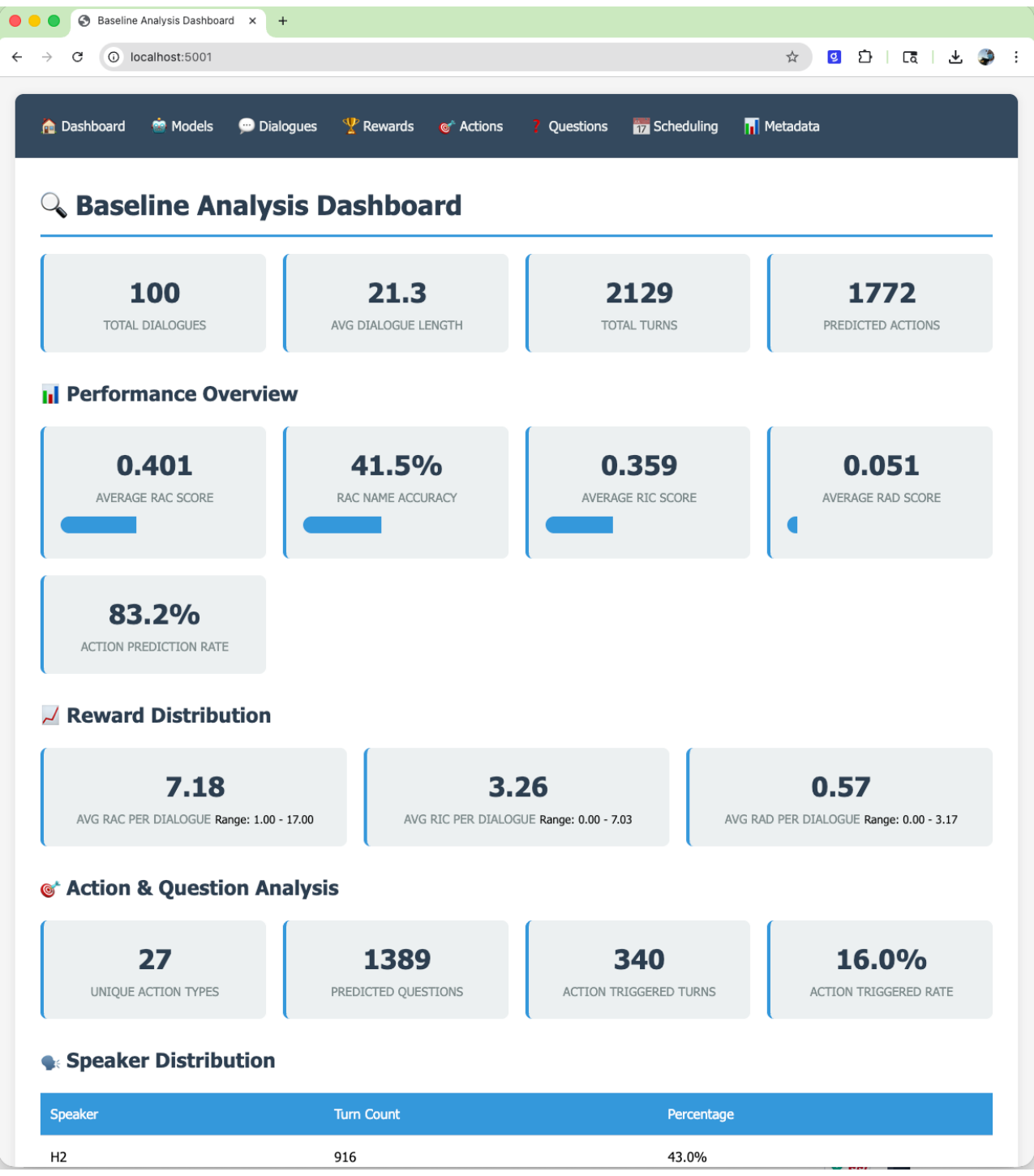}
    \caption{Web-based visualization interface for baseline performance analysis. The dashboard aggregates dialogue-level and turn-level metrics, including action prediction statistics, reward distributions, and trigger behaviors, and supports interactive inspection of individual dialogues and predicted action states.}
    \label{fig:10}
\end{figure}

Since the ABCD+ and Home Loan datasets consist of long dialogues (typically exceeding 20 turns), all performance metrics—including proactiveness metrics from the LLM-based judger—are computed online at the generation time of predicted action and raised question. All metrics and statistics are stored in JSON format for analysis. To support human inspection in nuanced cases, we additionally developed a web-based interface that loads both the aggregated metrics and corresponding prediction outputs, enabling evaluators to examine summary statistics as well as dialogue-level states and predictions (Figure~\ref{fig:10}).

\subsection{SFT Baseline}\label{sft_baseline_setup}

Considering that we apply LoRA training on the 4-bit quantized Qwen2.5-14B-Instruct model, we implement two SFT baseline settings:

\begin{enumerate}
    \item \textbf{LoRA Tuning}. 
    On 2$\times$H100 GPUs, we apply Low-Rank Adaptation (LoRA)~\cite{hu2022lora} with rank $r=16$, scaling factor $\alpha=32$, and dropout $0.05$. LoRA adapters are applied to all attention and MLP projection layers (\texttt{q\_proj}, \texttt{k\_proj}, \texttt{v\_proj}, \texttt{o\_proj}, \texttt{gate\_proj}, \texttt{up\_proj}, \texttt{down\_proj}). 
    Training uses a micro-batch size of 6, learning rate $2\times10^{-4}$, and a cosine learning-rate scheduler with 5\% warmup. We employ DeepSpeed ZeRO Stage~2 for memory optimization. The maximum sequence length is set to 8192 tokens.

    \item \textbf{Full Tuning}. 
    On 2$\times$H200 GPUs, we fine-tune all model parameters with a micro-batch size of 1 and gradient accumulation of 6 steps (effective batch size 6). We use a learning rate of $2\times10^{-5}$, weight decay $0.01$, and a cosine learning-rate scheduler with 5\% warmup. DeepSpeed ZeRO Stage~3 is applied to enable memory-efficient full-parameter training. The maximum sequence length is set to 8192 tokens.
\end{enumerate}

To integrate the proactiveness LLM judge, we extend the evaluation hook in the ArcticTraining framework so that metric computation and statistical aggregation remain consistent with the training-free baselines. Due to time constraints, we do not implement compatibility adaptations for the baseline workflow required by the visualization components in the SFT pipeline.

\section{ART-F Framework}\label{sec:art_f_framework}

This section introduces core components of ART-F, our open-source framework for end-to-end RL training, with a focus on how its inference cluster and training infrastructure maximize resource utilization under GPU constraints (Figure~\ref{fig:3}). The design and implementation of the ART-F client—which matches inference server capacity with training throughput—are detailed in Section~\ref{sec:art_f_client_implementation}.

\subsection{Inference Phase} \label{sec:inference_cluster_and_client}

Although ART’s original inference design supports only a single inference server per GPU, quantized LoRA models typically consume only a small fraction of GPU resources, leading to substantial under-utilization and limited inference throughput. To address this limitation, ART-F introduces an inference cluster composed of multiple inference server instances coordinated by a load-balancing proxy that monitors runtime status. This design improves GPU utilization and significantly increases rollout throughput. Moreover, by exposing real-time cluster status, the load balancer enables RL rollout clients to make informed scheduling decisions—such as selecting available servers or early-terminating rollouts under saturation—thereby enhancing runtime flexibility and system robustness.

\begin{figure}[t!]
    \centering
    \includegraphics[width=\columnwidth]{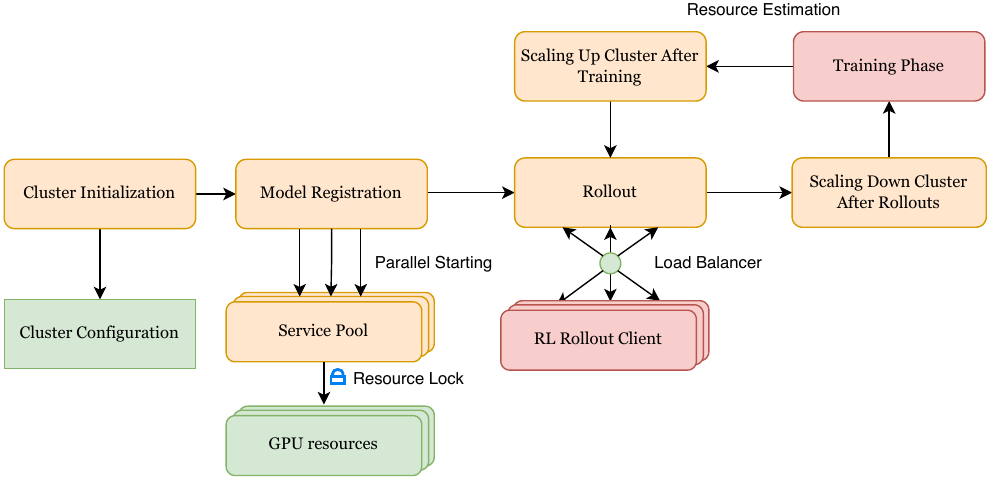}
    \caption{Inference Phase Summary: During inference, ART-F dynamically initializes and scales an inference cluster, allocates GPU resources via a service pool with resource locking, and executes parallel rollouts through the RL rollout client, before scaling down the cluster to release resources once rollouts are completed.}
    \label{fig:12}
\end{figure}

The entire lifecycle of the inference cluster is managed by the inference cluster manager. For simplification purposes, we use \textbf{a unique cluster configuration} for both inference and training, and allows the inference cluster manager and the training manager to access their separate sections. An overview of the inference phase is illustrated in Figure~\ref{fig:12}.

\paragraph{Initialization} The inference cluster manager reads the cluster configuration from the YAML configuration, which specifies the scaling policy—including the maximum number of servers per model (\texttt{max\_servers\_per\_model}), the server port allocation range (\texttt{port\_range\_start} to \texttt{port\_range\_end}, e.g., 8000–8010), and the load-balancing strategy (\texttt{load\_balancing\_strategy}, such as round-robin, random, or response-time–based). After validating it, the manager registers the model and initializes $N$ vLLM inference server instances, each reserving a fixed fraction of maximum GPU memory (\texttt{gpu\_memory\_per\_server}) on a single GPU.

To minimize startup latency while respecting Unsloth’s process-level lock~\cite{unsloth} during model loading, the $N$ vLLM instances are launched in parallel with a cross-process resource lock held throughout the loading phase. Once all inference servers successfully pass health-check (PING) tests, the inference cluster manager marks the cluster as ready and begins serving requests from RL rollout clients.

\paragraph{Parallel Rollout} After the inference cluster turns ready to handle rollout client requests, parallel requests will be distributed by the Load Balancer, which gets the next server from the server pool based on the load-balancing strategy. This Load Balancer also take charge of health monitoring \& failure recovery during request handling. It also tracks the request counters and response time metrics for each vLLM server instance.

\paragraph{Cluster Scaling} After the rollout phase, the inference cluster will be scaled down to allow all GPU resources used for training purposes. There are two different strategies for Non-DDP and DDP trainings:
\begin{enumerate}
    \item For Non-DDP training, we scale down the vLLM server instances to only keep the server with the lowest port number and pause server monitor before training so that GPU memory locked by N-1 servers can be utilized in the following training. After the training completion, we resume monitoring for the leftover server and scale up back to $N$ server and prepare for next rollout cycle.
    \item For DDP training, we shut down the inference cluster completely and let the DDP training manager start the training cycle. Once the training is completed, we restart the whole inference cluster and enter the next rollout cycle. The reason is that in ART-F's DDP training, we trigger a multi-processes collaboration to not only resume the normal DDP training, but also monitor the progress and guarantee the step synchronization, which needs higher resource demands. More training details will be covered in next section~\ref{training_art_f_phase}.
\end{enumerate}

\subsection{DDP-Based Training}
\label{training_art_f_phase}

\begin{figure}[t!]
    \centering
    \includegraphics[width=\columnwidth]{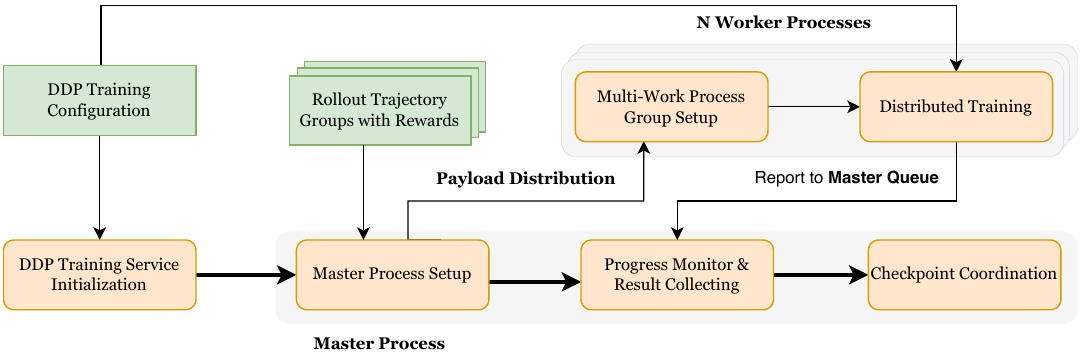}
    \caption{DDP-Based Training Phase Summary: ART-F orchestrates distributed data-parallel (DDP) training by initializing a master process and multiple worker processes, distributing training payloads—rollout trajectory groups with rewards—across workers, coordinating checkpoints, and aggregating progress and results throughout training.}
    \label{fig:13}
\end{figure}

Since non-DDP training largely follows the original ART framework with only minor adaptations, we focus this section on the DDP-based training phase implemented in ART-F. Unlike standard DDP setups that rely on \texttt{torchrun}, our training framework is \textbf{custom-built} to support advanced process coordination, fine-grained progress monitoring, and flexible data distribution. The framework is designed to fully utilize GPU resources on a single multi-GPU node. As illustrated in Figure~\ref{fig:13}, ART-F adopts a master–worker architecture in which a centrally managed controller process orchestrates multiple training workers. 

\paragraph{DDP Training Service} This service is compatible with the existing local Unsloth training service in the ART framework and is responsible for loading and validating the DDP training configuration. Specifically, it performs the following checks:
(1) verifies that the configured \texttt{world\_size} in \texttt{DDPTrainingConfig} does not exceed the number of available GPUs when \texttt{enable\_ddp} is set to \texttt{True} (2) validates the DDP backend selection (e.g., \texttt{nccl} or \texttt{gloo}) and ensures that all required process-coordination parameters—such as \texttt{master\_addr}, \texttt{port}, and \texttt{ddp\_timeout\_minutes}—are properly specified; and (3) checks that trainer arguments cloned across worker processes (e.g., \texttt{per\_device\_train\_batch\_size} and \texttt{per\_device\_eval\_batch\_size}) remain consistent, and that the effective runtime batch size is reasonable when \texttt{batch\_size\_allow\_adjusting} is enabled. If all validations pass, the service triggers the master process to initiate DDP training.

\paragraph{Master Process} \label{master_process_payload_distribution} 
This process, spawned by the DDP Training Service, serves two primary roles: 1) setting up the DDP process group and initializing $N$ worker processes; and 2) preparing a shared asynchronous queue—referred to as the Master Queue $MQ$ — through which worker processes report their execution progress. 

Specifically, we initialize the process context using \texttt{mp.get\_context("spawn")}, and use \texttt{mp.Process} to launch $N$ worker processes and create the shared $MQ$ within the same context. The master process synchronously waits for progress updates by invoking \texttt{MQ.get()}, which naturally blocks execution until feedback is received from the workers. Only after all trajectory group payloads dispatched to the worker processes have been fully processed does the master process proceed to the final synchronization point. This design guarantees that all worker processes terminate gracefully and that the DDP training phase completes in a consistent and orderly manner.

More importantly, the master process determines how the raw training payloads---rollout trajectory groups with rewards, denoted as $RG$---are distributed across worker processes according to the configuration option \texttt{replicate\_dataset\_across\_ranks}. We first align the effective payload size to the number of workers by rounding it down to $\lfloor \tfrac{RG}{N} \rfloor \cdot N$, ensuring an even partition across $N$ worker processes. 

If \texttt{replicate\_dataset\_across\_ranks} is set to \texttt{True}, ART-F performs \textbf{symmetric} DDP training, in which the aligned payload $RG$ is replicated identically across all worker processes. Otherwise, ART-F adopts \textbf{asymmetric} DDP training, where each worker process receives a disjoint subset of size $\lfloor \tfrac{RG}{N} \rfloor$. In theory, both settings are expected to converge under identical hyperparameter configurations. However, in our preliminary experiments, the heterogeneous rollout groups and the resulting gradient discrepancies in the \textbf{asymmetric} setting lead to noticeably less stable convergence compared to the \textbf{symmetric} setting. 

\label{symmetric_vs_non_symmetric_epoch}
Consequently, we adopt the \textbf{symmetric} DDP configuration with a reduced number of training epochs. For example, in our ABCD+ experiments on a single node with 4 $\times$ H200 GPUs, we set the number of epochs to 2 with \texttt{replicate\_dataset\_across\_ranks} = \texttt{True}, which yields performance comparable to training for 8 epochs with \texttt{replicate\_dataset\_across\_ranks} = \texttt{False} on the same payload $RG$.

\paragraph{Worker Processes} After being triggered by the master process, each worker process first initializes the DDP process group on its assigned GPU rank via PyTorch’s distributed interface \texttt{init\_process\_group()}, which primarily handles CUDA device assignment and validation. Given the payload allocated to each rank, workers then execute training in parallel on their assigned data partitions. 

Each worker process instantiates its own \texttt{GRPOTrainer} equipped with an asynchronous task queue. The trainer consumes data chunks from this queue and reports step-wise training progress to the shared \textbf{Master Queue} (\texttt{MQ}). To prevent synchronization issues during training, we employ \textbf{step-wise barrier synchronization to enforce batch-level gradient consistency} across ranks. Worker queue states are continuously monitored and propagated to all ranks to maintain a consistent global view of execution status. If any worker reports an error condition, the DDP training session is immediately terminated and the current model checkpoint is saved, preventing indefinite blocking or timeouts caused by stalled processes.

Once all worker processes successfully complete training, the master process enters the checkpoint coordination phase. In this phase, the rank-0 worker exclusively saves the final checkpoint to persist the model state to disk and then terminates gracefully. Afterward, the master process concludes process synchronization and releases all resources reserved by the worker processes.

\subsection{Other Features} \label{other_features}

Beyond the inference cluster and DDP-enabled training pipeline, ART-F introduces several additional features to simplify the end-to-end RL workflow:

\begin{enumerate}
    \item \textbf{Integrated experiment tracking.} Besides \texttt{wandb.ai}~\cite{wandb2025website}, ART-F provides built-in MLflow integration with extended system-level metrics, including GPU memory utilization, disk I/O, and network activity. We allow developers to extend the existing \texttt{MlflowSystemMonitor} to enables more customized monitoring.
    
    \item \textbf{Customizable RULER module.} We extend the original RULER to support user-defined evaluation rules (e.g., proactiveness assessment rubrics) and seamless integration of third-party models. Trajectory saving via W\&B Weave~\cite{wandb2025weave} is made optional to mitigate potential sensitive data leakage during training.
    
    \item \textbf{Databricks compatibility enhancements.} ART-F includes additional utilities to facilitate Databricks-based training workflows~\cite{databricks}, such as dynamic server address resolution at runtime, temporary directory management for memory-mapped files, trajectory compression, and incremental checkpoint saving.
\end{enumerate}

\subsection{ART-F Parameter Tuning} \label{art_f_parameter_tuning}

As our experiments target multiple GPUs within a single machine node, training and inference in ART-F share the same GPU and system resources. Careful tuning of training and inference parameters is therefore essential to prevent resource exhaustion and cascading “avalanche” failures, such as slow inference server startup, training deadlocks due to GPU memory locking, and frequent HTTP client timeouts. We accordingly consider two key aspects of parameter tuning in ART-F.

\paragraph{Dynamic Training Batch Size.} \label{dynamic_training_size}
When enabling dynamic training batch size, we observed that DDP training achieves approximately 5--10\% speedup per training step in our setting. This aligns with our setting: for an initial batch size $N$, the runtime batch size switches between $N$ and $N-1$. Thus, one additional batch can be processed when enabling this option, yielding approximately 5--10\% speedup, consistent with our observations. 

\paragraph{Rollout Group Size in Inference.}
This parameter significantly affects inference servers hosted on a single compute node, primarily determining how many parallel requests can be handled on that node. We observed that when setting the rollout group size to 250, rollout time per epoch step decreases initially but gradually slows as rollout continues due to the increasing context. Adjusting to 230 effectively alleviates this slowdown while maintaining a good request consumption rate.

\paragraph{Speed-up Analysis.} \label{tensor_parallelization_and_speed_up}

\begin{table}[t!]
\centering
\small
\setlength{\tabcolsep}{1.8pt}
\resizebox{\columnwidth}{!}{
\begin{tabular}{c|ccc|ccc|ccc}
\toprule
\textbf{Data Size}
& \multicolumn{3}{c|}{\textbf{ART (1$\times$H200)}}
& \multicolumn{3}{c|}{\textbf{ART-F (1$\times$H200)}}
& \multicolumn{3}{c}{\textbf{ART-F (4$\times$H200)}} \\
\cline{2-10}
& \textbf{\#Srv/\#R} & \textbf{Stat.} & \textbf{Time}
& \textbf{\#Srv/\#R} & \textbf{Stat.} & \textbf{Time}
& \textbf{\#Srv/\#R} & \textbf{Stat.} & \textbf{Time} \\
\midrule
100  & 1/1   & \greencheck & 00:26:40 & 2/50  & \greencheck & 00:14:30 & 8/230 & \greencheck & 00:05:30 \\
300  & 1/1   & \greencheck & 01:25:30 & 2/50  & \greencheck & 00:44:20 & 8/230 & \greencheck & 00:10:45 \\
1000 & 1/1   & \redcross   & -- & 2/50  & \redcross & -- & 8/230 & \greencheck & 00:42:45 \\
3000 & 1/1   & \redcross   & -- & 2/50  & \redcross & -- & 8/230 & \greencheck & 01:40:20 \\
5647 & 1/1   & \redcross   & -- & 2/50  & \redcross & -- & 8/230 & \greencheck & 03:03:03 \\
\bottomrule
\end{tabular}}
\caption{
Inference-time scaling under different rollout configurations. \#Srv denotes the number of inference servers and \#R the maximum number of parallel rollouts supported. Time reports the end-to-end rollout time per training epoch in \texttt{hh:mm:ss} format. ART-F achieves near-linear inference speedups by scaling inference servers and matching client request speed on multiple GPUs.
}
\label{tab:inference_speedup}\label{tab:inference_time_scaling}
\end{table}

To understand how ART-F accelerates and stabilizes RL training, we analyze three key aspects:

\textbf{(1) Rollout speed-up at the inference time}: By deploying a self-managed inference cluster with coordinated client request scheduling, we maximize GPU utilization and achieve near-linear inference throughput scaling with respect to the number of GPUs. Table~\ref{tab:inference_speedup} shows the inference-time scaling under different rollout configurations. Beyond rollout acceleration, we support concurrent startup of inference servers, retaining only a minimal process-level lock to avoid Unsloth initialization conflicts. This reduces startup latency per vLLM server from approximately 2 minutes to 1.5 minutes, corresponding to a 25\% speedup.

\textbf{(2) Training speed-up}: Although ART exposes a configurable training batch size, training tensors are constructed per instance when enqueued, which prevents the optimizer from fully exploiting batch-level parallelism. ART-F resolves this limitation by aligning tensor construction with the intended batch size, enabling true batched forward and backward passes. Combined with the dynamic batch sizing strategy described above, \textbf{ART-F achieves an approximately $N$-fold training throughput improvement, where $N$ denotes the effective batch size}. Another technical enhancement is that our DDP-supported ART-F training stabilizes RL training much earlier than the original ART single-GPU training. As a result, we observe smoother training curves when DDP is enabled.

\textbf{(3) Parallel Reward Evaluation and Tensor Persistence}: We also parallelize RULER client requests when groups of rollout trajectories are evaluated by the reward model cluster. We introduce the parameter \texttt{--max-concurrent-api-number} in the ART-F client, which enables concurrent reward-evaluation requests across multiple trajectories. This change reduces reward evaluation time from hours to approximately 10 minutes \textbf{per training step}, bringing the full-epoch reward evaluation time on ABCD+ (5,647 dialogues) down to approximately 130 minutes \textbf{per training epoch}. 

Another significant optimization is the parallelization of training-tensor persistence after tokenization of rollout trajectories. After introducing parallel persistence in ART-F, tensor processing completes in approximately 4.2 minutes (compared to 10--12 minutes originally), yielding a 2.5--2.9$\times$ speedup in tensor persistence.

\section{ART-F Client Design} \label{sec:art_f_client_implementation}

This section presents a typical ART-F client design to support end-to-end training of \textbf{ProAct-Q4}. Specifically, we describe (1) loading RL data and initializing the ART-F model according to client-specific configurations; (2) configuring parallel client requests at the inference phase; and (3) supporting multiple reward formulations, with particular emphasis on composite reward designs.

\begin{figure}[t]
    \centering
    \includegraphics[width=\columnwidth]{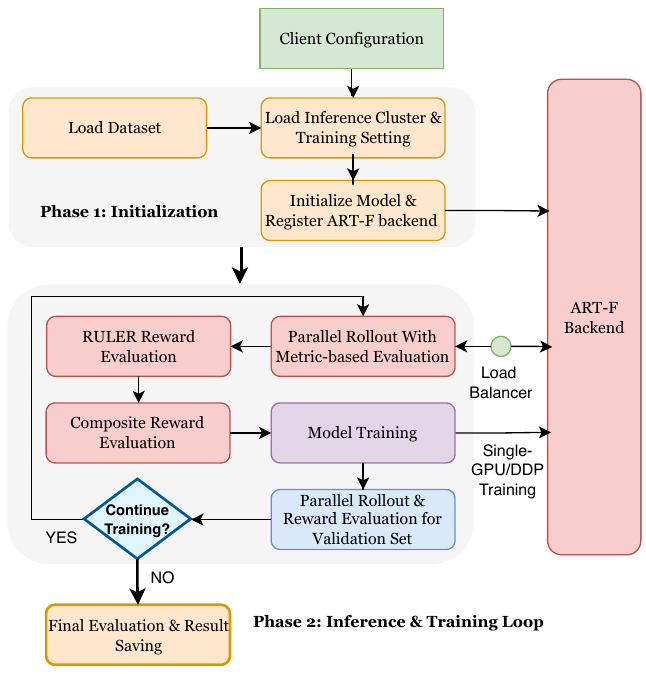}
    \caption{ART-F Client Workflow: The training client coordinates with the ART-F backend to manage parallel rollout requests through a load-balanced inference cluster, collect reward-annotated trajectories, and trigger single-GPU or DDP training, enabling efficient end-to-end reinforcement learning.}
    \label{fig:14}
\end{figure}

\subsection{Workflow}

As shown in Figure~\ref{fig:14}, a typical ART-F client workflow covers two phases: 

1) Initialization: The client first converts the raw dataset into rollout scenarios represented as serializable Python objects compatible with \texttt{Pydantic.BaseModel}, enabling them to be transmitted to the inference servers via HTTP requests. Dataset splitting into training, validation, and test sets is also performed at this stage. Next, the client loads a YAML-based configuration and merges it with the required command-line parameters to form a unified client configuration. This configuration is used to initialize the training job, register the model with the backend, and manage runtime behavior. For experiment tracking and reproducibility, the merged configuration is additionally logged to the corresponding Weights \& Biases~\cite{wandb2025website} and MLflow~\cite{zaharia2018mlflow} projects.

2) Inference \& Training Loop: Once the ART-F backend reaches the ready state, the client initiates parallel rollouts, with each inference server handling concurrent requests under semaphore-based rate limiting. During rollout, each scenario is executed $T$ times as specified by the client configuration, and each resulting trajectory is evaluated using metric-based rewards and tagged with the corresponding scores. If enabled, RULER-based reward evaluation is applied during rollout. 

Composite rewards are then computed, including \textbf{hybrid rewards} that combine RAC and RULER scores with fixed or dynamically scheduled weights, as well as \textbf{adaptive rewards} that emphasize different objectives (e.g., exploration, exploitation, or reliability) based on training progress. After reward processing, the client transitions to the training phase, where either single-GPU or distributed data-parallel (DDP) training optimizes the model using the collected trajectories. A validation rollout following the same procedure is scheduled after training.

Throughout the loop, comprehensive monitoring tracks training metrics, reward distributions, server utilization, and performance characteristics, with results logged to both Weights \& Biases~\cite{wandb2025website} and MLflow~\cite{zaharia2018mlflow} projects according to the project setting. The aforementioned loop continues until the predefined epoch limit is reached.

\subsection{Client Configuration} \label{cluster_configuration}

\begin{figure}[t!]
\begin{smallermdframed}

\text{\# Enhanced configuration for ART-F}\par
\text{logging:}\par
\text{\ \ wandb:}\par
\text{\ \ \ \ enabled: true  \# Enable logging to Weights \& Biases}\par
\text{\ \ \ \ project: "<Your project Name>" }\par
\text{\ \ \ \ model\_name: "<Your Model Name>"}\par
\text{\ \ \ \ weave\_trajectory\_tag: "<Trajectory Tag on Weave>"}\par
\text{\ \ \ \ weave\_enabled: false  \# Enable Weave on Wandb}\par
\text{\ \ mlflow:}\par
\text{\ \ \ \ enabled: true  \# Enable MLflow logging}\par
\text{\ \ \ \ local\_root: "./.mlruns"  \# Local MLflow tracking URI}\par
\text{\# LLM Settings}\par
\text{llm:}\par
\text{\ \ model: "Qwen/Qwen2.5-14B-Instruct"  \# model to use}\par
\text{\ \ max\_tokens: 9216      \# Maximum tokens for response}\par
\text{\ \ rollout:}\par
\text{\ \ \ \ sample\_num\_per\_training\_scenario: 4 \# Training Rollout Number}\par
\text{\ \ \ \ sample\_num\_per\_validation\_scenario: 2  \# Validation Rollout Number}\par
\text{\ \ \ \ temperature: 1.0      \# Temperature for rollout sampling}\par
\text{\ \ \ \ messages: "system.user.only" \# Message setting for rollout}\par
\text{\ \ \# Reward setting}\par
\text{\ \ trajectory\_reward\_rule: "schedule\_ruler\_weighted\_max\_rac\_score"} \par
\text{\ \ upper\_limit\_weight\_for\_scheduled\_ruler: 0.3} \par
\text{\ \ judger:}\par
\text{\ \ \ \ model: "openai/gpt-4.1-mini"  \# Model for judging actions}\par
\text{\ \ \ \ api\_key\_name: <INTERNAL\_API\_KEY>}\par
\text{\ \ \ \ base\_url: <INTERNAL\_SERVER\_BASE\_URL>}\par
\text{\ \ \ \ custom\_ruler\_placeholder:}\par
\text{\ \ \ \ \ \ - <Custom RULER>}\par
\text{\# Training Configuration}\par
\text{ddp\_training:}\par
\text{\ \ enable\_ddp: true                      \# Enable distributed data parallel}\par
\text{\ \ world\_size: 4                         \# Number of GPUs}\par
\text{\ \ ddp\_backend: "nccl"                   \# Backend: "nccl" (for CUDA GPUs)}\par
\text{\ \ ddp\_find\_unused\_parameters: false     \# Optimized for LoRA training}\par
\text{\ \ master\_addr: \$\{HOST\_IP\}              \# Master node address }\par
\text{\ \ master\_port: "29500"                  \# Master node port}\par
\text{\ \ ddp\_timeout\_minutes: 30              }\par
\text{\ \ batch\_size\_allow\_adjusting: true      \# Enable the dynamic training batch size}\par
\text{\ \ replicate\_dataset\_across\_ranks: true  \# Replicate data across all ranks}\par
\text{\# Inference Configuration}\par
\text{art\_multi\_server:}\par
\text{\ \ trainer\_args:}\par
\text{\ \ \ \ training\_batch\_size: 4                }\par
\text{\ \ \ \ logprob\_calculation\_chunk\_size: 1024 }\par
\text{\ \ \ \ max\_negative\_advantage\_importance\_sampling\_weight: 10 }\par
\text{\ \ server\_pool:}\par
\text{\ \ \ \ max\_servers\_per\_model: 8     \# Scale up to 8 servers}\par
\text{\ \ \ \ concurrent\_startup: true      \# Start multiple servers concurrently}\par
\text{\ \ \ \ port\_range\_start: 8000        \# Dedicated port range for training}\par
\text{\ \ \ \ port\_range\_end: 8100          \# 100 ports available for scaling}\par
\text{\ \ \ \ gpu\_memory\_per\_server: 0.4        \# 40\% GPU memory per GPU}\par
\text{\ \ \ \ scaling\_policy:               }\par
\text{\ \ \ \ \ \ enable\_auto\_scaling: true     }\par
\text{\ \ \ \ \ \ scale\_up\_before\_rollout: true }\par
\text{\ \ \ \ \ \ scale\_down\_before\_training: true }\par
\text{\ \ \ \ \ \ scaling\_timeout: 30         }\par
\text{\ \ rollout:}\par
\text{\ \ \ \ enable\_concurrent\_rollouts: true  \# Enable concurrent rollout processing}\par
\text{\ \ \ \ rollout\_batch\_size: 24             \# Process rollouts in batches of 24}\par
\text{\ \ \ \ load\_balancing\_strategy: "round\_robin"  \# Load Balancer Strategy}\par
\text{\# Rollout Prompts}\par
\text{tool\_catalog:}\par
\text{\ \ package: "Proactive\_Benchmark.synthetic\_data.proactive\_annotation}\par
\text{\ \ .abcd.tool\_catalog\_utils"}\par
\text{\ \ tool\_definition\_cls: "COMMON\_ABCD\_ANNOTATED\_TOOLING}\par
\text{\ \ \_API"}\par
\text{action\_prompts:}\par
\text{\ \ system\_prompt: |}\par
\text{\ \ \ \ You are an expert AI assistant for professional, specializing in analyzing}\par
\text{\ \ \ \  dialogues to identify automation opportunities and their parameters }\par
\text{\ \ \ \ based on the client and professional conversations.}\par
\text{\ \ \ \ Given the conversation history and available system action opportunities}\par
\text{\ \ \ \ , your task is to analyze the latest turn and determine the following}\par
\text{\ \ \ \ for all relevant actions:}\par
\text{\ \ \ \ 1. Action opportunities that could be triggered once ready}\par
\text{\ \ \ \ ......}\par
\text{\ \ task\_prompt: |}\par
\text{\ \ \ \ \#\# CONVERSATION AT TURN \{turn\_idx\}}\par
\text{\ \ \ \ \{dialogue\_context\}}\par
\end{smallermdframed}
\caption{ART-F Client YAML Configuration}
\label{fig:art_f_client_configuration}
\end{figure}

The ART-F client is initialized using a YAML configuration file that specifies all runtime settings for both inference and training, as illustrated in Figure~\ref{fig:art_f_client_configuration}. This configuration governs logging and experiment tracking (e.g., Weights \& Biases~\cite{wandb2025website} and MLflow~\cite{zaharia2018mlflow}), rollout behavior and reward setting, inference cluster scaling and load-balancing policies, distributed data-parallel (DDP)~\cite{pytorchDDP} training parameters, and prompt templates for action prediction. By externalizing these components into a declarative configuration, ART-F enables flexible experimentation and rapid system reconfiguration without modifying the client source code.

In this section, we will discuss the key settings for inference cluster and DDP training. Details of the rollout prompt design are discussed in Section~\ref{rollout_implementation}, while the reward formulation and client setup are described in Section~\ref{reward_setup}.

\paragraph{LLM Rollout Setting} To reduce rollout resource consumption, we support asymmetric sampling settings for training and validation during rollout, configured via \texttt{sample\_num\_per\_training\_scenario} and \texttt{sample\_num\_per\_validation\_scenario}.

\paragraph{Inference Cluster Setting} The \texttt{art\_multi\_server.server\_pool} configuration specifies inference cluster parameters, including \texttt{max\_servers\_per\_model}, which defines the maximum number of inference server instances that can be launched for a given model, and \texttt{gpu\_memory\_per\_server}, which controls the fraction of GPU memory reserved by each inference server on a single GPU. For instance, our LoRA-trained ProActor-Q4 model (9,216-token context) requires roughly 45 GB per inference server to support 15 concurrent rollout clients. On 4$\times$H200 GPUs (141 GB each), allocating \texttt{gpu\_memory\_per\_server}=0.4 enables two inference servers per GPU, allowing \texttt{max\_servers\_per\_model} to be safely set to 8. Parameters \texttt{port\_range\_start} and \texttt{port\_range\_end} allow us to specify the dedicated port range for the inference cluster, while enabling \texttt{concurrent\_startup} allows you to start the inference cluster in parallel with the minimal Unsloth process locking time. 

During inference cluster startup, ART-F resolves the server bind address by prioritizing \texttt{VLLM\_HOST\_IP}, then \texttt{HOST\_IP}, and finally falling back to \texttt{0.0.0.0}. This design is motivated by deployment environments such as Azure and Databricks, where binding to the default local or loopback address may fail.

\paragraph{Training Settings} General training parameters are specified under \texttt{art\_multi\_server.trainer\_args}, which defines settings shared by both single-GPU and DDP training, such as \texttt{training\_batch\_size} and \texttt{logprob\_calculation\_chunk\_size}. Distributed data-parallel (DDP)–specific configurations are isolated in a dedicated \texttt{ddp\_training} section. This section includes the DDP enable switch (\texttt{enable\_ddp}), standard PyTorch-style DDP parameters, support for dynamic batch size adjustment via \texttt{batch\_size\_allow\_adjusting}, and a symmetric data replication option (\texttt{replicate\_dataset\_across\_ranks}).

Similar to the bind-address resolution used during inference cluster initialization, ART-F also supports environment variable macros in \texttt{ddp\_training.master\_addr}, allowing the runtime to identify the IP address dynamically and bind with the DDP process group under heterogeneous or managed cloud environments.

\subsection{Rollout} \label{rollout_implementation}

\paragraph{Parallelization Setting} When \texttt{enable\_concurrent\_rollouts} is enabled, the ART-F client will launch \texttt{rollout\_batch\_size} parallel rollout groups for each scenario. Each rollout request is routed through the load balancer and assigned to an appropriate inference server according to the specified \texttt{load\_balancing\_strategy}, with fault recovery mechanisms to prevent bottlenecks on individual servers.

\paragraph{Rollout Prompt} For each scenario, the rollout client constructs inference requests by using \texttt{action\_prompts.system\_prompt} as the system message and injecting the current-turn dialogue context into \texttt{action\_prompts.task\_prompt} as the user message. The resulting prompt pair is sent to the assigned LLM inference server to generate corresponding action opportunities or questions. Along with the dialogue context, these messages form a dialogue-style rollout trajectory composed of interleaved SYSTEM and USER message segments.

ART-F supports three configurable message-construction strategies, specified via \texttt{llm.rollout.messages}, to accommodate both single-turn and multi-turn rollout scenarios:

\begin{enumerate}
\item \texttt{first.last} — Designed for multi-turn rollouts, this setting constructs the request using only the first and the last messages from the trajectory scenario, enabling lightweight context compression while preserving long-range intent.
\item \texttt{track} — Also intended for multi-turn rollouts, this setting includes the full sequence of messages from the trajectory scenario, providing the model with complete conversational context.
\item \texttt{system.user.only} — Used for single-turn rollouts, this setting restricts the request to the system and user messages only, omitting historical turns to minimize prompt length.
\end{enumerate}

\texttt{tool\_catalog} is also injected into the prompt as the tooling specification. Considering that \textbf{the standard MCP tooling call}~\cite{mcp2024} doesn't support partial parameter mapping, we send requests to the inference server as \textbf{a normal message request}, and save the response to the "Assistant" section with "tool\_call" mark in the returned response, yet \textbf{manually map the response body to the MCP-style action call} according to the action catalog specification. 

Based on the \texttt{llm.rollout} configuration, each training scenario is rolled out \texttt{sample\_num\_per\_training\_scenario} times, while each validation scenario is rolled out \texttt{sample\_num\_per\_validation\_scenario} times. All rollout trajectories generated from the same scenario are grouped into a rollout trajectory group, becoming the data used for RL training.

\paragraph{Metadata-Derived Metrics at Post-Rollout}\label{metadata_derived_metrics} As the metadata (dialogue information, reference actions, reference action range) has already been included in the rollout request body, \textbf{metadata-derived metrics}- \texttt{max\_rac\_score}, \texttt{rac\_score}, \texttt{timing\_reward\_in\_oracle}, and \texttt{fault\_trigger\_rate\_in\_oracle} - are computed immediately after rollout. Their values are then stored back in the metadata fields of the generated rollout trajectories. Meanwhile, ART-F automatically collects these metrics and reports them to the configured experiment-tracking backends.

\subsection{Reward Calculation} \label{reward_setup}

If \texttt{trajectory\_reward\_rule} specifies \textbf{a metadata-derived metric}, rewards are computed deterministically and stored in rollout metadata immediately after generation. By contrast, RULER-based and composite rewards require additional coordination and computation. We next describe how ART-F supports both reward categories in the client.

\paragraph{RULER-based rewards} \label{prompt_RULER} 

If \texttt{trajectory\_reward\_rule} is set to RULER-based rewards as "ruler", "hybrid\_ruler\_weighted\_rac\_score", "hybrid\_ruler\_weighted\_max\_rac\_score", "schedule\_ruler\_weighted\_rac\_score", "schedule\_ruler\_weighted\_max\_rac\_score", the collected rollout trajectories group will be further processed by the LLM judger to rate each trajectory in one comparison group based on the pre-defined rubric. 

The \texttt{judger} configuration specifies parameters for LLM-based evaluation, including \texttt{model}, \texttt{api\_key\_name}, \texttt{base\_url}, and an optional \texttt{custom\_ruler\_placeholder}. If no custom rules are provided, ART-F uses the default RULER rubric inherited from ART, referred to as \textbf{General RULER} (Section~\ref{general_ruler}). When custom rules are specified, they are appended to the judger prompt as additional evaluation criteria, yielding \textbf{Custom RULER} (Section~\ref{custom_ruler}).

\begin{figure}[t!]
\begin{smallermdframed}
\text{\#...Other Sections...}\par
\text{\ \ judger:}\par
\text{\ \ \ \ model: "openai/gpt-4.1-mini"  \# Model for judging actions}\par
\text{\ \ \ \ api\_key\_name: <INTERNAL\_API\_KEY>}\par
\text{\ \ \ \ base\_url: <INTERNAL\_SERVER\_BASE\_URL>}\par
\text{\ \ \ \ custom\_ruler\_placeholder:}\par
\text{\ \ \ \ \ \ - "Given the dialogue conversation, a trajectory with timely, context-}\par 
\text{\ \ \ \ \ \ aware actions should receive significantly higher scores than one that is}\par
\text{\ \ \ \ \ \ delayed or misaligned with the conversation. Specifically, 1. Reward }\par
\text{\ \ \ \ \ \ early and correct triggering: Give bonus credit to trajectories that mark}\par
\text{\ \ \ \ \ \ appropriate ready-to-trigger or triggered action opportunities as at the }\par
\text{\ \ \ \ \ \ earliest relevant dialogue turn in the conversation. The reward }\par
\text{\ \ \ \ \ \ diminishes as they are delayed. 2. Penalize inaccuracy: Deduct scores}\par
\text{\ \ \ \ \ \ for trajectories that incorrectly assign ready-to-trigger or triggered }\par
\text{\ \ \ \ \ \ status (false positives), or that fail to assign ready-to-trigger or }\par 
\text{\ \ \ \ \ \ triggered status when valid opportunities exist (false negatives)."}\par
\text{\#...Other Sections...}\par
\end{smallermdframed}
\caption{Our Custom RULER Setting for proactiveness evaluation}
\label{fig:custome_RULER_Setting_for_proactiveness_evaluation}
\end{figure}

Figure~\ref{fig:custome_RULER_Setting_for_proactiveness_evaluation} shows our criteria for proactiveness evaluation. From the experimental results, we clearly see that incorporating the proactiveness RULER can significantly improve the agent's behavior for proactive timing. ART-F framework takes custom RULER and merges with other rubrics, and from the completed prompt, as shown in Figure~\ref{fig:art_f_final_evaluation_prompt}, to evaluate a rollout trajectory group. 

\begin{figure}[t!]
\begin{smallermdframed}
\text{\textless system message\textgreater}\par
\text{All of the trajectories below have been given the same goal. Your job is to}\par
\text{consider each of them and assign a score between 0 and 1 based on your}\par
\text{best judgment of how well the agent achieves its goal.}\par
\text{Grading standards:}\par
\text{- A trajectory that achieves its goal should always receive a significantly}\par
\text{\ \ higher score than one that does not.}\par
\text{- A trajectory that achieves its goal more efficiently (e.g., avoiding}\par
\text{\ \ unproductive detours) should receive a higher score.}\par
\text{- If one trajectory is only slightly better than another, the score}\par
\text{\ \ difference should be small; if significantly better, the difference}\par
\text{\ \ should be large.}\par
\text{- \textless\textless Injected Custom RULER\textgreater\textgreater.}\par
\text{- Partial credit may be given for trajectories that make progress toward}\par
\text{\ \ the goal but do not complete it.}\par
\text{\textless/system message\textgreater}\par
\text{\textless user message\textgreater}\par
\text{\textless context\textgreater}\par
\text{[}\par
\text{\ \ \{"content": "\textless\textless Rollout System Message\textgreater\textgreater", "role": "user"\},}\par
\text{\ \ \{"content": "\textless\textless Rollout User Message\textgreater\textgreater", "role": "user"\}}\par
\text{]}\par
\text{\textless/context\textgreater}\par
\text{Trajectories:}\par
\text{\textless trajectory id="1"\textgreater}\par
\text{[}\par
\text{\ \ \{"role": "assistant", "content": "\textless think\textgreater\textless\textless ACTION Decision Explanation\textgreater\textgreater}\par
\text{\ \ \textless/think\textgreater\textbackslash n\textless action\textgreater\textless\textless ACTION\textgreater\textgreater\textless/action\textgreater"\}}\par
\text{]}\par
\text{\textless/trajectory\textgreater}\par
\text{\textless trajectory id="2"\textgreater}\par
\text{[}\par
\text{\ \ \{"role": "assistant", "content": "\textless think\textgreater\textless\textless ACTION Decision Explanation\textgreater\textgreater}\par
\text{\ \ \textless/think\textgreater\textbackslash n\textless action\textgreater\textless\textless ACTION\textgreater\textgreater\textless/action\textgreater"\}}\par
\text{]}\par
\text{\textless/trajectory\textgreater}\par
\text{\textless trajectory id="3"\textgreater}\par
\text{[}\par
\text{\ \ \{"role": "assistant", "content": "\textless think\textgreater\textless\textless ACTION Decision Explanation\textgreater\textgreater}\par
\text{\ \ \textless/think\textgreater\textbackslash n\textless action\textgreater\textless\textless ACTION\textgreater\textgreater\textless/action\textgreater"\}}\par
\text{]}\par
\text{\textless/trajectory\textgreater}\par
\text{\ldots If more trajectories are present.}\par
\end{smallermdframed}
\caption{LLM Judger Prompt for Trajectory-Level Proactiveness Evaluation with Custom RULER Injection}
\label{fig:art_f_final_evaluation_prompt}
\end{figure}

The Figure~\ref{fig:custome_RULER_result} shows how a sample rating in the JSON format from the LLM judger for the evaluation on a rollout group with four different trajectories.

\begin{figure}[t!]
\begin{smallermdframed}
\text{\{} \par
\text{\ \ "scores": [} \par
\text{\ \ \ \ \{} \par
\text{\ \ \ \ \ \ "trajectory\_id": "1",} \par
\text{\ \ \ \ \ \ "explanation": "<explanation for rating>",} \par
\text{\ \ \ \ \ \ "score": 0.95} \par
\text{\ \ \ \ \ \},} \par
\text{\ \ \ \ \{} \par
\text{\ \ \ \ \ \ "trajectory\_id": "2",} \par
\text{\ \ \ \ \ \ "explanation": "<explanation for rating>",} \par
\text{\ \ \ \ \ \ "score": 0.90} \par
\text{\ \ \ \ \ \},} \par
\text{\ \ \ \ \{} \par
\text{\ \ \ \ \ \ "trajectory\_id": "3",} \par
\text{\ \ \ \ \ \ "explanation": "<explanation for rating>",} \par
\text{\ \ \ \ \ \ "score": 0.85} \par
\text{\ \ \ \ \ \},} \par
\text{\ \ \ \ \ldots} \par
\text{\ \ \ \ \{} \par
\text{\ \ \ \ \ \ "trajectory\_id": "N",} \par
\text{\ \ \ \ \ \ "explanation": "<explanation for rating>",} \par
\text{\ \ \ \ \ \ "score": 0.30} \par
\text{\ \ \ \ \ \}} \par
\text{\ \ ]} \par
\text{\}} \par
\end{smallermdframed}
\caption{JSON Response Format Returned by the LLM Judger for Proactiveness Evaluation}
\label{fig:custome_RULER_result}
\end{figure}

\paragraph{Composite Rewards} \label{compositive_reward_calculation}

When both metadata-derived metrics and RULER-based metrics are available, ART-F supports composing them into \emph{composite rewards} according to user-defined reward formulations. Broadly, we categorize composite rewards into two classes:

\begin{enumerate}
\item \textbf{Weighted rewards.}
This category combines multiple metrics using \textbf{fixed weighting coefficients specified in the client configuration}. The weighting scheme is independent of training progress and does not require access to the current or total number of training steps, including \texttt{weighted\_max\_rac\_score}, \texttt{weighted\_rac\_score}, \texttt{hybrid\_ruler\_weighted\_rac\_score}, and \texttt{hybrid\_ruler\_weighted\_max\_rac\_score}.

\item \textbf{Stage-aware weighted rewards.}  
In contrast, this category explicitly incorporates training progress $\frac{u}{U}$ into reward computation by adapting metric weights according to the current training step $u$ relative to the total number of steps $U$, including \texttt{adaptive\_metric\_score}, \texttt{schedule\_ruler\_weighted\_max\_rac\_score}, and \texttt{schedule\_ruler\_weighted\_rac\_score}. Detailed formulations are provided in Equations~\ref{eq:adaptive_metrics} and~\ref{eq:adaptive_ruler}.
\end{enumerate}

\section{RL Methodology Details}\label{sec:rl_details}

This section provides additional mathematical formalization for the RL approach described in Section~\ref{sec:rl_method}, including both turn-level and trajectory-level considerations.

\subsection{Task Formulation}

Given a conversational dataset $\mathbf{D}$, each dialogue $D[i]$ consists of $N = |D[i]|$ annotated turns:
\begin{equation}
D[i,t] = \big(C[i,t],\, A^{\text{ref}}_{i,t}\big), \quad t \in \{1,\dots,N\},
\end{equation}
where $C[i,t]$ denotes the dialogue content at turn $t$, and $A^{\text{ref}}_{i,t}$ is the subset of reference actions annotated for that turn.

Let $\tau[i,\le t] = \{C[i,1], \dots, C[i,t]\}$ denote the dialogue trajectory up to turn $t$.
Given an unified action catalog $\mathcal{A}$ and a reward function $R$, the model defines a stochastic policy $\pi_\theta$ that conditions on $\tau[i,\le t]$ and samples a set of action candidates
\begin{equation}
\hat{A}_{i,t} = \{\hat{a}^{(k)}_{i,t}\}_{k=1}^{K}, \quad \hat{a}^{(k)}_{i,t} \sim \pi_\theta(\cdot \mid \tau[i,\le t]).
\end{equation}

The objective is to optimize $\pi_\theta$ to maximize the expected reward comparing predicted and reference action sets:
\begin{equation}
\max_{\theta}\;
\mathbb{E}_{i,t}\!\left[
R\!\left(\hat{A}_{i,t},\, A^{\text{ref}}_{i,t} \mid \tau[i,\le t]\right)
\right].
\end{equation}

Here the notation is a little bit different from Section~\ref{sec:proactiveness_metrics} and Section~\ref{sec:rl_method}, where we directly use $\hat{A}_t$ and $\hat{A}^{\text{ref}}_t$ at the dialogue turn $t$ and simply omits the dialogue index $i$, as it is clear from context.

\subsection{Metric-Based Rewards}

The trajectory-level accumulated RAC reward up to turn $t$ is defined as
\begin{equation}
\text{RAC}(\tau[i,\le t]) 
= \frac{1}{t}\sum_{t'=1}^{t} 
\text{AC}\!\left(\hat{A}_{i,t'},\, A^{\text{ref}}_{i,t'}\right),
\end{equation}
where $\hat{A}_{i,t'}$ and $A^{\text{ref}}_{i,t'}$ denote the predicted and reference action sets at turn $t'$.

The corresponding turn-level RAC reward is
\begin{equation}
\text{RAC}(\tau[i,t]) 
= \text{AC}\!\left(\hat{A}_{i,t},\, A^{\text{ref}}_{i,t}\right).
\end{equation}

Similarly, the trajectory-level Max RAC reward up to turn $t$ is
\begin{equation}
\resizebox{\columnwidth}{!}{$
\text{Max RAC}(\tau[i,\le t]) 
= \frac{1}{t} \sum_{t'=1}^{t} 
\max_{\hat{a} \in \hat{A}_{i,t'}} 
\text{AC}\!\left(\{\hat{a}\},\, A^{\text{ref}}_{i,t'}\right),
$}
\end{equation}
which selects the highest-consistency action within the predicted set at each turn.

The corresponding turn-level Max RAC reward is
\begin{equation}
\text{Max RAC}(\tau[i,t]) 
= \max_{\hat{a} \in \hat{A}_{i,t}} 
\text{AC}\!\left(\{\hat{a}\},\, A^{\text{ref}}_{i,t}\right).
\end{equation}

\subsection{RULER-Based Rewards}

RULER (Relative Universal LLM-Elicited Rewards)~\cite{openpipe2025ruler} evaluates the quality of predicted action candidates using rubric-based LLM judgments. 
Given a dialogue trajectory $\tau[i,\le t]$, the trajectory-level accumulated RULER reward is defined as
\begin{equation}
\resizebox{0.9\columnwidth}{!}{$
\text{RULER}(\tau[i,\le t]) 
= \frac{1}{t}\sum_{t'=1}^{t} 
\text{Judger}\!\left(\hat{A}_{i,t'},\, P,\, \tau[i,\le t']\right),
$}
\end{equation}
where $P$ denotes the instruction prompt used by the LLM-based evaluator.

The corresponding turn-level RULER reward is
\begin{equation}
\resizebox{0.9\columnwidth}{!}{$
\text{RULER}(\tau[i,t]) 
= \text{Judger}\!\left(\hat{A}_{i,t},\, P,\, \tau[i,\le t]\right).
$}
\end{equation}

To capture proactiveness nuances not fully specified by $P$, we further introduce a set of custom rubric rules $C$, yielding the turn-level Custom RULER reward:
\begin{equation}
\resizebox{0.9\columnwidth}{!}{$
\text{CUSTOM RULER}(\tau[i,t]) 
= \text{Judger}\!\left(\hat{A}_{i,t},\, P,\, C,\, \tau[i,\le t]\right).
$}
\end{equation}

Similarly, a trajectory-level custom ruler can be defined as:

\begin{equation}
\resizebox{0.9\columnwidth}{!}{$
\text{CUSTOM RULER}(\tau[i,\le t]) 
= \frac{1}{t}\sum_{t'=1}^{t} 
\text{Judger}\!\left(\hat{A}_{i,t'},\, P,\, C,\, \tau[i,\le t']\right),
$}
\end{equation}

From these formulations, we observe that preliminary trajectory-level reward designs fail to capture fine-grained action dynamics along rollout trajectories and therefore are ill-suited for providing dense reward signals to optimize RL models. However, \textbf{hierarchical reward formulations that combine turn-level and trajectory-level signals can better balance reward density and learning efficiency}, as demonstrated in prior work on hierarchical and temporally abstract reinforcement learning~\cite{zhou2024archer, ji2024hierarchical, wientjes2024successor}.

\subsection{Turn-Level GRPO Formulation} \label{turn_level_grpo_equation}
For each turn $t$ in conversation $c$, we sample $K$ action candidates and compute turn-specific rewards. At the optimization step $u$, the policy gradient update is:
\begin{equation}
\resizebox{\columnwidth}{!}{$
\nabla_\theta \mathcal{L}_{\text{cap-clip}}
=
\mathbb{E}_{u,c}\!\left[
\sum_{k=1}^{K}
g_u^{(k)} \, \nabla_\theta \log \pi_\theta(a_u^{(k)}|s_u)
\right]
$}
\end{equation}

\begin{equation}
\resizebox{\columnwidth}{!}{$
g_u^{(k)}
=
\begin{cases}
\bar r_u^{(k)} A_u^{(k)}, & \text{if } \bar r_u^{(k)} A_u^{(k)} \le
\operatorname{clip}(\bar r_u^{(k)}, 1-\epsilon, 1+\epsilon_{\text{high}})A_u^{(k)},\\
\operatorname{clip}(\bar r_u^{(k)}, 1-\epsilon, 1+\epsilon_{\text{high}})A_u^{(k)}, & \text{otherwise.}
\end{cases}
$}
\end{equation}
where $c$ denotes the episode-level context (e.g., dialogue history and task configuration) sampled from the training distribution, $A_u^{(k)} \neq 0$ are computed from turn-level rewards rather than trajectory returns, and $\bar r_u^{(k)}$ denotes the importance sampling ratio cap ($\bar r_u^{(k)} =  10$, parameter tuning via performance consistency check, refer to Section~\ref{importance_sampling_ratio_cap}).

\section{Dataset Details}\label{sec:dataset_details}

This section provides additional details on the datasets used for training and evaluation. We create two datasets representing the two general types encountered in real-world task scheduling: (1) systems \textit{with} historical action execution logs, where observed triggers enable alignment validation, and (2) systems \textit{without} such logs, where only conversation transcripts are available. This design validates that our framework generalizes across both scenarios.

\subsection{ABCD+ Dataset Details}

We enhance ABCD~\citep{chen2021action} with proactive annotations. The original data contains 10,042 task-oriented dialogues with 30 actions across customer service scenarios. Critically, our oracle annotation operates purely from conversation context---observed triggers are \textit{not} required for the annotation process itself, enabling the same pipeline to work across domains. However, when observed triggers are available, they enable post-annotation quality validation: we measure alignment between oracle annotations and actual system executions, achieving 87.14\% alignment. After filtering dialogues with quality scores $>0.8$, we retain approximately 70\% of the data---7,042 dialogues (5,647/703/692 train/dev/test) with 114,978 proactive action annotations. Validation details are in Appendix~\ref{sec:annotation_pipeline}.

\paragraph{Quality Validation.}
We validate annotation quality using the reference alignment validation~\ref{alignment_validation}. For each triggered action, we compute Early Ready Criteria (EC)---whether oracle annotations predict actions before/at trigger time, as mentioned in Section~\ref{alignment_validation_on_abcd}. Our validation on the full dataset shows 87.14\% EC success rate, with 70.27\% of dialogues achieving quality scores $>0.8$.

\paragraph{ABCD+ Statistics.}
The enhanced dataset contains 114,978 proactive action annotations with 80.89\% coverage of triggered actions. Each dialogue averages 3.6 actions with $5.98 \pm 2.86$ valid reference ranges. For high-quality training data, we filter dialogues with oracle scores $>0.8$, yielding 7,042 dialogues (5,647/703/692 train/dev/test). This filtered dataset serves as our primary benchmark for proactive dialogue agent evaluation.

\subsection{Home Loan Dataset Details}

To demonstrate generalizability, we apply our pipeline to real mortgage consultations---phone call transcripts between loan officers and clients in a financial services domain, distinct from ABCD's customer support context. This dataset contains only conversational data without any observed software action triggers. The dataset covers 13 action types across 6 workflow categories. Domain adaptation requires \textit{only configuration changes}---no code modifications---validating our pipeline's domain-agnostic design. 

\paragraph{Challenges Without Action Observations.}
Without recorded action triggers, our LLM-based oracle must infer appropriate action timing purely from conversational context. We address this through enhanced annotation guidelines: (1) conservative triggering requiring clear evidence of readiness, (2) strict parameter tracking to ensure all required inputs are present before marking actions as ready, and (3) multi-level readiness assessment considering both client intent and information completeness.

\paragraph{Privacy and Infrastructure.}
Due to stringent privacy requirements, all data processing is conducted within secure environments with limited GPU resources. These constraints reflect realistic enterprise settings and highlight our framework’s ability to operate effectively while preserving annotation quality. Moreover, the lack of observed triggers, together with strict privacy restrictions, makes this dataset particularly challenging. Despite these limitations, our automated annotation pipeline can be readily configured to identify proactive opportunities throughout mortgage consultations without additional coding, enabling effective automation of complex workflows even in the absence of historical action data.

\section{Training Setup}\label{sec:training_setup}

\paragraph{ABCD+ Configuration.}
On our internal platform, we use a 4$\times$H200 GPU server (each GPU with 141 GB memory) to conduct RL training on the ABCD+ dataset. For the inference server cluster, ART-F launches 8 inference instances, each configured to use 40\% of a single GPU's memory. To match the cluster's throughput, we allocate 100--230 parallel rollout groups with training rollout sizes of 4 and validation rollout sizes 2, which fully saturates GPU and process capacity while maintaining stable performance. For training, we set the model's maximum sequence length to 9,216 tokens and use a maximum training batch size of 4 , with three parallel data-loading workers. We enable dynamic batch sizing, which automatically adjusts the effective batch size to respect a fixed training budget of 4 $\times$ 9,216 tokens per step. To ensure training stability and prevent memory over-utilization, we cap GPU utilization at 90\% per device. We further enable Distributed Data Parallel (DDP) training \texttt{enable\_ddp} and symmetric data mode \texttt{replicate\_dataset\_across\_ranks}.

\paragraph{Home Loan Configuration.}
On Databricks, we set up 8$\times$H100 GPU servers to do RL training on the Home Loan Dataset. Each of the GPUs has 80 GB of memory. Considering more GPU numbers while having less memory on each node, we reduce the training batch size to 2--3 with a maximum token length of 9,216, while using 8 inference instances, each configured to use 45\% of a single GPU's memory. Similarly, we also enable Distributed Data Parallel (DDP) training \texttt{enable\_ddp} and symmetric data mode \texttt{replicate\_dataset\_across\_ranks}.

\paragraph{LoRA Setting}
We use LoRA with rank $r{=}8$ and scaling factor $\alpha{=}16$, applied to attention ($q,k,v,o$) and MLP ($\text{gate},\text{up},\text{down}$) projections, with dropout disabled for maximal throughput, which aligns with the default ART-F PEFT setting as shown in Figure~\ref{fig:lora_setting_artf}.

\begin{figure}[t!]
\begin{smallermdframed}
\text{\# LoRA configuration used in ART-F}\par
\par
\text{\{} \par
\text{\ \ max\_lora\_rank: 8, \ \ \ \ \ \ \ \ \ \ \ \ \ \ \ \ \# Rank for efficiency}\par
\text{\ \ lora\_alpha: 16, \ \ \ \ \ \ \ \ \ \ \ \ \ \ \ \ \ \ \# Alpha = 2$\times$ rank}\par
\text{\ \ lora\_dropout: 0, \ \ \ \ \ \ \ \ \ \ \ \ \ \ \ \ \ \# Optimized for performance}\par
\text{\ \ target\_modules: [}\par
\text{\ \ \ \ 'q\_proj', 'k\_proj', 'v\_proj', 'o\_proj',}\par
\text{\ \ \ \ 'gate\_proj', 'up\_proj', 'down\_proj'}\par
\text{\ \ ]}\par
\text{\}} \par
\end{smallermdframed}
\caption{LoRA configuration used in ART-F.}
\label{fig:lora_setting_artf}
\end{figure}

\paragraph{Inference Time Scaling.}
Table~\ref{tab:inference_speedup} reports inference-time scaling under different rollout configurations. ART-F achieves near-linear inference speedups by scaling inference servers to match client request throughput across multiple GPUs, effectively mitigating GPU under-utilization and addressing scalability limitations.

\section{Extended Experimental Analysis}\label{sec:extended_analysis}

This section provides additional analyses of baseline performance and reinforcement learning improvements that were summarized in the main text. We also include GPT-4.1-mini, an early baseline, to complete the comparative evaluation.

\subsection{Rationale}

Proactive agents must master three interconnected capabilities: \textit{what} action to take, \textit{how} to parameterize it correctly, and \textit{when} to execute it. Traditional task-oriented dialogue evaluation focuses almost exclusively on the first two---measuring whether the agent selects correct actions with accurate parameters. However, for proactive automation, \textit{timing} is equally critical: an action predicted too late provides no proactive value, while one predicted too early may lack sufficient information for correct parameterization. We therefore organize our evaluation into two categories:

\begin{itemize}
\item \textbf{Action Consistency} (AC, Max AC, Consistency Difference) measures \textit{what}, \textit{how well}, and \textit{how reliably}. AC averages consistency across all predictions, while Max AC captures the best single-action quality per turn, accommodating agents that track multiple hypotheses. Consistency Difference quantifies prediction reliability---the relative gap between best-case and average performance. A high Difference indicates that the model occasionally ``gets lucky'' with one good action but is unreliable on average---a signature of guesswork rather than genuine understanding. Low Difference signals consistent, trustworthy predictions suitable for production deployment.

\item \textbf{Timing} (PT, FTR, RAR) measures \textit{when} and \textit{how aggressively}---whether the agent predicts action readiness at appropriate moments. Proactive Timing (PT) rewards predictions aligned with oracle readiness windows. Fault Trigger Rate (FTR) penalizes predictions of non-existent actions. Ready Action Rate (RAR) captures overall proactive engagement---the proportion of predictions marked as ready.
\end{itemize}

This design reveals the fundamental challenge of proactive agents: an agent optimizing purely for Action Consistency may become overly conservative, waiting until parameters are complete---achieving high precision but low proactiveness. Conversely, aggressive timing may sacrifice parameter accuracy and consistency. Effective proactive agents must balance both dimensions while maintaining reliable predictions, a challenge we find no baseline fully resolves.

\begin{table*}[t]
    \centering
    \setlength\tabcolsep{3pt}
    \resizebox{\textwidth}{!}{
    \begin{tabular}{l|c|ccc|ccc}
    \hline\toprule
     & \textbf{PRI}{\scriptsize~$\uparrow$}
     & \multicolumn{3}{c|}{\textbf{Consistency}}
     & \multicolumn{3}{c}{\textbf{Timing}} \\
    \cline{3-8}
     &
     & \textbf{AC}{\scriptsize~$\uparrow$}
     & \textbf{Max AC}{\scriptsize~$\uparrow$}
     & \textbf{Difference}{\scriptsize~$\downarrow$}
     & \textbf{Proactive Timing}{\scriptsize~$\uparrow$}
     & \textbf{Fault Trigger Rate}{\scriptsize~$\downarrow$}
     & \textbf{Ready Action Rate}{\scriptsize~$\uparrow$}  \\
    \hline\toprule

    \multicolumn{6}{l}{\textbf{ABCD+}} \\
    \hline
    GPT-4.1-mini Non-Reasoning
    & 0.4405
    & 0.363$\pm$0.097 & 0.524$\pm$0.040 & 0.363$\pm$0.097
    & 0.0989$\pm$0.0703 & 0.0080$\pm$0.0060 & 0.161$\pm$0.115 \\
    GPT-4.1-mini Reasoning
    & 0.3986
    & 0.315$\pm$0.111 & 0.550$\pm$0.036& 0.315$\pm$0.111
    & 0.0587$\pm$0.0766 & \textbf{0.0044$\pm$0.0062} & 0.214$\pm$0.003 \\
    GPT-4.1-mini Reasoning + ASG
    & 0.4092
    & 0.303$\pm$0.072 & 0.537$\pm$0.058	& 0.303$\pm$0.072
    & 0.0625$\pm$0.0864 & \textbf{0.0067$\pm$0.0095} & 0.281$\pm$0.003 \\
    GPT-5.1 Non-Reasoning
    & 0.5104
    & 0.318$\pm$0.005 & 0.706$\pm$0.012 & 0.717$\pm$0.013
    & 0.2023$\pm$0.0019 & 0.0460$\pm$0.0009 & 0.419$\pm$0.010 \\
    GPT-5.1 Reasoning
    & 0.6003
    & 0.429$\pm$0.006 & 0.789$\pm$0.002 & 0.839$\pm$0.026
    & 0.1643$\pm$0.0018 & 0.0165$\pm$0.0002 & 0.214$\pm$0.003 \\
    GPT-5.1 Reasoning + ASG
    & 0.5547
    & 0.420$\pm$0.004 & 0.733$\pm$0.007 & 0.712$\pm$0.063
    & 0.1874$\pm$0.0022 & 0.0647$\pm$0.0019 & 0.281$\pm$0.003 \\
    Gemini-2.5-flash Non-Reasoning $^{(4)}$
    & \textbf{0.6251}
    & 0.417$\pm$0.004 & \textbf{0.834$\pm$0.001} & 1.000$\pm$0.019
    & 0.2133$\pm$0.0030 & 0.0399$\pm$0.0006 & 0.288$\pm$0.003	 \\
    Gemini-2.5-flash Reasoning
    & 0.6216
    & \textbf{0.430$\pm$0.001} & 0.813$\pm$0.002 & 0.891$\pm$0.006
    & 0.1782$\pm$0.0011 & 0.0245$\pm$0.0006 & 0.252$\pm$0.001	 \\
    Gemini-2.5-flash Reasoning + ASG
    & 0.5257
    & 0.384$\pm$0.001 & 0.737$\pm$0.001 & 0.919$\pm$0.006
    & 0.1715$\pm$0.0015 & 0.0286$\pm$0.0005 & 0.241$\pm$0.004	 \\
    Claude-4 Non-Reasoning
    & 0.6216
    & 0.427$\pm$0.001 & 0.831$\pm$0.001 & 0.946$\pm$0.005
    & 0.2119$\pm$0.0013 & 0.0526$\pm$0.0012 & 0.297$\pm$0.001 \\
    Claude-4 Reasoning $^{(3)}$
    & \textbf{0.6318}
    & 0.421$\pm$0.001 & 0.749$\pm$0.004 & 0.779$\pm$0.010
    & 0.2136$\pm$0.0023 & 0.0482$\pm$0.0023 & 0.314$\pm$0.003 \\
    Claude-4 Reasoning + ASG
    & 0.6031
    & 0.403$\pm$0.005 & \textbf{0.852$\pm$0.004} & 1.114$\pm$0.028
    & 0.2269$\pm$0.0022 & 0.0634$\pm$0.0032 & 0.360$\pm$0.002 \\
    Qwen2.5-14B-Instruct Non-Reasoning
    & 0.2996
    & 0.295$\pm$0.004 & 0.564$\pm$0.002 & 0.914$\pm$0.004
    & 0.2237$\pm$0.0012	 & 0.0773$\pm$0.0010 & 0.515$\pm$0.002 \\
    Qwen2.5-14B-Instruct Reasoning
    & 0.6246
    & 0.423$\pm$0.005 & 0.449$\pm$0.006 & 0.062$\pm$0.019
    & 0.1842$\pm$0.0012 & 0.0307$\pm$0.0008 & 0.279$\pm$0.003 \\
    Qwen2.5-14B-Instruct Reasoning + ASG
    & 0.4331
    & 0.316$\pm$0.004 & 0.385$\pm$0.004 & 0.218$\pm$0.020
    & 0.1918$\pm$0.0043 & 0.0432$\pm$0.0011 & 0.359$\pm$0.004 \\
    Qwen2.5-14B-Instruct + SFT (LoRA)
    & 0.1700
    & 0.272$\pm$0.001 & 0.533$\pm$0.0002 & 0.960$\pm$0.020
    & 0.2097$\pm$0.0007 & 0.0912$\pm$0.0003 & 0.531$\pm$0.003 \\
    Qwen2.5-14B-Instruct + SFT (Full-tune)
    & 0.5553
    & 0.482$\pm$0.001 & 0.486$\pm$0.0003 & 0.008$\pm$0.002
    & 0.1141$\pm$0.0010 & 0.0034$\pm$0.0020 & 0.130$\pm$0.010 \\
    \hline
    Qwen2.5-14B-ProActor-Q4 + Custom RULER $^{(1)}$
    & \textbf{0.7293}
    & 0.426$\pm$0.015 & 0.484$\pm$0.019 & 0.136$\pm$0.048
    & \textbf{0.2347$\pm$0.0201} & 0.0708$\pm$0.0078 & \textbf{0.546$\pm$0.036} \\
    Qwen2.5-14B-ProActor-Q4 + Adaptive RULER $^{(2)}$
    & \textbf{0.6842}
    & \textbf{0.431$\pm$0.022} & 0.586$\pm$0.044 & 0.320$\pm$0.123
    & \textbf{0.2515$\pm$0.0272} & 0.1089$\pm$0.0083 & \textbf{0.521$\pm$0.052} \\
    \hline\toprule

    \multicolumn{6}{l}{\textbf{Home Loan}} \\
    \hline
    GPT-4.1-mini Non-Reasoning
    & 0.4835
    & 0.237$\pm$0.011 & 0.265$\pm$0.010 & 0.118$\pm$0.067
    & 0.0219$\pm$0.0015 & 0.0009$\pm$0.0002 & 0.074$\pm$0.006 \\
    GPT-4.1-mini Reasoning
    & 0.4882
    & 0.300$\pm$0.010 & 0.329$\pm$0.013 & 0.097$\pm$0.057
    & 0.0118$\pm$0.0026 & \textbf{0.0003$\pm$0.0002} & 0.057$\pm$0.002 \\
    GPT-4.1-mini Reasoning + ASG
    & 0.4652
    & 0.188$\pm$0.002 & 0.232$\pm$0.005 & 0.234$\pm$0.030
    & 0.0281$\pm$0.0098 & 0.0012$\pm$0.0010 & 0.151$\pm$0.013 \\
    GPT-5.1 Non-Reasoning
    & 0.5047
    & 0.272$\pm$0.001 & 0.467$\pm$0.003 & 0.717$\pm$0.013
    & 0.0462$\pm$0.0013 & 0.0049$\pm$0.0002 & 0.116$\pm$0.003	 \\
    GPT-5.1 Reasoning
    & 0.5047
    & 0.363$\pm$0.007 & \textbf{0.579$\pm$0.004} & 0.595$\pm$0.033
    & 0.0186$\pm$0.0026 & 0.0003$\pm$0.0003 & 0.031$\pm$0.002\\
    GPT-5.1 Reasoning + ASG
    & 0.5010
    & 0.281$\pm$0.007 & 0.481$\pm$0.013 & 0.712$\pm$0.063
    & 0.0374$\pm$0.0061 & 0.0034$\pm$0.0004	 & 0.105$\pm$0.010  \\
    Gemini-2.5-flash Non-Reasoning
    & 0.6165
    & 0.349$\pm$0.004 & \textbf{0.688$\pm$0.004} & 0.972$\pm$0.024
    & 0.0632$\pm$0.0011 & 0.0072$\pm$0.0005 & 0.173$\pm$0.005 \\
    Gemini-2.5-flash Reasoning $^{(1)}$
    & \textbf{0.7303}
    & 0.345$\pm$0.004 & 0.527$\pm$0.006	& 0.528$\pm$0.024
    & 0.0757$\pm$0.0011 & \textbf{0.0001$\pm$0.0001} & 0.241$\pm$0.002 \\
    Gemini-2.5-flash Reasoning + ASG
    & 0.6067
    & 0.283$\pm$0.007 & 0.479$\pm$0.002 & 0.693$\pm$0.042
    & 0.0764$\pm$0.0039 & 0.0063$\pm$0.0018 & 0.251$\pm$0.012 \\
    Claude-4 Non-Reasoning
    & 0.5416
    & 0.224$\pm$0.003 & 0.347$\pm$0.004 & 0.549$\pm$0.024
    & \textbf{0.0811$\pm$0.0029} & 0.0118$\pm$0.0008 & \textbf{0.332$\pm$0.002}\\
    Claude-4 Reasoning $^{(3)}$
    & \textbf{0.7039}
    & 0.332$\pm$0.006 & 0.472$\pm$0.003 & 0.422$\pm$0.028
    & 0.0742$\pm$0.0027 & 0.0063$\pm$0.0016 & 0.261$\pm$0.003\\
    Claude-4 Reasoning + ASG $^{(2)}$
    & \textbf{0.7262}
    & \textbf{0.375$\pm$0.002} & 0.607$\pm$0.006 & 0.619$\pm$0.022
    & 0.0760$\pm$0.0031 & 0.0127$\pm$0.0013 & 0.307$\pm$0.005\\
    Qwen2.5-14B-Instruct Non-Reasoning
    & 0.4288
    & 0.253$\pm$0.002 & 0.383$\pm$0.003 & 0.514$\pm$0.017
    & 0.0120$\pm$0.0006 & 0.0003$\pm$0.0004 & 0.054$\pm$0.005 \\
    Qwen2.5-14B-Instruct Reasoning
    & 0.4012
    & 0.217$\pm$0.005 & 0.239$\pm$0.009 & 0.101$\pm$0.049
    & 0.0037$\pm$0.0010 & 0.0006$\pm$0.0001 & 0.032$\pm$0.003 \\
    Qwen2.5-14B-Instruct Reasoning + ASG
    & 0.3223
    & 0.113$\pm$0.011 & 0.144$\pm$0.006 & 0.274$\pm$0.133
    & 0.0184$\pm$0.0083 & 0.0033$\pm$0.0022 & 0.090$\pm$0.014 \\
    \hline
    Qwen2.5-14B-ProActor-Q4 + Custom RULER
    & 0.5603
    & 0.206$\pm$0.024 & 0.234$\pm$0.021  & 0.137$\pm$0.161
    & \textbf{0.0846$\pm$0.0095} & 0.0355$\pm$0.0079 & \textbf{0.465$\pm$0.074}  \\
    Qwen2.5-14B-ProActor-Q4 + Adaptive RULER $^{(4)}$
    & \textbf{0.6232}
    & \textbf{0.395$\pm$0.029} & 0.466$\pm$0.038 & 0.180$\pm$0.129
    & 0.0501$\pm$0.0055 & 0.0131$\pm$0.0013 & 0.156$\pm$0.009 \\
    \bottomrule
    \end{tabular}}
    \caption{Model performance comparison: Baselines are averaged over 3 runs, while Qwen2.5-14B-ProActor-Q4 are aggregated by last $N=4$ steps. Adaptive RULER is using Max RAC with $\lambda_{\max}=0.3$. $^{(1)}$, $^{(2)}$, $^{(3)}$, $^{(4)}$ mark top-1, top-2, top-3, top-4 methods by PRI (Proactiveness Ranking Index, see Section~\ref{proactiveness_ranking_index}).}
    \label{table:main_results_extended}
\end{table*}

\subsection{Interpretive Principles}

Before analyzing results, we highlight three critical interpretive principles:

\textit{On Action Consistency}: AC measures alignment with oracle annotations, which represent \textbf{one valid timing choice among potentially many}. Oracle annotations are constructed with hindsight: annotators observe the complete conversation and label actions that \emph{were} or \emph{could have been} triggered. A genuinely proactive agent operating in real-time may identify opportunities \textit{earlier} than the oracle, before all parameters are available---valid proactive behavior that nonetheless receives lower AC scores. We complement AC with PT, which evaluates whether predictions \textit{will become} ready ($\exists \tau \geq t$ in reference range) rather than requiring exact timing match.

\textit{On Consistency Difference}: A high Consistency Difference (>0.5) is a strong indicator of \textbf{unreliable, guess-based predictions}. For example, Qwen2.5-14B-Instruct Non-Reasoning achieves the highest Max AC (0.564) on ABCD+ but also the highest Consistency Difference (0.914), revealing that while it occasionally identifies good actions, its average predictions are poor. Similar cases also appear on Gemini-2.5-flash, Claude-4.0. In contrast, reasoning-based methods dramatically reduce Consistency Difference (Qwen2.5-14B-Instruct: 0.914$\rightarrow$0.062), demonstrating that explicit reasoning improves prediction consistency, not just accuracy. For production deployment, a low or moderate Consistency Difference is essential---a model that is ``sometimes right'' is less valuable than one that is ``consistently good.''

\textit{On Fault Trigger Rate}: For proactive agents, \textbf{lower FTR is not always better}. FTR$\approx$0 often indicates that a model has abandoned proactiveness---achieving ``precision'' by rarely predicting actions as ready. The appropriate evaluation considers \textit{net proactive value} (correct actions minus weighted faults) rather than minimizing FTR in isolation.

\subsection{On Timing}

True proactiveness requires the agent to reason beyond the current dialogue context and schedule action opportunities \textbf{as early as possible once the necessary conditions are confidently satisfied}. To assess this capability, we consider: (1) \textbf{Proactive Timing (PT)}, measuring the proportion of predicted ready actions that \emph{will} become valid (true positive ready-action rate); (2) \textbf{Fault Trigger Rate (FTR)}, quantifying the proportion that \emph{will not} become valid (false positive ready-action rate); and (3) \textbf{Ready Action Rate (RAR)}, capturing overall proactive engagement. Effective proactive agents should achieve strong PT and high RAR while keeping FTR within acceptable bounds. Formal definitions can be found in Section~\ref{sec:proactiveness_metrics}.

\subsection{Proactiveness Ranking Index for Model Comparison}\label{proactiveness_ranking_index}
\label{sec:proactiveness_ranking_index}

While individual proactive metrics (AC, PT, FTR, RAR) provide specific insights into model behavior, comparing models across different datasets and scales requires a unified evaluation framework. We introduce the \textbf{Proactiveness Ranking Index (PRI)}, a composite ranking metric that aggregates consistency and timing performance into a single comparable measure for systematic model ordering and selection.

\subsubsection{Motivation and Design Principles}

Proactive conversational agents must balance two critical aspects: \textit{consistency} in action prediction and \textit{timing} in proactive interventions. A model that achieves high action consistency (AC, Max\_AC) but poor timing (high FTR, low PT) may be less effective than one with moderate consistency but excellent timing. The PRI metric addresses this by:

\begin{enumerate}
    \item \textbf{Dimensional Reduction}: Aggregating multiple correlated metrics into interpretable indices
    \item \textbf{Balanced Evaluation}: Using harmonic mean to prevent one dimension from dominating
    \item \textbf{Relative Ranking}: Enabling fair comparison across different dataset scales and characteristics
\end{enumerate}

\subsubsection{Mathematical Formulation}\label{PRI_definition}

Given a set of evaluation metrics $\mathbf{M} \equiv \{\text{AC}, \text{Max\_AC}, \text{Difference}, \text{PT}, \text{FTR}, \text{RAR}\}$ and a comparison group $G = \{M_1, M_2, \ldots, M_n\}$ evaluated on the same dataset, we compute PRI through the following steps:

\paragraph{Step 1: Metric Normalization} For each metric $m \in \mathbf{M}$, we apply min-max normalization within group $G$:

\begin{equation}
m_{\text{norm}}^{(i)} = \frac{m^{(i)} - \min_{j \in G} m^{(j)}}{\max_{j \in G} m^{(j)} - \min_{j \in G} m^{(j)}}
\end{equation}

where $m^{(i)}$ denotes the metric value for model $M_i$. When $\max = \min$, we set $m_{\text{norm}}^{(i)} = 0.5$.

\paragraph{Step 2: Index Computation}
We define two composite indices and assume each factor takes an equal weight:

\textbf{Consistency Index (CI):}
\begin{equation}
\text{CI}^{(i)} = \frac{\text{AC}_{\text{norm}}^{(i)} + \text{Max\_AC}_{\text{norm}}^{(i)} + (1 - \text{Diff}_{\text{norm}}^{(i)})}{3}
\end{equation}

\textbf{Timing Index (TI):}
\begin{equation}
\text{TI}^{(i)} = \frac{\text{PT}_{\text{norm}}^{(i)} + (1 - \text{FTR}_{\text{norm}}^{(i)}) + \text{RAR}_{\text{norm}}^{(i)}}{3}
\end{equation}

The CI measures action prediction consistency, where higher AC and Max\_AC values and lower Difference values indicate better performance. The TI captures timing effectiveness, where higher PT and RAR values and lower FTR values indicate better proactive timing.

\paragraph{Step 3: Harmonic Mean Aggregation}
The final Proactiveness Ranking Index uses harmonic mean to ensure balanced performance:

\begin{equation}
\text{PRI}^{(i)} = \frac{2 \times \text{CI}^{(i)} \times \text{TI}^{(i)}}{\text{CI}^{(i)} + \text{TI}^{(i)}}
\end{equation}

To ensure numerical stability, we apply epsilon protection: $\text{CI}^{(i)}, \text{TI}^{(i)} \geq \epsilon = 0.001$.

\subsubsection{Interpretation and Properties}

\paragraph{Index Range and Meaning}
PRI values range from 0 to 1, where:
\begin{itemize}
    \item \textbf{PRI $\approx$ 1.0}: Excellent performance in both consistency and timing
    \item \textbf{PRI $\approx$ 0.5}: Moderate performance or strong in one dimension but weak in another
    \item \textbf{PRI $\approx$ 0.0}: Poor performance in at least one critical dimension
\end{itemize}

\paragraph{Harmonic Mean Properties}
The harmonic mean penalizes imbalanced performance more severely than arithmetic mean. For example:
\begin{itemize}
    \item Balanced: CI=0.8, TI=0.8 $\Rightarrow$ PRI=0.8
    \item Imbalanced: CI=1.0, TI=0.6 $\Rightarrow$ PRI=0.75 (not 0.8)
    \item Severely imbalanced: CI=1.0, TI=0.1 $\Rightarrow$ PRI=0.18 (not 0.55)
\end{itemize}

This design encourages models that perform well across both dimensions rather than excelling in only one.

\subsubsection{Usage in Model Comparison}\label{usage_in_model_comparison}

We apply the proposed PRI to rank model performance in both the baseline comparison (Table~\ref{table:main_results_extended}) and the reward ablation study (Table~\ref{table:ablation_reward_merged}).

\paragraph{Baseline and ProActor Comparison.}
We construct two comparison groups corresponding to the ABCD+ and Home Loan datasets. Each group includes all prompting-based baselines as well as two ProActor-Q4 variants, optimized using the Custom RULER and Adaptive RULER rewards, respectively.

\paragraph{ProActor Reward Variation.}
To analyze the impact of different reward designs, we construct three comparison groups:
(1) a small-scale ABCD+ setting with 100 training and 50 test dialogues, used to isolate the effects of single-objective rewards;
(2) the full ABCD+ dataset with 5{,}647 training and 692 test dialogues, used to evaluate all proposed reward variations at scale; and
(3) the full Home Loan dataset with 774 training and 97 test dialogues, used to assess the same reward variations, including combinations of RAC and Max RAC with both the Weighted Metric and Adaptive RULER rewards.

\subsubsection{Limitations and Considerations}

\paragraph{Relative Nature}
PRI is a \textit{relative ranking metric}, not an absolute performance measure. A model with PRI=0.8 in one group may perform differently than a PRI=0.8 model in another group. Cross-group comparisons should be interpreted with caution.

\paragraph{Normalization Sensitivity}
In small comparison groups, outlier models can skew normalization. We address this by:
\begin{enumerate}
    \item Ensuring minimum group size of 3 models when possible
    \item Reporting raw metric values alongside PRI scores
    \item Documenting outlier cases (e.g., the ABCD+ 3000/600 performance degradation)
\end{enumerate}

\paragraph{Metric Weight Assumptions}
The equal weighting of AC, Max\_AC, and Difference in CI assumes these metrics have similar importance. Similarly, PT, FTR, and RAR receive equal weight in TI. Future work could explore learned or domain-specific weightings.

\subsection{Baseline Patterns}\label{baseline_pattern_details}

\paragraph{(1) Reasoning improves consistency but not timing.} 
Introducing explicit reasoning substantially reduces Consistency Difference (on average 19–20\% for non-Qwen baselines, and a much larger 80–93\% drop for Qwen), while its effect on AC and Max AC varies across models and settings. In contrast, reasoning does not reliably improve timing-related metrics: Proactive Timing frequently degrades, sometimes sharply (e.g., GPT-5.1 and Qwen on Home Loan), with only a few exceptions (Claude-4 on ABCD+ and Gemini-2.5-flash on Home Loan). These results suggest that while reasoning helps models produce actions more consistently, it does not reliably improve \emph{when} actions should be taken.

\begin{table}[t!]
\centering
\setlength{\tabcolsep}{4pt}
\resizebox{\columnwidth}{!}{
\begin{tabular}{l|rrrrrr}
\toprule
\textbf{Model} & $\Delta$AC & $\Delta$MaxAC & $\Delta$Difference & $\Delta$PT & $\Delta$FTR & $\Delta$RAR \\
\midrule
\multicolumn{7}{l}{\textbf{ABCD+ (Non-Reasoning $\rightarrow$ Reasoning)}}\\
\midrule
GPT-5.1                & +34.9\% & +11.8\% & -31.2\% & -18.8\% & -64.1\% & -48.9\% \\
Gemini-2.5-flash       &  +3.1\% &  -2.5\% & -10.9\% & -16.5\% & -38.6\% & -12.5\% \\
Claude-4               &  -1.4\% &  -9.9\% & -17.7\% &  +0.8\% &  -8.4\% &  +5.7\% \\
\hline
Qwen2.5-14B-Instruct   & +43.4\% & -20.4\% & -93.3\% & -17.7\% & -60.3\% & -45.8\% \\
\midrule
\multicolumn{7}{l}{\textbf{Home Loan (Non-Reasoning $\rightarrow$ Reasoning)}}\\
\midrule
GPT-5.1                & +33.5\% & +24.0\% & -17.0\% & -75.6\% & -93.9\% & -73.3\% \\
Gemini-2.5-flash       &  -1.1\% & -23.4\% & -45.6\% & +19.8\% & -98.6\% & +39.3\% \\
Claude-4               & +48.2\% & +36.0\% & -23.1\% &  -8.5\% & -46.6\% & -21.4\% \\
\hline
Qwen2.5-14B-Instruct   & -14.2\% & -37.6\% & -80.4\% & -69.2\% & +100.0\% & -40.7\% \\
\bottomrule
\end{tabular}}
\caption{Relative change (\%) from Non-Reasoning to Reasoning baselines: $\Delta_{\%}=\frac{\text{R}-\text{NR}}{\text{NR}}\times100\%$. (Large percentage changes may occur when baseline values are close to zero.)}
\label{tab:reasoning_change_ratio}
\end{table}

\paragraph{(2) ASG destabilizes reasoning policies with model-specific fault amplification.}
Adding ASG on top of reasoning baselines consistently amplifies fault risks and destabilizes consistency, with similar failure patterns observed across model families. For example, on ABCD+, \textbf{GPT-5.1} exhibits only modest gains in proactive timing (+14.1\%) and ready action rate (+31.3\%), but incurs a sharp increase in Fault Trigger Rate (+292.1\%) alongside declines in AC and Max AC. \textbf{Gemini-2.5-flash} shows no meaningful timing improvement (–3.8\% PT) while still increasing FTR (+16.7\%) and consistency difference (+3.1\%). \textbf{Claude-4} improves Max AC (+13.7\%), yet this comes at the cost of substantially higher inconsistency (+43.0\% Difference) and elevated FTR (+31.5\%). The degradation is most severe for \textbf{Qwen2.5-14B-Instruct}, where ASG causes a large consistency collapse (Difference +251.6\%) and a significant rise in FTR (+40.7\%) despite marginal timing gains.
  
A similar trend appears in the Home Loan domain, where ASG leads to extreme fault amplification across all models (e.g., FTR +1033\% for GPT-5.1 and +6200\% for Gemini-2.5-flash), even when proactive timing or action coverage improves. Together, these results indicate that ASG primarily increases action firing frequency without stabilizing decision quality, resulting in inconsistent and unsafe behavior when layered onto reasoning policies.

\begin{table}[t!]
\centering
\setlength{\tabcolsep}{4pt}
\resizebox{\columnwidth}{!}{
\begin{tabular}{l|rrrrrr}
\toprule
\textbf{Model} & $\Delta$AC & $\Delta$MaxAC & $\Delta$Difference & $\Delta$PT & $\Delta$FTR & $\Delta$RAR \\
\midrule
\multicolumn{7}{l}{\textbf{ABCD+ (Reasoning $\rightarrow$ Reasoning + ASG)}}\\
\midrule
GPT-5.1              &  -2.1\% &  -7.1\% & -15.1\% & +14.1\% & +292.1\% & +31.3\% \\
Gemini-2.5-flash     & -10.7\% &  -9.3\% &  +3.1\% &  -3.8\% & +16.7\%  &  -4.4\% \\
Claude-4             &  -4.3\% & +13.7\% & +43.0\% &  +6.2\% & +31.5\%  & +14.6\% \\
\hline
Qwen2.5-14B-Instruct & -25.3\% & -14.3\% & +251.6\% &  +4.1\% & +40.7\%  & +28.7\% \\
\midrule
\multicolumn{7}{l}{\textbf{Home Loan (Reasoning $\rightarrow$ Reasoning + ASG)}}\\
\midrule
GPT-5.1              & -22.6\% & -16.9\% & +19.7\% & +101.1\% & +1033.3\% & +238.7\% \\
Gemini-2.5-flash     & -18.0\% &  -9.1\% & +31.3\% &  +0.9\% & +6200.0\% &  +4.1\% \\
Claude-4             & +13.0\% & +28.6\% & +46.7\% &  +2.4\% & +101.6\% & +17.6\% \\
\hline
Qwen2.5-14B-Instruct & -47.9\% & -39.7\% & +171.3\% & +397.3\% & +450.0\%  & +181.3\% \\
\bottomrule
\end{tabular}}
\caption{Relative change (\%) from Reasoning to Reasoning + ASG baselines:
$\Delta_{\%}=\frac{\text{R+ASG}-\text{R}}{\text{R}}\times100\%$. (Large percentage changes may occur when baseline values are close to zero.)} 
\label{tab:asg_change_ratio}
\end{table}

\paragraph{(3) No baseline resolves the consistency--timing trade-off.}
\label{no_baseline_strick_balance}
Across both ABCD+ and Home Loan, existing baselines fail to jointly achieve stable action selection and timely execution. 

On \textbf{ABCD+}, strong baselines frequently attain high Max AC (up to 0.85), but this comes with substantially inflated Consistency Difference (typically 0.6--1.1), indicating unstable action selection. To control this instability, many models effectively behave conservatively, resulting in limited proactive timing and ready action rates (PT $\leq 0.22$, RAR $\leq 0.32$). While a few baselines achieve PT values closer to ProActor (e.g., Claude-4 and Gemini-2.5-flash), these gains are consistently accompanied by severe consistency degradation.

On \textbf{Home Loan}, the trade-off is more pronounced. Baselines that maintain lower Consistency Difference exhibit extremely conservative behavior, yielding very low proactive timing (typically PT $<0.08$) and ready action rates (RAR $<0.33$). Conversely, models that improve PT do so at the cost of sharply increased inconsistency, suggesting that timing gains arise from aggressive over-triggering rather than reliable temporal decision-making. 

In contrast, ProActor-Q4 uniquely maintains low Consistency Difference ($\approx 0.14$) while simultaneously achieving the highest proactive timing and ready action rates across both datasets, demonstrating that existing baselines do not resolve the consistency--timing trade-off.

\paragraph{(4) Near-zero FTR on baselines reflects conservative failure, not precision.}
A model that never makes mistakes is often a model that never attempts to be proactive. This failure mode is most evident on Home Loan, where GPT-5.1 Reasoning achieves a high Max AC (0.579$\pm$0.004) but exhibits extremely low proactive behavior (PT=0.0186$\pm$0.0026, RAR=0.031$\pm$0.002) alongside a near-zero Fault Trigger Rate (FTR=0.0003$\pm$0.0003). Combined with its large Consistency Difference (0.595$\pm$0.033), this pattern reveals a model that can identify well-matched actions but systematically avoids committing to them. The resulting low FTR is therefore achieved through excessive conservatism rather than precise temporal decision-making.

This behavior recurs across conservative baselines on Home Loan, where near-zero FTR values (typically 0.01--0.3\%) consistently coincide with severely diminished proactive capacity (RAR 3--16\%). Similar observations occur on ABCD+, where reasoning-based baselines maintain low FTR (typically below 2\%) yet fail to translate high Max AC into timely or reliable readiness decisions. Although these models occasionally act, their proactive timing and ready action rates remain capped (PT $\leq 0.17$, RAR $\leq 0.32$), indicating hesitation rather than confidence. Among baselines, \textbf{Gemini-2.5-flash }shows relatively stronger proactive timing (PT $\approx$ 0.21 on ABCD+ and up to 0.076 on Home Loan). However, this improvement comes with poor action consistency, reflected by a high Consistency Difference (often $>0.9$) and unstable AC/Max AC alignment, suggesting that better timing is achieved at the expense of reliable action selection.

In contrast, ProActor-Q4 adopts a recall-oriented but controlled strategy across both datasets. On Home Loan, it achieves substantially higher readiness (RAR=46.5\%) with a bounded FTR of 3.6\%, producing nearly 3$\times$ more correct ready actions than conservative baselines. These results suggest that minimizing FTR in isolation is a misleading objective; instead, effective proactive agents should optimize \emph{net proactive value}, balancing correct ready actions against manageable fault costs.

\paragraph{(6) No prompting strategy resolves the precision--consistency--timing trade-off.}
Across both ABCD+ and Home Loan, even the strongest prompting-based baselines consistently optimize one objective at the expense of others. Non-reasoning variants of state-of-the-art models (e.g., Gemini-2.5-flash and Claude-4) achieve very high Max AC on ABCD+ (up to 0.83--0.85), but suffer from severe inconsistency, with Consistency Difference frequently exceeding 0.9 and unstable alignment between AC and Max AC. 

Introducing explicit reasoning substantially improves action consistency and average AC across models, but leads to conservative decision policies that suppress proactive behavior. As a result, reasoning-based baselines exhibit low proactive timing and ready action rates (PT $\leq 0.17$, RAR $\leq 0.32$ on ABCD+; PT $<0.02$, RAR $\approx 0.03$ on Home Loan), despite maintaining strong Max AC. 

Augmenting reasoning with ASG partially improves timing awareness, yielding the highest PT among prompting strategies, but does so at a high cost to reliability: AC and Max AC degrade, Consistency Difference increases, and fault trigger rates rise sharply. Consequently, no prompting configuration simultaneously achieves high action quality, low inconsistency, timely execution, and meaningful readiness under acceptable fault rates. These systematic limitations motivate our reinforcement learning approach, which learns adaptive policies that balance precision, consistency, and timing beyond the limits of static prompting.

\paragraph{(7) SFT exhibits polarized failure modes.}\label{sft_baseline_comparison}

 For \textbf{SFT (LoRA)}, it drops to PRI = 0.1700, the lowest of all 19 models. It learns to fire actions aggressively (RAR=0.531, comparable to RL), but with catastrophic inconsistency: AC=0.272 (worst in group) with Difference=0.960 (near-maximum). This means it occasionally gets a good action (Max AC=0.533) but is wildly unreliable on average. The CI collapses to 0.098, and the harmonic mean punishes this severely, resulting in PRI = 0.17. LoRA SFT essentially memorizes surface patterns of "when to act" without reliably learning what to act on.
 
When it comes to \textbf{SFT (Full-tune)}, PRI turns to a mediocre value of 0.5553. The opposite pathology: it achieves the highest AC in the group (0.482) with near-zero Difference (0.008) — extremely consistent. But it becomes overly conservative: PT=0.1141 and RAR=0.130 are among the lowest, with near-zero FTR (0.0034). It learns to replicate reference actions accurately but refuses to act proactively. CI is excellent (0.881), but TI is poor (0.406) — the classic "never-wrong-because-never-tries" failure.

In comparison, \textbf{ProActor-Q4 resolves what SFT cannot}. Custom RULER achieves PRI=0.7293 (+329\% over LoRA SFT, +31\% over Full-tune SFT) by jointly optimizing consistency and timing through composite rewards. It maintains a low Difference (0.136) while achieving the highest PT (0.2347) and RAR (0.546) — neither of which SFT variants achieve. The root cause of the gap reflects on our assumption: SFT optimizes a single objective (next-token prediction on reference outputs), which forces it into one of two failure modes depending on capacity:

\begin{enumerate}
    \item \textbf{Underfitting (LoRA)}: Learns the timing signal (when to act) but lacks the capacity to learn the content signal (what action to predict), leading to high recall but low precision.
    \item \textbf{Overfitting (Full-tune)}: Learns the content signal (what action to predict) but fails to generalize the timing signal (when to act), leading to high precision but low proactiveness, with additional formatting errors during evaluation due to catastrophic forgetting.
\end{enumerate}

RL's composite reward (RAC + RULER) explicitly optimizes both dimensions simultaneously, enabling a calibrated trade-off that static supervised learning cannot discover.

\subsection{Detailed ProActor-Q4 Analysis}

In contrast, ProActor-Q4 models achieve a better balance between reference action alignment and proactive timing. They maintain performance comparable to strong baselines on action consistency, while substantially improving proactive timing and ready action rates, without incurring an unacceptable increase in fault trigger rates.

\paragraph{Q1: Does ProActor-Q4 preserve robustness in action consistency?}

ProActor-Q4 maintains \textbf{strong and well-balanced action consistency} across domains. We evaluate robustness using \textit{Consistency Difference}, which measures the gap between average and best-case action alignment and reflects susceptibility to sporadic or ``lucky'' matches.

\begin{enumerate}
    \item \textbf{ABCD+.} ProActor-Q4 + Custom RULER achieves the lowest Consistency Difference ($0.136\pm0.048$) with solid alignment (AC $=0.426\pm0.015$, Max AC $=0.484\pm0.019$). The Adaptive RULER variant increases action quality substantially (AC $=0.431\pm0.022$, Max AC $=0.586\pm0.044$) while keeping the Consistency Difference at a moderate level ($0.320\pm0.123$), far below those of strong prompting baselines such as Gemini-2.5-flash Non-Reasoning (Difference $=1.000\pm0.019$) and Claude-4 Reasoning + ASG (Difference $=1.114\pm0.028$). This indicates a favorable trade-off: Adaptive RULER sacrifices some consistency to obtain markedly stronger action alignment, without entering the unstable regime observed in prompting methods.
    
    \item \textbf{Home Loan.} Under domain shift, Adaptive RULER exhibits an even clearer balance. Compared to Custom RULER (AC $=0.206\pm0.024$, Max AC $=0.234\pm0.021$, Difference $=0.137\pm0.161$), Adaptive RULER nearly doubles alignment quality (AC $=0.395\pm0.029$, Max AC $=0.466\pm0.038$) while incurring only a modest increase in Consistency Difference ($0.180\pm0.129$). Both variants remain substantially more stable than strong prompting baselines such as GPT-5.1 Reasoning (Difference $=0.595\pm0.033$) and Gemini-2.5-flash Non-Reasoning (Difference $=0.972\pm0.024$).
\end{enumerate}

Overall, Adaptive RULER achieves a \textbf{more favorable consistency--alignment balance}, trading a controlled increase in variance for large gains in action quality, while avoiding the severe instability that characterizes prompting-based approaches.

\paragraph{Q2: ProActor-Q4 exhibits higher FTR due to increased action recall rather than over-triggering.}
To better understand the elevated Fault Trigger Rate (FTR) of ProActor-Q4 relative to more conservative baselines, we analyze the distribution of predicted ready actions per dialogue turn on the test sets (Table~\ref{tab:ready_action_distribution}). Despite higher FTR, the absolute number of actions predicted per turn remains small: on ABCD+, the model predicts fewer than one action per turn on average (mean 0.96, median 1), and on Home Loan the mean is 0.65 with a median of 0.5. In both datasets, 75\% of turns contain at most one predicted action, indicating that ProActor-Q4 does not excessively over-trigger actions.

\begin{table}[t!]
\centering
\setlength{\tabcolsep}{6pt}
\small
\begin{tabular}{l|cc}
\toprule
\textbf{Statistic} & \textbf{ABCD+} & \textbf{Home Loan} \\
\hline
Count              & 4{,}533 & 806 \\
Mean               & 0.96 & 0.65 \\
Std. Dev.          & 0.61 & 0.53 \\
Minimum            & 0.00 & 0.00 \\
25th Percentile    & 1.00 & 0.00 \\
Median (50\%)      & 1.00 & 0.50 \\
75th Percentile    & 1.00 & 1.00 \\
Maximum            & 10.00 & 3.00 \\
\bottomrule
\end{tabular}
\caption{Ready action number per dialogue turn delivered by the RL-trained model on test sets}
\label{tab:ready_action_distribution}
\end{table}

The observed increase in FTR therefore reflects \textbf{a deliberate shift toward higher action recall}, rather than uncontrolled over-triggering. This pattern holds for both ProActor-Q4 variants: Custom RULER maintains the lowest inconsistency with moderate FTR, while Adaptive RULER further increases recall with a controlled rise in fault rate. In both cases, the absolute number of predicted actions per turn remains small, indicating that higher FTR arises from acting more often when appropriate, not from action spamming.

Compared to aggressive ASG-based variants, which exhibit unstable consistency and sharply inflated fault rates, both ProActor-Q4 variants maintain substantially lower FTR while achieving stronger proactive timing and ready action rates. This suggests that the additional fault cost introduced by recall-oriented optimization is bounded and acceptable given the overall gains in proactive effectiveness.

\paragraph{Q3: Why does ProActor-Q4 succeed where prompting-based methods fall short?}

ProActor-Q4 succeeds by optimizing proactive behavior through \textbf{learning-based exploration of the action space}, rather than relying on static oracle examples or hand-crafted prompting rules. \textbf{The Custom RULER reward} explicitly encodes proactiveness requirements via rubric-based evaluation, providing dense, turn-level feedback that directly incentivizes timely and appropriate action readiness. This design leads to substantially stronger improvements in proactive timing and ready action rates than metric-only or prompting-based alternatives.

By adopting \textbf{composite rewards} that jointly account for action consistency and timing, ProActor-Q4 learns a calibrated trade-off between acting early and acting reliably. As demonstrated in our analysis, this allows ProActor-Q4 to significantly improve timing performance while keeping consistency degradation bounded. In contrast, prompting-based baselines exhibit polarized failure modes: reasoning-based methods tend toward excessive conservatism (e.g., GPT-5.1 Reasoning), while structure-driven prompting such as ASG induces uncalibrated aggressiveness with elevated fault rates and instability. These results highlight the advantage of reinforcement learning with explicit proactiveness supervision for resolving the precision--consistency--timing trade-off in proactive decision-making.

\section{More Ablations}\label{sec:ablation_details}

This section provides comprehensive ablation study results, including reward type comparisons, data scaling effects, and additional analysis tables.

\begin{table*}[t!]
\centering
\setlength\tabcolsep{3pt}
\resizebox{\textwidth}{!}{
\begin{tabular}{l|c|c|cc|cc|cc|cc|cc}
\hline\toprule
\textbf{\#Dialogs} 
& \textbf{Reward Type}
& \textbf{\textbf{PRI}{\scriptsize~$\uparrow$}}
& \multicolumn{2}{c|}{\textbf{AC}{\scriptsize~$\uparrow$}}
& \multicolumn{2}{c|}{\textbf{Max AC}{\scriptsize~$\uparrow$}}
& \multicolumn{2}{c|}{\textbf{Proactive Timing}{\scriptsize~$\uparrow$}}
& \multicolumn{2}{c|}{\textbf{Fault Trigger Rate}{\scriptsize~$\downarrow$}} 
& \multicolumn{2}{c}{\textbf{Ready Action Rate}{\scriptsize~$\uparrow$}}  \\
\cline{4-13}
\textbf{Train/Test} 
& 
&
& Statistics & $\Delta$
& Statistics & $\Delta$
& Statistics & $\Delta$
& Statistics & $\Delta$ 
& Statistics & $\Delta$\\
\hline

\multicolumn{10}{l}{\rule{0pt}{2.5ex}\textbf{ABCD+}} \\
\hline

100/50 
& RAC   
& 0.2596   
& 0.3239±0.01606 & -0.1528
& 0.3881±0.02590 & -0.1348 
& 0.1223±0.02364 & -0.1030 
& 0.0640±0.01960 & 0.9636  
& 0.3002±0.08088 &  0.2770  \\ 

100/50 
& Max RAC      
& 0.2264    
& 0.3263±0.02468 & -0.4740	
& 0.4384±0.02264 & -0.1587
& 0.1643±0.00207 & -0.0616	 
& 0.0886±0.01212 & 2.6500
& 0.3862±0.01111 & 0.2262 \\

100/50 
& General Ruler  
& 0.4918     
& 0.3494±0.02714  & -0.2750
& 0.3978±0.02984 & -0.2436
& 0.2324±0.01984  & 0.9056
& 0.1019±0.01910 & 3.7188 
& 0.5945±0.05405 & 1.4867 \\

100/50 
& Custom Ruler  $\dagger$
& \textbf{0.6140}    
& 0.3850±0.02544 & -0.1995
& 0.4203±0.03200  & -0.2476
& 0.2315±0.02393 & 0.6755
& 0.1068±0.02327  & 4.8358 
& 0.5707±0.05717  & 1.2846 \\

100/50 
& Weighted Metric (Max RAC)
& 0.5800     
& 0.3929±0.03797 & -0.0381
& 0.5099±0.06267 & 0.17834
& 0.2030±0.02414 & 0.55646
& 0.0818±0.01114 & 1.74 
& 0.4860±0.03368 & 0.6535 \\

100/50 
& Adaptive Metric   $\star$ 
& \textbf{0.8160}        
& 0.4429$\pm$0.02656 & -0.0291
& 0.4931$\pm$0.02914 & -0.01426
& 0.1989$\pm$0.01303 & 0.18889
& 0.0631$\pm$0.01157 & 0.6667 
& 0.4642$\pm$0.01688 & 0.5163\\






\hline

5647/692
& RAC 
& 0.4352      
& 0.3368$\pm$0.02401 & -0.1478
& 0.3857$\pm$0.01949 & -0.1327
& 0.1799$\pm$0.00452 & 0.0128
& 0.0621$\pm$0.00765 & 0.1760 
& 0.4148$\pm$0.02354 & 0.1818  \\

5647/692
& Max RAC  
& 0.3737      
& 0.3745$\pm$0.03439 & -0.0926
& 0.5014$\pm$0.06809 & 0.0492
& 0.1172$\pm$0.03605 & -0.0351
& \textbf{0.0365$\pm$0.01013} & 0.0919
& 0.2426±0.06160 & -0.0008  \\

5647/692
& Custom Ruler $\star$  
& \textbf{0.7217}    
& 0.4257$\pm$0.01469 & -0.4791
& 0.4837$\pm$0.01848 & -0.4731
& 0.2347$\pm$0.02014 & 0.56561
& 0.0708$\pm$0.00784 & 2.4654 
& 0.5456±0.04169 & 1.8482 \\

5647/692
& Weighted Metric (RAC) $\dagger$
& \textbf{0.7078}
& \textbf{0.4547±0.01284} &  0.0016
& 0.5130±0.01260 &  0.0932
& 0.1920±0.00905 &  0.2096
& 0.0503±0.00149 &  0.5371
& 0.4329±0.01393 &  0.4391\\

5647/692
& Weighted Metric (Max RAC) 
& 0.2169  
& 0.3068±0.02606 & -0.2975
& 0.4399±0.05253 & 0.0011
& 0.1465±0.03091 & 0.0338
& 0.0700±0.01253 & 1.5204
& 0.3769±0.05893 & 0.3607\\

5647/692
& Adaptive Metric 
& 0.5159   
& 0.3537$\pm$0.02264 & -0.1950
& 0.3922$\pm$0.03489 & -0.2524
& 0.2268$\pm$0.04591 & 0.2966
& 0.1137$\pm$0.01925 & 2.5661
& 0.6332±0.11113  & 1.0600 \\

5647/692
& Adaptive RULER ($\lambda_{\max}=0.3$, RAC) 
& 0.5921   
& 0.4515±0.01795 & 0.1139
& 0.5134±0.01762 & 0.1487
& 0.2076±0.01269 & 0.3667
& 0.1004±0.00314 & 1.4526
& 0.4612±0.05153 & 0.4523\\

5647/692
& Adaptive RULER ($\lambda_{\max}=0.3$, Max RAC) 
& 0.6026   
& 0.4314$\pm$0.02240 & -0.05045
& \textbf{0.5861$\pm$0.04335} & 0.1150  
& \textbf{0.2515$\pm$0.02722} & 0.4410 
& 0.1089$\pm$0.00832 & 1.9003
& 0.5212±0.05153 & 0.5818 \\

\hline\toprule


\multicolumn{10}{l}{\textbf{Home Loan}} \\
\hline

774/97
& RAC   
& 0.5668   
& \textbf{0.4701±0.03285} & 0.8483 
& \textbf{0.5195±0.03016} & 0.8462
& 0.0264±0.00590 & -0.2502 
& 0.0041±0.00245 & -0.7494
& 0.0612±0.01856 & -0.5365 \\

774/97
& Max RAC  
& 0.4311
& 0.3894±0.01327 & 0.4891
& 0.4780±0.02065 & 0.6927
& 0.0093±0.00378 & -0.8214 
& \textbf{0.0015±0.00072} & -0.9409
& 0.0177±0.00787 & -0.9178\\

774/97
& Custom Ruler 
& 0.4220   
& 0.2057±0.02409 & -0.3760
& 0.2338±0.02114 & -0.3329
& \textbf{0.0846±0.00951} & 2.5245
& 0.0355±0.00785 & 5.2040
& 0.4653±0.07412 & 4.8508\\

774/97
& Weighted Metric (RAC) 
& 0.5376   
& 0.4359±0.01376 & 0.6695
& 0.4805±0.02399 & 0.7160 
& 0.0205±0.00349 & -0.2859
& 0.0020±0.00066 & -0.8762
& 0.0481±0.00722 & -0.6039  \\

774/97
& Weighted Metric (Max RAC) 
& 0.4809  
& 0.4133±0.02947 & 0.6461
& 0.5346±0.03620  & 0.9601
& 0.0270±0.00730 & -0.4212 
& 0.0049±0.00090 & -0.4268 
& 0.0506±0.02000 & -0.7042\\

774/97
& Adaptive Metric 
& 0.5742  
& 0.4677$\pm$0.01760 & 0.8090
& 0.5184$\pm$0.02184 & 0.8390 
& 0.0282$\pm$0.00552 & -0.3170
& 0.0042$\pm$0.00121 & -0.7683
& 0.0653±0.01628  & -0.5919 \\

774/97
& Hybrid RULER ($\lambda=0.3$, RAC)
& 0.6672  
& 0.4418±0.02502 & 0.7347
& 0.4632±0.02591 & 0.6569
& 0.0492±0.00122 & 0.7427
& 0.0107±0.00091 & 0.4341
& 0.1725±0.00620 & 0.6519 \\

774/97
& Adaptive RULER ($\lambda_{\max}=0.3$, RAC) 
& 0.6154   
& 0.4173$\pm$0.00768 & 0.5111
& 0.4403$\pm$0.01019 & 0.4510
& 0.0397$\pm$0.00528 & 0.4382
& 0.0071$\pm$0.00070 & -0.1682  
& 0.1231$\pm$0.01985  & 0.4329     \\

774/97
& Hybrid RULER ($\lambda=0.5$, RAC) $\star$
& \textbf{0.6836}
& 0.4233$\pm$0.02586 & 0.6608
& 0.4501$\pm$0.02397 & 0.5636
& 0.0613$\pm$0.00751 & 1.2351
& 0.0154$\pm$0.00184 & 1.4814
& 0.2236±0.03846 & 1.4644 \\

774/97
& Adaptive RULER ($\lambda_{\max}=0.5$, RAC)
& 0.5416 
& 0.3919$\pm$0.01226 & 0.4050
& 0.4483$\pm$0.01509 & 0.4791
& 0.0325$\pm$0.00553 & 0.3638 
& 0.0063$\pm$0.00225 & -0.1435
& 0.0946±0.01411 & 0.1204 \\

774/97
& Hybrid RULER ($\lambda=0.75$, RAC) $\dagger$ 
& \textbf{0.6711}
& 0.3584$\pm$0.01861 & 0.3707
& 0.3786$\pm$0.01823 & 0.2833
& 0.0720$\pm$0.01011 & 1.5767
& 0.0181$\pm$0.00349 & 2.5504 
& 0.2931±0.05276 & 2.2065 \\

774/97
& Adaptive RULER ($\lambda_{\max}=0.75$, RAC)
& 0.6227 
& 0.4019$\pm$0.02993 & 0.5041 
& 0.4258$\pm$0.02664 & 0.4238
& 0.0447$\pm$0.00239 & 0.5757
& 0.0105$\pm$0.00150 & 0.3950 
& 0.1708±0.01436 & 0.8176 \\

774/97
& Hybrid RULER ($\lambda=0.3$, Max RAC) 
& 0.6112
& 0.3868$\pm$0.02672 & 0.4823 
& 0.4329$\pm$0.04006 & 0.5458
& 0.0595$\pm$0.00465 & 0.9393
& 0.0210$\pm$0.00250 & 1.9204
& 0.2540±0.01615 & 1.5509 \\

774/97
& Adaptive RULER ($\lambda_{\max}=0.3$, Max RAC) 
& 0.5734 
& 0.3950$\pm$0.02936 & 0.5366
& 0.4658$\pm$0.03821 & 0.6764
& 0.0501$\pm$0.00547 & 1.0451
& 0.0131$\pm$0.00131 & 1.5299
& 0.1561±0.00930 & 0.5190  \\

774/97
& Hybrid RULER ($\lambda=0.5$, Max RAC)
& 0.6533   
& 0.4114$\pm$0.02512 & 0.5827
& 0.4491$\pm$0.02862 & 0.5771
& 0.0580$\pm$0.00210 & 0.9779
& 0.0169$\pm$0.00252 & 0.8783
& 0.2385±0.01071 & 1.4148\\

774/97
& Adaptive RULER ($\lambda_{\max}=0.5$, Max RAC) 
& 0.5849   
& 0.3983$\pm$0.01492 & 0.4339
& 0.4536$\pm$0.01820 & 0.5338
& 0.0436$\pm$0.00421 & 0.9312
& 0.0086$\pm$0.00157 & 0.3971
& 0.1295±0.01488 & 0.5244 \\

774/97
& Hybrid RULER ($\lambda=0.75$, Max RAC)
& 0.5431
& 0.2795$\pm$0.01821 & -0.0153
& 0.3095$\pm$0.01610 & -0.0287
& 0.0711$\pm$0.01088 & 1.9046
& 0.0264$\pm$0.00743 & 2.5972
& 0.3683±0.07846 & 3.0949\\

774/97
& Adaptive RULER ($\lambda_{\max}=0.75$, Max RAC)
& 0.3357  
& 0.2448±0.01290 & -0.1612
& 0.3382±0.01703 & 0.0401
& 0.0715±0.01286 & 2.2181
& 0.0362±0.01368 & 6.5137
& 0.2977±0.09482 & 3.1734\\

\bottomrule
\end{tabular}}
\caption{Performance given different reward settings. \#Dialogs Train/Test denotes the dialogue number for training and testing. Statistics report mean$\pm$std on the test set. $^{\star}$, $^{\dagger}$ mark top-1, top-2 methods by PRI (Proactiveness Ranking Index, see Section~\ref{proactiveness_ranking_index}).}
\label{table:ablation_reward_extended}
\end{table*}

\subsection{Reward Type Comparison}

The full ablation results are presented in Table~\ref{table:ablation_reward_extended} in the main text. Here we provide detailed analysis.

\paragraph{Reward Types.} \label{reward_type} 

Among all evaluated reward types, \textbf{Custom RULER} achieves the most balanced performance, which supports our reward decision to leverage it when further scaling up data volumes. 

In terms of Action Consistency, Custom RULER attains the highest AC score (0.5540$\pm$0.0160), outperforming RAC, Max RAC, and General RULER, while maintaining competitive Max AC (0.6004$\pm$0.0243). In contrast, Adaptive RULER substantially harms action consistency (AC drops to 0.232$\pm$0.0089), suggesting that overly aggressive emphasis on timing can destabilize action alignment at small data scales. 

From a timing perspective, Custom RULER improves Proactive Timing (0.1548$\pm$0.0164) over RAC and General RULER, while keeping the Fault Trigger Rate at the lowest level among all reward variants (0.0106$\pm$0.0037). Although Weighted Metric and Adaptive Metric rewards achieve higher Proactive Timing, they do so at the cost of significantly increased Fault Trigger Rates (e.g., 0.0778 and 0.0631, respectively), indicating overly aggressive action triggering and reduced reliability.

We further compute the \textbf{Consistency Difference} for each reward on \textbf{ABCD+} (training 100 + test 50), as shown in Table~\ref{tab:ac_difference_ablation_reward_type}: 

\begin{table}[t!]
\centering
\small
\setlength{\tabcolsep}{6pt}
\begin{tabular}{l|c}
\toprule
\textbf{Reward Type} & \textbf{Consistency Difference} ($\downarrow$) \\
\midrule
\textbf{Custom RULER}            & \textbf{0.084 $\pm$ 0.054} \\
RAC                              & 0.088 $\pm$ 0.088 \\
General RULER                    & 0.111 $\pm$ 0.063 \\
Adaptive Metric                  & 0.113 $\pm$ 0.094 \\
Weighted Metric (Max RAC)        & 0.323 $\pm$ 0.078 \\
Max RAC                          & 0.540 $\pm$ 0.132 \\
Adaptive RULER                   & 0.664 $\pm$ 0.101 \\
\bottomrule
\end{tabular}
\caption{Consistency Difference on ABCD+ (training 100/test 50) for reward type decision.}
\label{tab:ac_difference_ablation_reward_type}
\end{table}

Custom RULER achieves the lowest AC Difference among all reward types, indicating the most consistent alignment between AC and Max AC and supporting its selection as a safe and robust reward choice before scaling.

\paragraph{Data Scaling Effects.} \label{data_scaling_effects}

Next, we analyze the effect of scaling training data for Custom RULER on the ABCD+ dataset, ranging from training 300 + test 100 to the full dataset (training 5647 + test 692).

\textbf{Action Consistency:} As training data on ABCD+ increases, Action Consistency remains largely stable. AC improves from 0.554 (100 samples) to 0.597 (1k samples), and while it decreases at larger scales (e.g., 0.320 at 3k, 0.426 at full scale), the corresponding AC Difference stays moderate (e.g., 0.079--0.136), suggesting that the gap between AC and Max AC does not widen significantly. This indicates that the model does not collapse into optimizing a narrow subset of actions; this behavior aligns with our design goal: Custom RULER prioritizes proactive decision-making over strictly maintaining alignment with reference actions at every turn.

\textbf{Timing Behavior:} Proactive Timing consistently improves with more data, increasing from 0.155 (100 samples) to 0.235 at full scale, demonstrating stronger anticipation of future-ready actions. This improvement is accompanied by a moderate increase in Fault Trigger Rate (from 0.010 to 0.071), but the values remain controlled. Overall, scaling data strengthens proactive timing while preserving reasonable action consistency, aligning with the intended design of Custom RULER.

\section{Further Discussions}

This section provides additional analysis of several design decisions and evaluation factors that influence RL training performance. Section~\ref{reward_granularity} examines our reward granularity choices through controlled experiments, while Section~\ref{grpo_setting} discusses the rationale behind selecting the importance sampling ratio cap $\bar r_u^{(k)}$. We also analyze two complementary aspects of the evaluation pipeline: Section~\ref{sec:human_annotation_protocol} introduces a human annotation protocol for validating proactive timing references, and Section~\ref{sec:judger_model} investigates the impact of different LLM judge models on RL training outcomes.

\subsection{Reward Granularity Decision} \label{reward_granularity} 

To assess the effectiveness of trajectory-level versus turn-level reward formulations, we construct two comparative experimental groups. Each group incorporates both metric-based rewards (RAC and Max RAC) and RULER-based rewards (General RULER and Custom RULER). Experiments are conducted on a reduced ABCD+ subset comprising 100 training dialogues and 50 validation dialogues. We evaluate each reward design by tracking \textbf{four metadata-derived metrics} (Section~\ref{metadata_derived_metrics}) on the training set throughout the learning process.

We collect RL training statistics (training reward, loss, entropy) together with metadata-derived metrics (proactive timing, AC, Max AC) for each run. Each comparison group organizes these six metric diagrams in this way: Top row displays timing reward and training reward, demonstrating proactive timing and training performance stability differences. Middle row shows RAC score and Max RAC score evolution, indicating the effectiveness of proactive action consistency. Bottom row presents training loss and entropy metrics.

\subsubsection{RAC Group}

\begin{figure}[t!]
  \centering
  \includegraphics[width=\columnwidth]{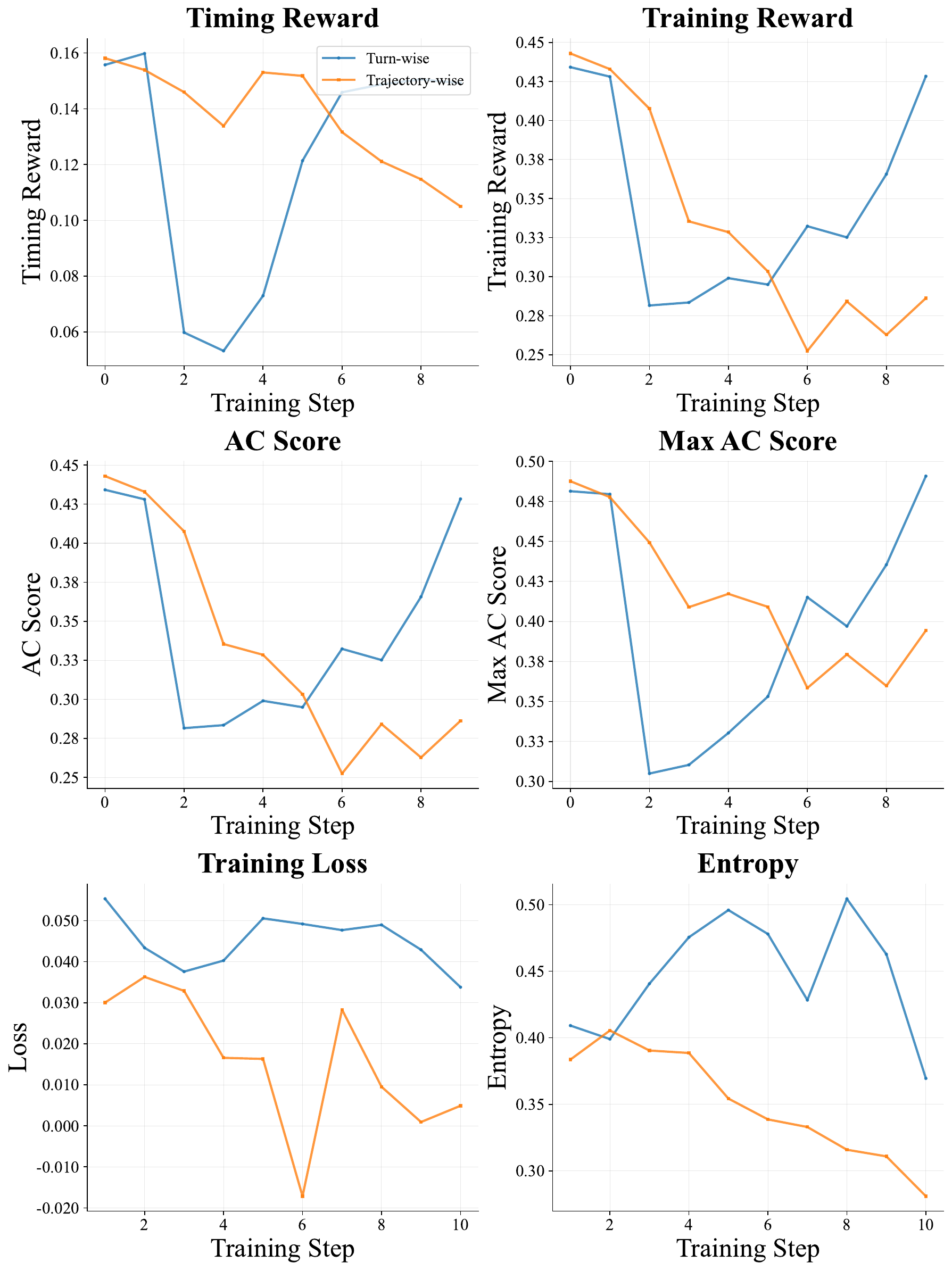}
  \caption{Turn-level(blue) vs Trajectory-level(orange) \textbf{RAC} Training Comparison. }
  \label{fig:turn_vs_trajectory_rac_comparison}
\end{figure}

Beside Figure~\ref{fig:turn_vs_trajectory_rac_comparison}, a brief summary in Table~\ref{tab:rac_stability} shows that the \textbf{turn-wise advances in exploration diversity and reward progression}, while the \textbf{trajectory-level} demonstrates superior \textbf{better convergence} (64\% improvement) and \textbf{AC consistency} with 37\% lower variance.

\begin{table}[t]
\centering
\small
\resizebox{\columnwidth}{!}{
\begin{tabular}{@{}lccc@{}}
\toprule
\textbf{Metric} & \textbf{Turn-level} & \textbf{Trajectory-level} & \textbf{Better} \\
\midrule
Loss Stability & 0.0450$\pm$0.0066 & \textbf{0.0159$\pm$0.0168} & Trajectory-level \\
Entropy Diversity & \textbf{0.4463$\pm$0.0445} & 0.3502$\pm$0.0411 & Turn-level \\
Reward Progression & \textbf{Positive trend} & Declining trend & Turn-level \\
Timing Reward & 0.121$\pm$0.0427 & 0.1369$\pm$0.0186 & Trajectory-level \\
AC Consistency & 0.3473$\pm$0.0625 & 0.3336$\pm$0.0703 & Turn-level \\
Max AC Performance & 0.3998$\pm$0.0721 & 0.4142$\pm$0.0452 & Trajectory-level \\
\bottomrule
\end{tabular}
}
\caption{Training stability comparison between turn-level and trajectory-level RAC reward computation.}
\label{tab:rac_stability}
\end{table}

\subsubsection{Max RAC Group}

\begin{figure}[t!]
  \centering
  \includegraphics[width=\columnwidth]{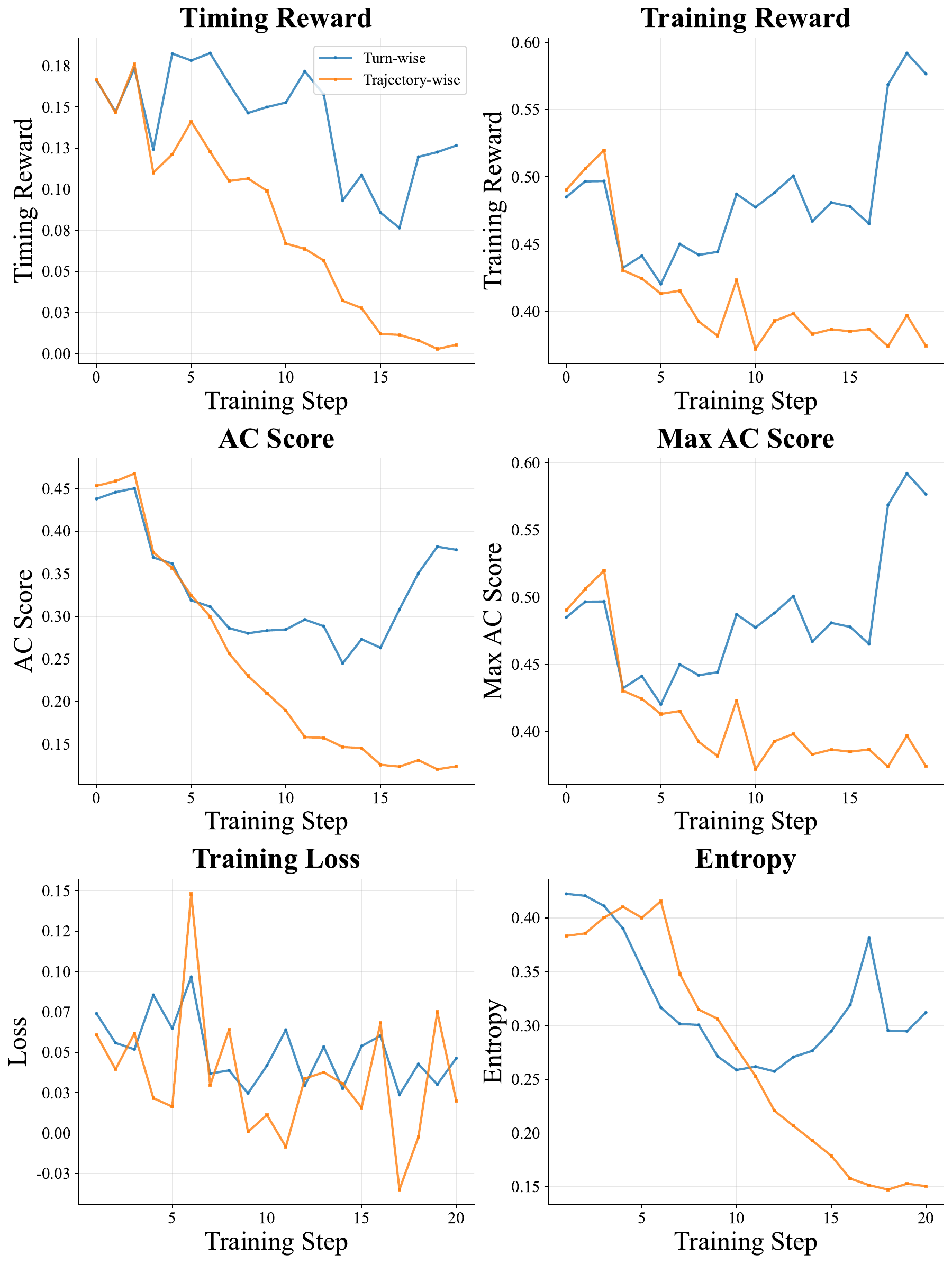}
  \caption{Turn-level(blue) vs Trajectory-level(orange) \textbf{Max RAC} Training Comparison.}
  \label{fig:turn_vs_trajectory_max_rac_comparison_acl}
\end{figure}

When it comes to Max RAC as shown in Figure~\ref{fig:turn_vs_trajectory_max_rac_comparison_acl}, \textbf{the turn-level reward demonstrates superior performance across 4 out of 5 metrics}, including 17.5\% higher Max AC performance, better training rewards, superior exploration stability, and more consistent AC scores, making it the optimal choice for Max RAC training despite trajectory-level showing better loss convergence. 

\begin{table}[t!]
\centering
\small
\resizebox{\columnwidth}{!}{
\begin{tabular}{@{}lccc@{}}
\toprule
\textbf{Metric} & \textbf{Turn-level} & \textbf{Trajectory-level} & \textbf{Better} \\
\midrule
Loss Stability & 0.0501$\pm$0.0201 & \textbf{0.0344$\pm$0.0390} & Trajectory \\
Training Reward & \textbf{0.4845$\pm$0.0469} & 0.4124$\pm$0.0439 & Turn-level \\
Entropy Diversity & \textbf{0.3205$\pm$0.0558} & 0.2728$\pm$0.1029 & Turn-level \\
Timing Reward & \textbf{0.1415$\pm$0.0328} & 0.0791$\pm$0.0576 & Turn-level \\
AC Consistency & \textbf{0.3308$\pm$0.0628} & 0.2428$\pm$0.1233 & Turn-level \\
Max AC Performance & \textbf{0.4845$\pm$0.0469} & 0.4124$\pm$0.0439 & Turn-level \\
\bottomrule
\end{tabular}
}
\caption{Training stability comparison between turn-level and trajectory-level Max RAC reward.}
\label{tab:max_rac_stability}
\end{table}

\subsubsection{General RULER Group}

\begin{figure}[t!]
  \centering
  \includegraphics[width=\columnwidth]{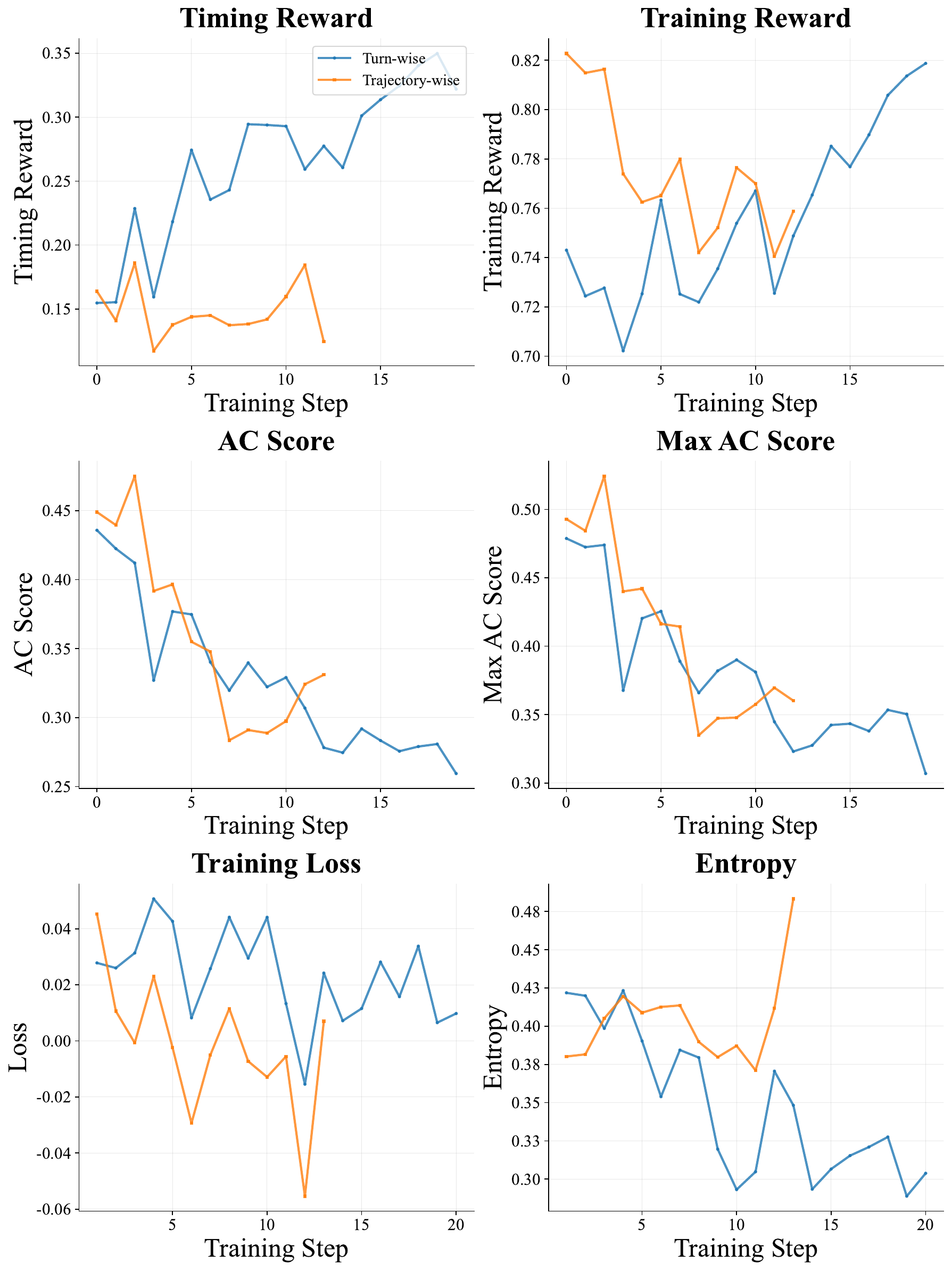}
  \caption{Turn-wise(blue) vs Trajectory-wise(orange) \textbf{General RULER} Training Comparison.}
  \label{fig:turn_vs_trajectory_general_ruler_comparison_acl}
\end{figure}

In our experiments, we observe that the trajectory-level context length grows rapidly as training progresses. As shown in Figure~\ref{fig:turn_vs_trajectory_general_ruler_comparison_acl}, even at this smaller scale, the context length explodes around training step 12, causing training to fail due to \textbf{out-of-memory (OOM) errors} on a 1$\times$H200 GPU. Although \textbf{trajectory-level rewards achieve better performance} in Table~\ref{tab:general_ruler_compatability}, we adopt \textbf{a conservative choice in favor of turn-level modeling for its stability and feasibility}.

\begin{table}[t!]
\centering
\small
\resizebox{\columnwidth}{!}{
\begin{tabular}{@{}lccc@{}}
\toprule
\textbf{Metric} & \textbf{Turn-level} & \textbf{Trajectory-level} & \textbf{Better} \\
\midrule
Loss Performance & 0.0232$\pm$0.0162 & \textbf{0.0016$\pm$0.0242} & Trajectory \\
Training Reward & 0.7560$\pm$0.0338 & 0.7750$\pm$0.0273 & Trajectory \\
Entropy Diversity & 0.3482$\pm$0.0468 & \textbf{0.4034$\pm$0.0289} & Trajectory \\
Timing Reward & \textbf{0.2649$\pm$0.0591} & 0.1477$\pm$0.0206 & Turn-level \\
AC Consistency & 0.3265$\pm$0.0532 & 0.3593$\pm$0.0657 & Trajectory \\
Max AC Performance & 0.3788$\pm$0.0513 & \textbf{0.4102$\pm$0.0631} & Trajectory \\
\bottomrule
\end{tabular}
}
\caption{Training stability comparison between turn-level and trajectory-level General RULER reward computation. }
\label{tab:general_ruler_compatability}
\end{table}

\subsubsection{Custom RULER Group}

\begin{figure}[t!]
  \centering
  \includegraphics[width=\columnwidth]{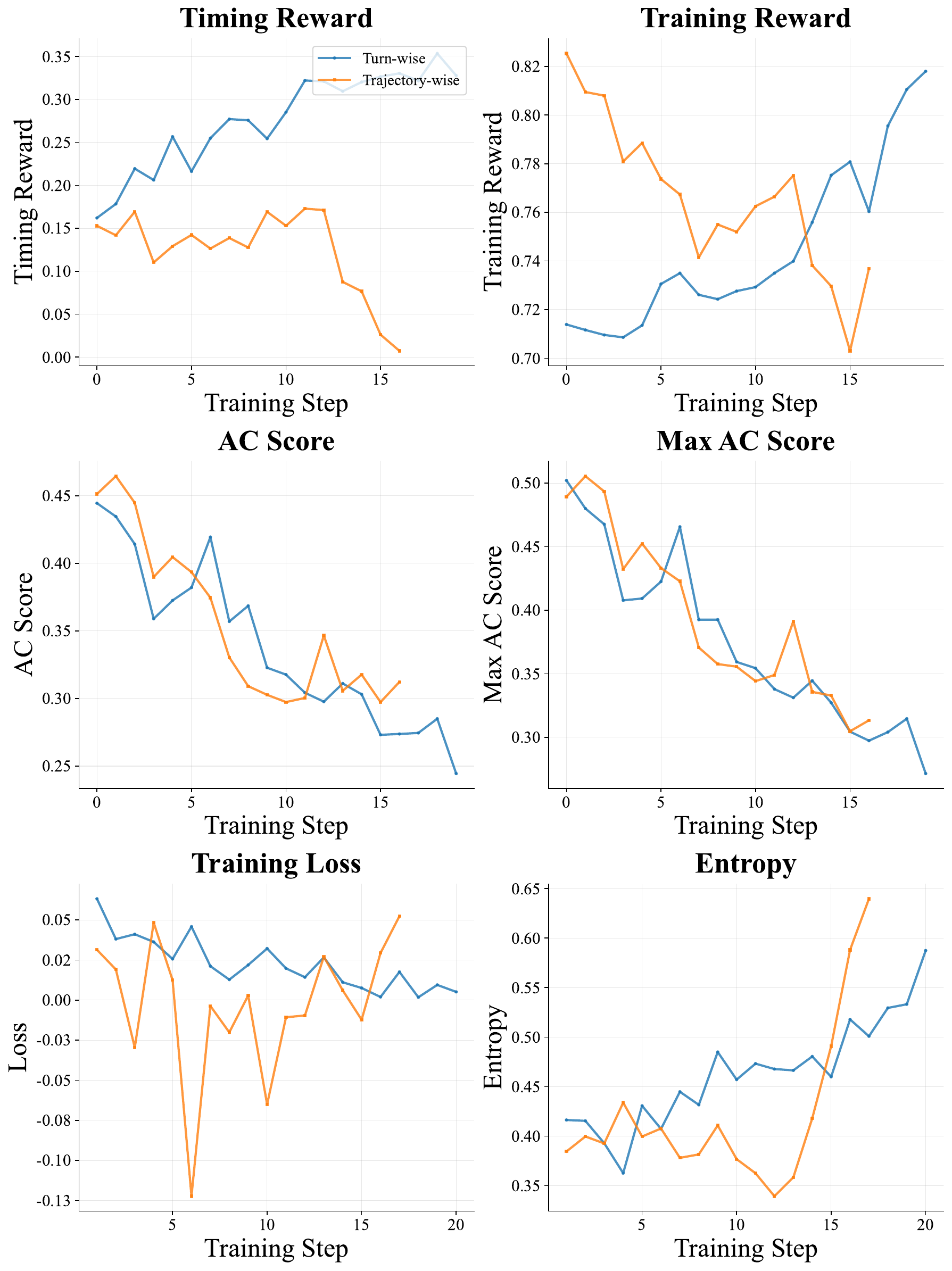}
  \caption{Turn-wise(blue) vs Trajectory-wise(orange) \textbf{Custom RULER} Training Comparison.}
  \label{fig:turn_vs_trajectory_custom_ruler_comparison_acl}
\end{figure}

Similar to the General RULER group, Figure~\ref{fig:turn_vs_trajectory_custom_ruler_comparison_acl} shows that training terminates abruptly around step 16 due to context explosion. Further analysis in Table~\ref{tab:custom_ruler_stability} indicates that turn-level rewards achieve superior entropy diversity and timing reward optimization, motivating our recommendation of turn-level Custom RULER despite the advantages of trajectory-level rewards on other metrics.

\begin{table}[t!]
\centering
\small
\resizebox{\columnwidth}{!}{
\begin{tabular}{@{}lccc@{}}
\toprule
\textbf{Metric} & \textbf{Turn-level} & \textbf{Trajectory-level} & \textbf{Better} \\
\midrule
Loss Performance & 0.023$\pm$0.016 & \textbf{0.003$\pm$0.043} & Trajectory \\
Training Reward & 0.745$\pm$0.034 & 0.766$\pm$0.032 & Trajectory \\
Entropy Diversity & 0.463 ± 0.054 & 0.421$\pm$0.080 & Turn-level \\
Timing Reward & \textbf{0.276$\pm$0.056} & 0.124$\pm$0.049 & Turn-level \\
AC Consistency & 0.338$\pm$0.059 & 0.355$\pm$0.059 & Trajectory \\
Max AC Performance & 0.374$\pm$0.068 & 0.393$\pm$0.065 & Trajectory \\
\bottomrule
\end{tabular}
}
\caption{Training stability comparison between turn-level and trajectory-level Custom RULER reward computation.}
\label{tab:custom_ruler_stability}
\end{table}

Considering the importance of training stability in 4$\times$H200 and 8$\times$H100 environments, together with the clear performance advantages of turn-level rewards, we adopt \textbf{turn-level modeling} as our initial strategy. This choice allows us to systematically study effective proactive timing optimization and establish a solid foundation for future extensions to \textbf{trajectory-guided multi-turn reward learning}.

\subsection{GRPO Parameters}\label{grpo_setting}

This section focuses on hyperparameter tuning on the GRPO algorithm.

\subsubsection{Importance Sampling Ratio Cap \texorpdfstring{$\bar r_u^{(k)}$}{ru(k) bar}} \label{importance_sampling_ratio_cap}

\paragraph{Experiment} To identify an appropriate value for the gamma clamp parameter $\gamma$ without degrading model performance consistency, we conduct a controlled comparison study using two configurations: $\gamma \in \{10, 100\}$.  We employ the Home Loan dataset at full scale with the Qwen2.5-14B-Instruct model using distributed data parallel (DDP) training~\cite{pytorchDDP}. Each $\gamma$ configuration is evaluated through two independent training runs to assess run-to-run consistency, resulting in four total experimental runs. We monitor eight key training metrics categorized into two groups: (1) \textit{ML statistics}, including training loss, entropy, reward, reward standard deviation, and policy loss; and (2) \textit{Proactive statistics}, including timing reward, AC score, and Max AC score. These metrics provide comprehensive coverage of both general machine learning training behavior and proactive agent-specific performance indicators. 

We further define {Combined Consistency Score} to measure the experiment that better captures the balance between performance value consistency and trajectory correlation.

\paragraph{Final Value Consistency Score} how similar the final training values are between runs within each group:
\begin{equation}
    C_{\text{final}}(\text{metric}) = 1 - \frac{|v_{1,\text{final}} - v_{2,\text{final}}|}{|v_{1,\text{final}}| + |v_{2,\text{final}}|}
\end{equation}

Where final values are extracted from the training time series:
\begin{equation}
    v_{i,\text{final}} = x_{i,U_i} \quad \text{where } U_i = \max\{u : x_{i,u} \text{ exists}\}
\end{equation}
and $v_{1,\text{final}}, v_{2,\text{final}}$ are the last recorded training values from run 1 and run 2, $x_{i,u}$ is the metric value at training step $u$ for run $i$, $U_i$ is the final training step for run $i$.

\paragraph{Trajectory correlation} measures how well the training curves follow similar patterns over time using
\begin{equation}
\resizebox{\columnwidth}{!}{$
C_{\text{trajectory}}(\text{metric}) = \rho(\mathbf{x}_1, \mathbf{x}_2) = \frac{\sum_{u=1}^{n}(x_{1,u} - \bar{x}_1)(x_{2,u} - \bar{x}_2)}{\sqrt{\sum_{u=1}^{n}(x_{1,u} - \bar{x}_1)^2 \sum_{u=1}^{n}(x_{2,u} - \bar{x}_2)^2}}
$}
\end{equation}
where $\rho$ is the \textbf{Pearson correlation coefficient}~\cite{pearson1895correlation}, $\mathbf{x}_1, \mathbf{x}_2$ are complete time series trajectories from run 1 and run 2, $x_{i,u}$ is the metric value at training step $u$ for run $i$, $\bar{x}_i = \frac{1}{n}\sum_{u=1}^{n} x_{i,u}$ is the mean value across all time steps for run $i$, $n$ is the number of overlapping training steps between runs.

\paragraph{Combined Consistency Score} averages final value consistency and trajectory correlation across all metrics:
\begin{equation}
\resizebox{\columnwidth}{!}{$
C_{\text{combined}}(\text{group}) = \frac{1}{2}\left(\frac{1}{m}\sum_{i=1}^{m} C_{\text{final}}(\text{metric}_i) + \frac{1}{m}\sum_{i=1}^{m} C_{\text{trajectory}}(\text{metric}_i)\right)
$}
\end{equation}
Where $m$ is the total number of metrics evaluated for the group, $\text{metric}_i$ represents the $i$-th training metric (loss, entropy, reward, etc.)

\paragraph{Performance Analysis}
Table~\ref{tab:gamma_consistency_compact} summarizes the consistency results across all monitored metrics. Overall, $\gamma=10$ demonstrates stable and reliable training behavior, characterized by uniformly high trajectory correlations ($\rho \in [0.650, 0.974]$) and strong final value alignment (consistency scores between 0.974 and 0.988). No anti-correlated behaviors are observed, and run-to-run variance remains low across metrics (average 1.4\%), indicating reproducible optimization dynamics. In contrast, $\gamma=100$ exhibits less predictable training behavior, with substantially more variable trajectory correlations ($\rho \in [0.362, 0.964]$) and an anti-correlated timing reward trajectory ($\rho=-0.586$), suggesting instability in timing-sensitive optimization. Although $\gamma=100$ achieves higher absolute performance on certain RAC-related metrics, its increased variance (2.4\%) and inconsistent dynamics reduce reliability, making $\gamma=10$ a more dependable choice for stable training.

\begin{table}[t!]
\centering
\small
\resizebox{\columnwidth}{!}{
\begin{tabular}{@{}lccc@{}}
\toprule
\textbf{Metric} & \textbf{Final} & \textbf{Trajectory} & \textbf{Variance} \\
& \textbf{Consistency} & \textbf{Correlation} & \textbf{(\%)} \\
\midrule
\multicolumn{4}{l}{\textit{$\gamma=10$ Group}} \\
\midrule
Training Loss & 0.974 & 0.899 & -- \\
Training Entropy & 0.978 & \textbf{0.650} & -- \\
Training Reward & \textbf{0.985} & \textbf{0.948} & \textbf{1.5} \\
RAC Score & 0.984 & 0.753 & 1.6 \\
Max RAC Score & \textbf{0.988} & 0.828 & \textbf{1.2} \\
Timing Reward & \textbf{0.962} & \textbf{0.974} & -- \\
\textbf{Average} & \textbf{0.975} & \textbf{0.840} & \textbf{1.4} \\
\midrule
\multicolumn{4}{l}{\textit{$\gamma=100$ Group}} \\
\midrule
Training Loss & \textbf{0.979} & 0.887 & -- \\
Training Entropy & 0.978 & 0.362 & -- \\
Training Reward & 0.952 & 0.945 & 4.8 \\
RAC Score & \textbf{0.998} & \textbf{0.964} & \textbf{0.2} \\
Max RAC Score & 0.978 & \textbf{0.951} & 2.2 \\
Timing Reward & 0.522 & -0.586 & -- \\
\textbf{Average} & 0.886 & 0.669 & 2.4 \\
\midrule
\multicolumn{4}{l}{\textit{Overall Comparison}} \\
\midrule
Combined Score & \multicolumn{3}{c}{\textbf{$\gamma=10$: 0.907} vs $\gamma=100$: 0.778} \\
Anti-Correlated Metrics & \multicolumn{3}{c}{\textbf{$\gamma=10$: 0} vs $\gamma=100$: 1} \\
Reproducibility & \multicolumn{3}{c}{\textbf{$\gamma=10$: High} vs $\gamma=100$: Moderate} \\
\bottomrule
\end{tabular}
}
\caption{Training consistency comparison between $\gamma=10$ and $\gamma=100$ groups arranged sequentially for single-column format. $\gamma=10$ achieves 16.6\% higher combined consistency score with no anti-correlated behaviors.}
\label{tab:gamma_consistency_compact}
\end{table}

\begin{figure}[t!]
  \centering

  \begin{minipage}{\columnwidth}
    \centering
    \includegraphics[width=\columnwidth]{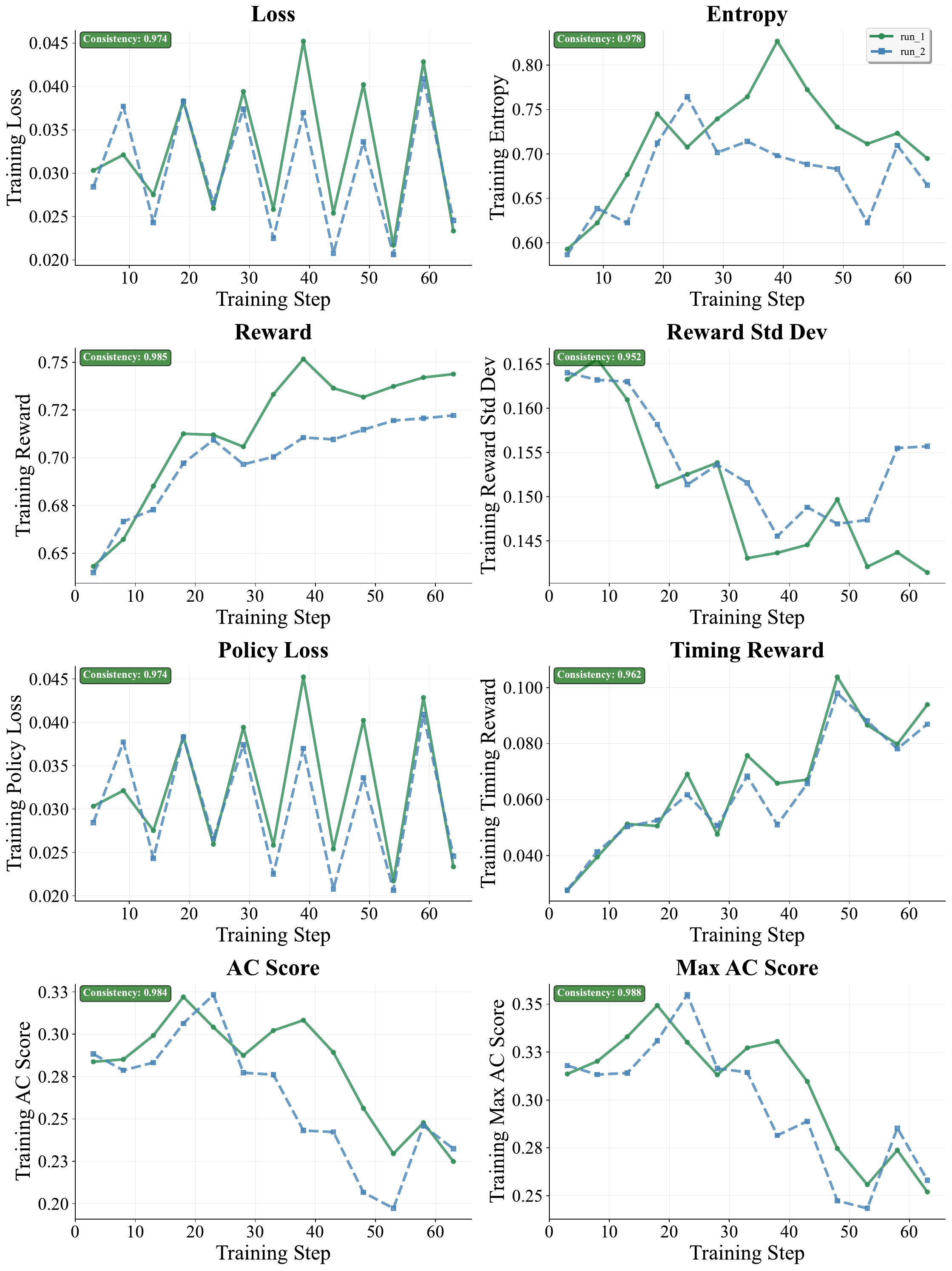}
    \vspace{-0.4em}
    {\footnotesize\textbf{(a)} $\bar r_u^{(k)} = 10$}
  \end{minipage}

  \vspace{0.6em}

  \begin{minipage}{\columnwidth}
    \centering
    \includegraphics[width=\columnwidth]{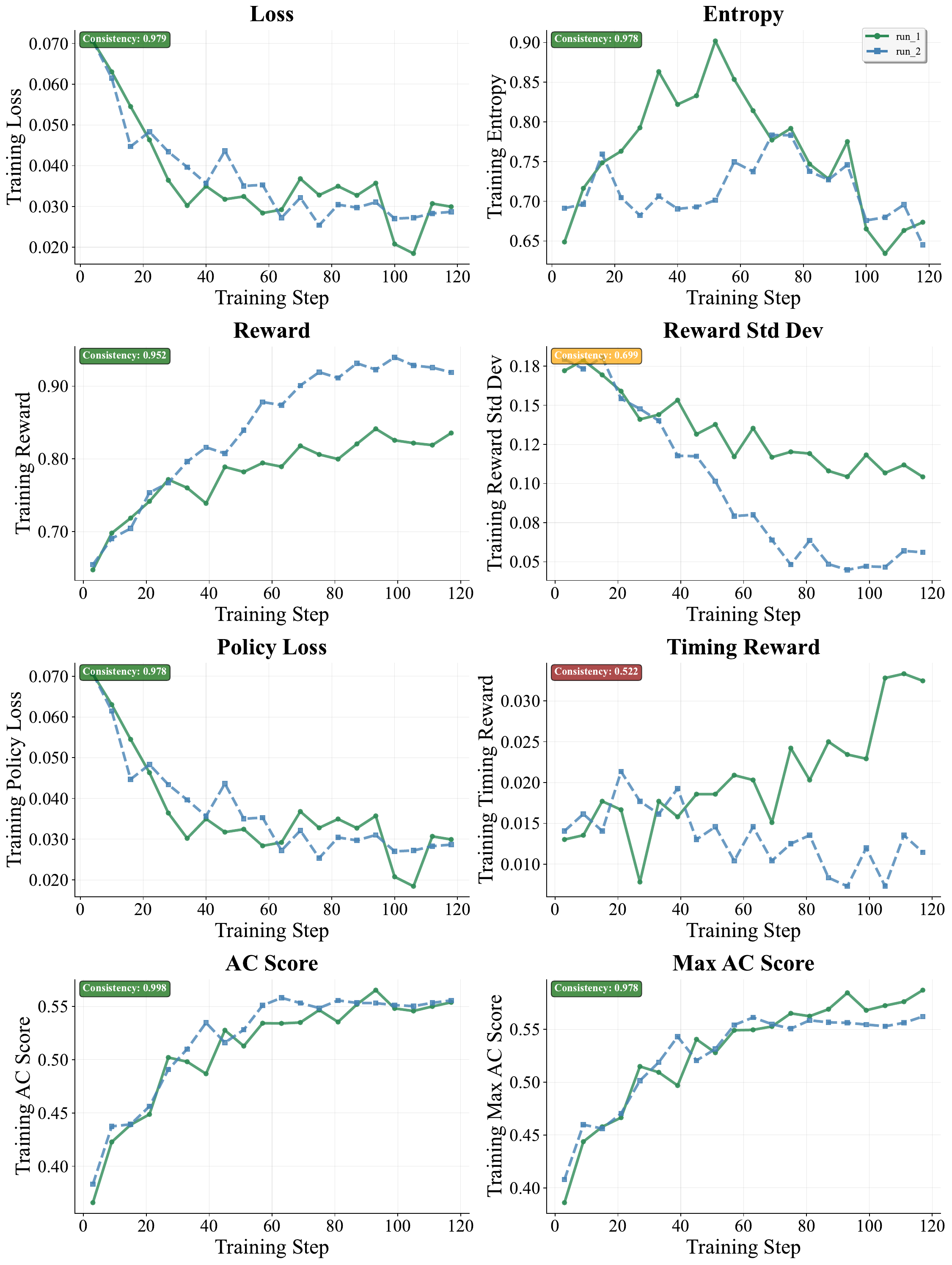}
    \vspace{-0.4em}
    {\footnotesize\textbf{(b)} $\bar r_u^{(k)} = 100$}
  \end{minipage}

  \caption{Training curve consistency analysis across two independent runs with different $\bar r_u^{(k)}$.}
  \label{fig:gamma_consistency}
\end{figure}

\paragraph{Training Dynamics Visualization}
Figure~\ref{fig:gamma_consistency} further illustrates the training dynamics under different importance sampling ratio caps. For $\gamma=10$, independent training runs closely follow similar trajectories across all metrics, exhibiting smooth convergence and consistent optimization patterns throughout training. By contrast, $\gamma=100$ shows larger fluctuations and increasingly divergent trajectories as training progresses, with pronounced instability in timing-related signals. These fluctuations indicate that higher ratio caps require more updates to stabilize and may amplify sensitivity to stochastic effects. Overall, the qualitative trends in Figure~\ref{fig:gamma_consistency} align with the quantitative consistency scores in Table~\ref{tab:gamma_consistency_compact}, reinforcing $\gamma=10$ as the more stable and reproducible configuration.

\section{Human Annotation Protocol for Proactive Timing Evaluation}\label{sec:human_annotation_protocol}

While our automated oracle pipeline enables scalable annotation (Section~\ref{sec:auto_data_pipeline}), human evaluation remains essential for validating annotation quality and establishing ground-truth timing references. This section describes a domain-agnostic protocol for human annotation of proactive action timing in task-oriented dialogues.

\subsection{Task Definition}

Given a dialogue transcript and the unified action catalog $\mathcal{A}$ (Section~\ref{metadata_annotation}), annotators identify proactive action opportunities at each dialogue turn and record the \textbf{earliest turn at which each action becomes ready to trigger}. This produces a human-verified reference ready action range that can be directly compared against oracle-generated annotations using the alignment metrics defined in Section~\ref{alignment_validation}.

\subsection{Human Annotation Procedure}

Human annotators process each dialogue sequentially, one turn at a time:

\begin{enumerate}
    \item \textbf{Action Candidate Initialization.} For each dialogue, the annotator receives a set of candidate actions from the action catalog. These may be pre-suggested by the automated pipeline or identified by the annotator during review. Annotators may also flag additional actions not in the initial candidate set.

    \item \textbf{Turn-Level Status Assignment.} At each dialogue turn $t$, the annotator assigns one of the following statuses to every tracked action $a \in \mathcal{A}$:
    \begin{itemize}
        \item \textsc{PENDING}: An opportunity for $a$ is identified (e.g., partial prerequisites observed), but one or more required parameters remain unresolved or unconfirmed.
        \item \textsc{READY\_TO\_TRIGGER}: All required parameters for $a$ are available and confirmed---the action can be appropriately triggered at turn $t$.
        \item \textsc{TRIGGERED}: The action was already executed by the human professional in the conversation.
        \item \textsc{DISMISSED}: A previously identified opportunity becomes irrelevant due to changed context (e.g., the user declines).
    \end{itemize}

    \item \textbf{Timing Record.} For each action that reaches \textsc{READY\_TO\_TRIGGER}, the annotator records:
    \begin{itemize}
        \item The \textit{earliest ready turn} $t^*_a$: the first turn where all prerequisites are satisfied.
        \item (Optional, if annotation budget permits) The full \textit{valid triggering window} $[t^*_a, t^{\text{end}}_a]$, capturing the range of turns over which triggering remains appropriate.
    \end{itemize}
    Recording only the earliest ready turn is sufficient for computing the core evaluation metrics (AC, PT, RAR); the full window enables finer-grained analysis of timing quality.

    \item \textbf{Supplementary Notes.} For actions that remain in \textsc{PENDING} throughout the dialogue (never reaching readiness) or that are \textsc{DISMISSED}, annotators provide a brief reason (e.g., ``missing required authorization,'' ``user declined'').
\end{enumerate}

\subsection{Annotator Requirements}

\begin{itemize}
    \item \textbf{Domain familiarity}: Annotators should understand the task domain sufficiently to judge whether action prerequisites are met from conversational context.
    \item \textbf{Catalog training}: Before annotation, each annotator completes a calibration session covering the action catalog, required vs.\ optional parameters, and status definitions.
    \item \textbf{Inter-annotator agreement}: A subset of dialogues (recommended $\geq$10\%) is independently annotated by at least two annotators to compute agreement on the earliest ready turn $t^*_a$, measured by exact match and within-$k$-turn tolerance.
\end{itemize}

\subsection{Output Format}

Each annotated dialogue produces a structured record per action:

\begin{table}[h!]
\centering
\small
\setlength{\tabcolsep}{4pt}
\begin{tabular}{p{2.2cm}|p{4.8cm}}
\hline\hline
\textbf{Field} & \textbf{Description} \\ \hline
Action & Action name from the unified catalog \\
First Identified & Turn where opportunity first observed (\textsc{PENDING}) \\
Earliest Ready & Turn where all prerequisites met ($t^*_a$) \\
Valid Window & (Optional) Range $[t^*_a, t^{\text{end}}_a]$ \\
Final Status & Terminal status (\textsc{TRIGGERED}/\textsc{DISMISSED}/\textsc{PENDING}) \\
Notes & Reason for non-triggering or special conditions \\ \hline
\end{tabular}
\caption{Per-action annotation output fields for human timing evaluation.}
\label{tab:human_annotation_output}
\end{table}

\subsection{Relationship to Automated Annotations}

Human annotations serve two complementary purposes: (1)~\textbf{validation}---comparing human-labeled earliest ready turns against oracle-generated reference ranges to quantify automated annotation quality using EC scores (Section~\ref{alignment_validation}), and (2)~\textbf{expanded coverage}---capturing valid timing opportunities that were never executed in the original dialogue and are therefore invisible to observation-based validation. This directly addresses the Action Observation Scope limitation discussed in Section~\ref{sec:limitations_observation_scope}: observed triggers represent only a subset of valid proactive opportunities, and human annotation can establish a more comprehensive ground truth.

\section{Judge Model Consideration}\label{sec:judger_model}

To examine the influence of different judge models on RL training outcomes, we run Adaptive RULER with Max RAC using three judge models from different providers—GPT-4.1-mini (OpenAI), Gemini-2.5-flash (Google), and Claude-Opus-4.6 (Anthropic)—on 100 training and 50 test dialogues. The resulting evaluation metrics are summarized in Table~\ref{tab:judge_comparison}.

\begin{table}[hbtp!]
\centering
\setlength{\tabcolsep}{4pt}
\small
\resizebox{\columnwidth}{!}{
\begin{tabular}{lccccc}
\toprule
\textbf{Judge} & \textbf{PT$\uparrow$} & \textbf{RAR$\uparrow$} & \textbf{AC$\uparrow$} & \textbf{MaxAC$\uparrow$} & \textbf{FTR$\downarrow$} \\
\midrule
GPT-4.1-mini & .190$\pm$.036 & .413$\pm$.087 & .426$\pm$.024 & .487$\pm$.027 & .057$\pm$.024 \\
Gemini-2.5-flash & .169$\pm$.013 & .380$\pm$.064 & .400$\pm$.032 & .456$\pm$.024 & .049$\pm$.021 \\
Claude-Opus-4.6 & .169$\pm$.009 & .365$\pm$.050 & .429$\pm$.017 & .478$\pm$.017 & .047$\pm$.016 \\
\bottomrule
\end{tabular}
}
\caption{Comparison of LLM judges on proactive action evaluation metrics.}
\label{tab:judge_comparison}
\end{table}

Across the three judge models, the resulting policies exhibit highly similar proactive behavioral profiles. Timing-related metrics remain stable (PT: 0.169–0.190, RAR: 0.365–0.413), indicating consistent learning of action timing signals. Action correctness is also comparable across judges: GPT-4.1-mini and Claude-Opus-4.6 produce nearly identical AC values (.426 vs .429), while Gemini-2.5-flash is modestly lower (~7\%), which likely reflects differences in how each model evaluates action quality. Overall, the results suggest that RL training under the Adaptive RULER objective is robust to reasonable variation in the choice of LLM judge.

\end{document}